\documentclass[]{fairmeta}

\usepackage{amsmath, amssymb, bm, bbm, mathrsfs}
\usepackage{siunitx}

\usepackage[utf8]{inputenc}
\usepackage[T1]{fontenc}
\usepackage{microtype}

\usepackage[dvipsnames,table]{xcolor}
\usepackage{soul}
\setul{0.5ex}{0.3ex}

\usepackage{booktabs, tabularx, threeparttable, multirow, array, colortbl, makecell}
\usepackage{tabularray}
\UseTblrLibrary{booktabs}

\usepackage{enumitem}
\setlist[enumerate]{noitemsep, topsep=0pt}
\setlist[itemize]{noitemsep, topsep=0pt}

\usepackage{subcaption}

\usepackage[ruled,vlined,algo2e]{algorithm2e}

\usepackage{url}
\usepackage{pifont}
\usepackage{setspace}
\usepackage{wrapfig}
\usepackage{tikz}
\usepackage{arydshln}
\usepackage{graphicx}
\usetikzlibrary{positioning}
\usepackage[most]{tcolorbox}
\usepackage{wasysym}

\makeatletter
\newcommand{\thickhline}{%
    \noalign {\ifnum 0=`}\fi \hrule height 1pt
    \futurelet \reserved@a \@xhline
}
\makeatother
\newcolumntype{I}{!{\vrule width 1pt}}

\usepackage{graphicx}
\usepackage{csquotes}
\usepackage{afterpage}
\usepackage{mathtools}
\usepackage{amsthm}
\usepackage{xspace}
\usepackage{nicefrac}
\newcommand{\chartfont}{\ttfamily}

\newcommand{\bench}{Tiresias-Bench}

\newcommand{\ms}[2]{#1\textsubscript{$\pm$#2}}

\makeatletter
\DeclareRobustCommand\onedot{\futurelet\@let@token\@onedot}
\def\@onedot{\ifx\@let@token.\else.\null\fi\xspace}

\makeatother

\definecolor{w_1}{RGB}{66,138,244}
\definecolor{w_2}{RGB}{73,144,245}
\definecolor{w_3}{RGB}{79,148,246}
\definecolor{w_4}{RGB}{87,154,247}
\definecolor{w_5}{RGB}{95,160,248}

\definecolor{uw}{RGB}{197,180,228}

\definecolor{NavigationColor}{HTML}{4E91C4}
\definecolor{RoboticsColor}{HTML}{E98445}
\definecolor{GamingColor}{HTML}{C94440}
\definecolor{DrivingColor}{HTML}{7562A8}
\definecolor{myblue}{HTML}{E9F2FA}

\definecolor{MetricPurple}{HTML}{7562A8}
\definecolor{MetricBlue}{HTML}{4E91C4}
\definecolor{MetricOrange}{HTML}{E98445}
\definecolor{MetricRed}{HTML}{C94440}
\definecolor{LegendGray}{HTML}{777777}
\definecolor{GridGray}{HTML}{D9D9D9}
\definecolor{TextGray}{HTML}{333333}

\definecolor{fbApp}{HTML}{c8e7fa}
\definecolor{fbPurple3}{HTML}{f0ebf5}

\definecolor{citecolor}{HTML}{0071BC}
\definecolor{linkcolor}{HTML}{ED1C24}

\definecolor{citecolor}{HTML}{0071BC}
\definecolor{linkcolor}{HTML}{ED1C24}

\usepackage{caption}
\usepackage{newfloat}
\usepackage{listings}
\usepackage{adjustbox}

\usepackage{algorithm}
\usepackage{algorithmic}

\usepackage{pgfplots}
\usepgfplotslibrary{groupplots,fillbetween}
\pgfplotsset{compat=1.18}

\definecolor{abRandom}{HTML}{79ADD2}
\definecolor{abOurs}{HTML}{E7847C}
\definecolor{abTr}{HTML}{F2B36F}
\definecolor{abGray}{HTML}{8A8A8A}
\definecolor{abBlue}{HTML}{4E91C4}
\definecolor{abOrange}{HTML}{E98445}
\definecolor{abPurple}{HTML}{7562A8}
\definecolor{abHighlight}{HTML}{EEEAF7}
\definecolor{abGrid}{HTML}{D9D9D9}
\definecolor{abText}{HTML}{333333}
\newlength{\abPlotWidth}
\newlength{\abPlotHeight}
\newlength{\abPlotGap}
\pgfplotsset{
 abAxis/.style={
  scale only axis,width=\abPlotWidth,height=\abPlotHeight,
  axis lines*=left,axis line style={black!45,line width=.4pt},
  tick align=outside,tick style={black!45,line width=.4pt},
  major tick length=2pt,
  ymajorgrids,grid style={abGrid,line width=.3pt},
  tick label style={font=\fontsize{7.5}{8}\selectfont,color=abText},
  label style={font=\fontsize{8}{9}\selectfont,color=abText},
  title style={font=\fontsize{8}{9}\selectfont\bfseries,yshift=1pt},
  xlabel style={at={(axis description cs:.5,-.21)},anchor=north},
  ylabel style={at={(axis description cs:-.18,.5)},anchor=south},
  scaled ticks=false,/pgf/number format/fixed,
  legend style={font=\fontsize{6.8}{7.5}\selectfont,draw=none,
    fill=none,inner sep=1pt,row sep=-1pt,column sep=1pt},
  legend cell align=left,legend image post style={xscale=.65},
  every axis plot/.append style={line width=.9pt,mark size=1.5pt},
 },
 abError/.style={error bars/.cd,y dir=both,y explicit,
  error bar style={black!65,line width=.4pt},
  error mark options={rotate=90,mark size=1.4pt,line width=.4pt}},
}

\DeclareCaptionStyle{ruled}{
    labelfont=normalfont,
    labelsep=colon,
    strut=off
}

\DeclareFloatingEnvironment[
    fileext=lst,
    placement=tb,
    name=Listing
]{listing}

\colorlet{PromptColor}{NavigationColor}

\newtcolorbox{annotationprompt}[1][NavigationColor]{
    enhanced,
    colback=#1!2,
    colframe=#1!38,
    boxrule=0.5pt,
    arc=1.5pt,
    left=3pt,
    right=3pt,
    top=0pt,
    bottom=0pt,
    before skip=0pt,
    after skip=0pt,
    before upper={%
        \colorlet{PromptColor}{#1}%
    }
}

\newcommand{\promptfield}[1]{%
    \par\addvspace{1pt}%
    \noindent
    \textbf{%
        \textcolor{PromptColor!75!black}{#1.}%
    }%
    \enspace
}

\newcommand{\promptsection}[1]{%
    \par\addvspace{1pt}%
    \noindent
    \textbf{%
        \textcolor{PromptColor!80!black}{#1}%
    }%
    \enspace
    {%
        \color{PromptColor!105}%
        \leaders\hrule height 0.45pt\hfill\kern0pt
    }%
    \par\addvspace{1pt}%
}

\newtcolorbox{promptstyle}[2]{
    enhanced,
    colback=PromptColor!#2,
    frame hidden,
    borderline west={1pt}{0pt}{PromptColor!80},
    sharp corners,
    left=1pt,
    right=1pt,
    top=0pt,
    bottom=0pt,
    before skip=2pt,
    after skip=2pt,
    title={\texttt{#1}},
    coltitle=PromptColor!65!black,
    fonttitle=\bfseries,
    attach title to upper,
    after title={.\quad}
}

\newtcolorbox{promptconstraint}{
    enhanced,
    colback=gray!7,
    frame hidden,
    borderline west={1pt}{0pt}{gray!40},
    borderline north={0.4pt}{0pt}{gray!20},
    borderline south={0.4pt}{0pt}{gray!20},
    sharp corners,
    left=1pt,
    right=1pt,
    top=0pt,
    bottom=0pt,
    before skip=2pt,
    after skip=2pt
}

\newtcolorbox{promptexample}[2]{
    enhanced,
    colback=PromptColor!#2,
    frame hidden,
    borderline west={1pt}{0pt}{PromptColor!65},
    sharp corners,
    left=1pt,
    right=1pt,
    top=0pt,
    bottom=0pt,
    before skip=2pt,
    after skip=2pt,
    title={\texttt{#1}},
    coltitle=PromptColor!65!black,
    fonttitle=\bfseries,
    attach title to upper,
    after title={:\quad}
}

\usepackage{orcidlink}                 

\DeclareRobustCommand{\afflogo}[2][1.3em]{%
  \ensuremath{\vcenter{\hbox{\includegraphics[height=#1]{#2}}}}}

\title{\textcolor[HTML]{F9746B}{Ti}\textcolor[HTML]{762A9D}{re}%
            \textcolor[HTML]{3F73B3}{si}\textcolor[HTML]{F1B287}{as}: A Benchmark and Agent for \\Language-Guided  Planning with World Models}

\author[1,*]{Fengyi~Wu}
\author[1,*]{Yifei~Dong}
\author[1,*]{Yetong~Sha}
\author[1]{Lingdong~Kong}
\author[1]{Qiyu~Hu}
\author[1]{Bohan~Xiong}
\author[2]{Qi~Dai}
\author[3]{Ruixiang~Tang}
\author[4]{Jiachen~Li}
\author[1,\dagger]{Zhi-Qi~Cheng}

\affiliation[1]{\afflogo[1.15em]{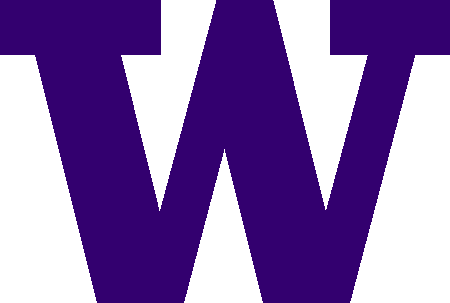}~University~of~Washington}
\affiliation[2]{\hspace{-0.25em}\smash{~\raisebox{0.1em}{\afflogo[1.3em]{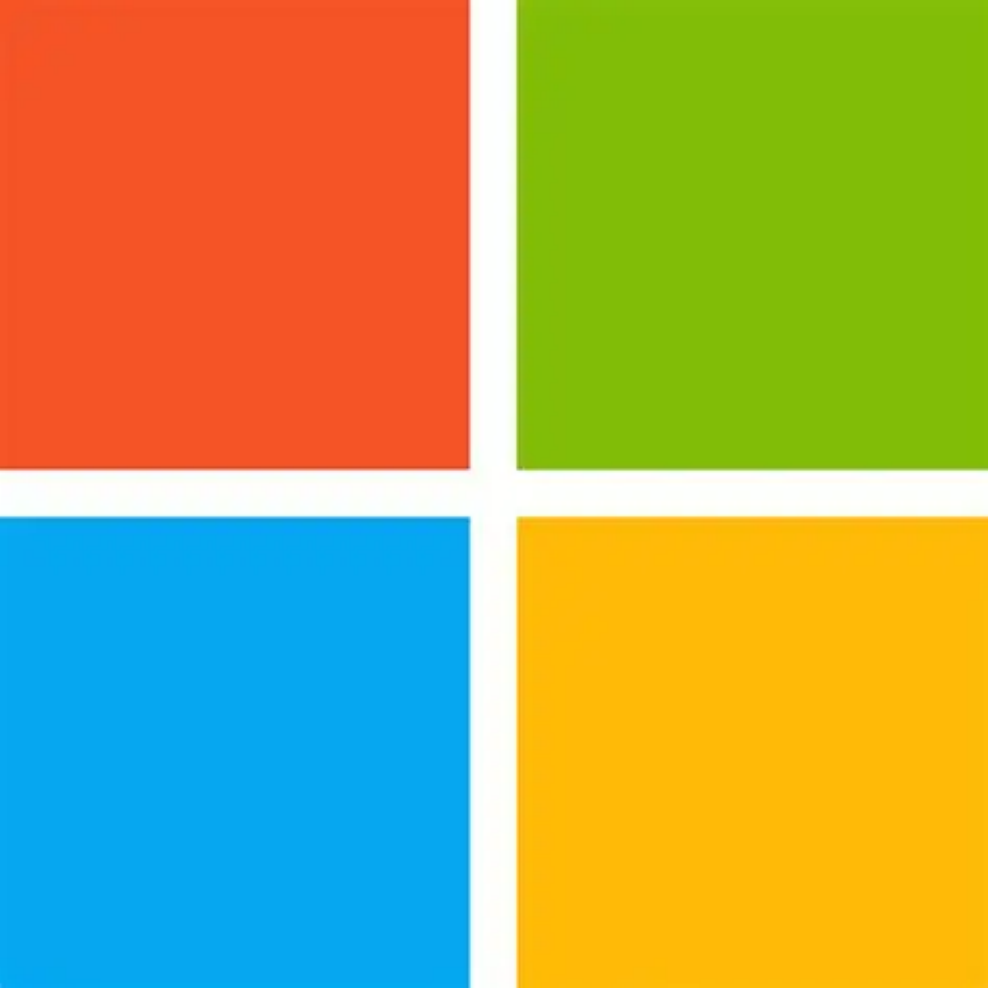}}}\hspace{-0.3em}~~Microsoft Research}
\affiliation[3]{\smash{\raisebox{0.1em}{\afflogo[1.65em]{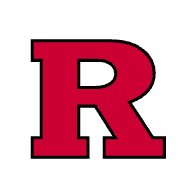}}}~Rutgers~University}
\affiliation[4]{\afflogo[1.4em]{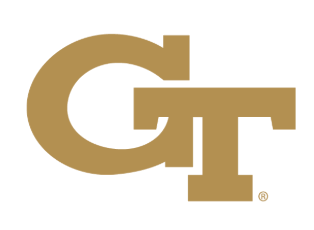}~Georgia~Institute~of~Technology}

\contribution[*]{Equal contribution}
\contribution[\dagger]{Corresponding author.}

\abstract{
World models enable planning through internal simulation, while natural language provides intuitive task specification. Yet, language-conditioned world modeling remains fragmented across domains with disparate action spaces, datasets, and evaluation protocols. We present Tiresias, a framework towards language-guided planning with world models that comprises a shared task, a benchmark, and a reference agent. The task, language-conditioned world modeling, maps an initial agent-view observation and a language instruction to an action sequence that fulfills the instruction. The benchmark, \bench{}, integrates 15 datasets across navigation, robotics, 3D gaming, and driving under a shared action interface and evaluation protocol, pairing 142{,}987 episodes with 428{,}961 instructions in three styles. The agent, Tiresias-Agent, plans via selective imagination: a language-conditioned diffusion policy proposes candidate trajectories, a trajectory ranker shortlists them, a video-pretrained world model imagines the visual consequences of the shortlist, and a visual ranker scores these rollouts; trajectory and visual scores jointly select the final plan. On \bench{}, Tiresias-Agent attains overall scores of 63 and 51 on the seen and unseen validation splits, improving over the strongest baseline by 9 points on each. Together, \bench{} and Tiresias-Agent provide a unified evaluation setting and a strong starting point for language-guided planning with world models.
}
  
\date{September 30, 2026}

\metadata[Code]{\url{https://github.com/UWMILab/Tiresias}}

\begin{document}

\maketitle
   
\section{Introduction}
\label{sec-intro}
\begin{flushright}
\small
\textit{``The blind seer, whose mind abides steadfast.''}\\
\textcolor{gray}{\small --- Homer, \textit{The Odyssey}, Book X,
trans.\ A.~T.~Murray}
\end{flushright}

Generalist agents operating in complex environments~\citep{chu2026agentic}, such as a robot in a kitchen, a vehicle in traffic, or an avatar in a game, require both world modeling and language grounding. World models predict how observations evolve under actions, enabling agents to consider possible outcomes before execution~\citep{brooks2024video,zhu2024sora,dong2026towards}, while language grounding connects natural-language instructions to intended behavior~\citep{wu2025govig}. Bringing these capabilities together requires more than generating visually plausible futures: agents need to identify actions that follow the instruction and assess their consequences in the current scene.

\begin{figure}[t]
    \centering
    \includegraphics[width=\linewidth]{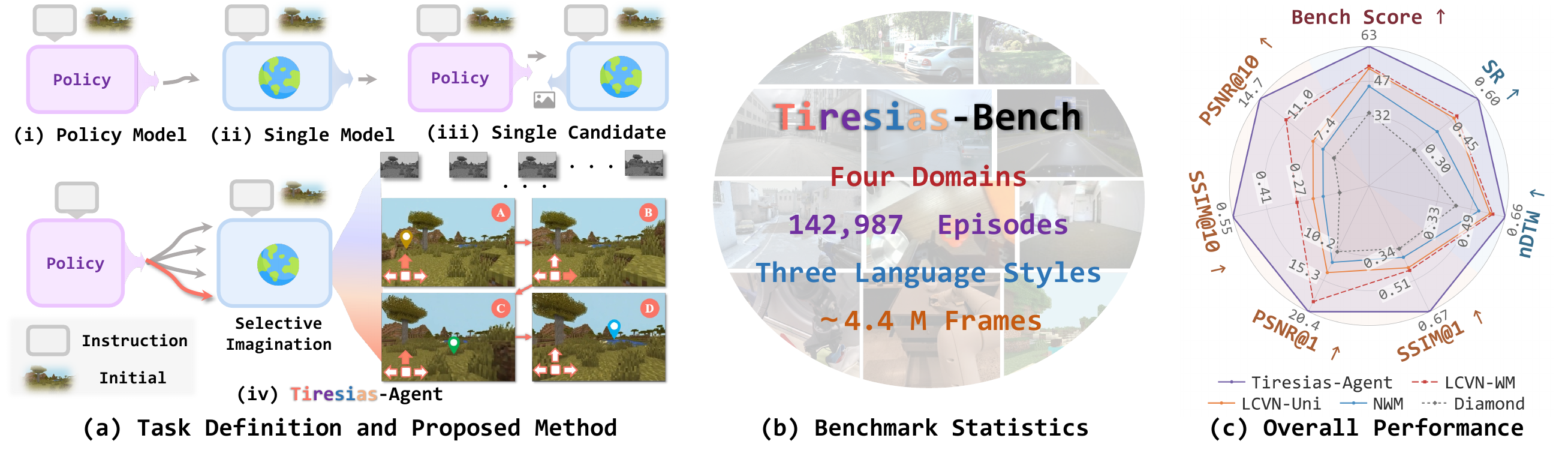}
    \caption{%
        \textbf{Overview of the \textbf{\textcolor[HTML]{F9746B}{Ti}%
        \textcolor[HTML]{762A9D}{re}%
        \textcolor[HTML]{3F73B3}{si}\textcolor[HTML]{F1B287}{as}} framework}.
        \textbf{(a)} Given an initial observation and a language
        instruction, the task is to predict an action sequence that
        fulfills the instruction. Unlike (i) policy-only models,
        (ii) single unified models, and (iii) policy--world-model
        pairs that roll out a single candidate, (iv) \textbf{Tiresias-Agent}
        proposes candidate trajectories and imagines a shortlisted subset to select the final plan (red).
        Frames A--D show the imagined rollout of the selected plan.
        \textbf{(b)} \textbf{Tiresias-Bench} unifies 15 datasets across four domains,
        with 142{,}987 episodes ($\sim$4.4M frames) annotated in
        three instruction styles.
        \textbf{(c)} Planning and imagination results on the \bench{} val seen
        split.
    }
    \label{fig:task}
\end{figure}

Existing approaches are largely evaluated within individual domains~\citep{dong2026language,zhou2024robodreamer,yu2025gamefactory,worldlens}: navigation, robotics, gaming, and driving differ in control interfaces, datasets, and metrics, which hinders comparison across different domains and leaves it unclear whether one approach can accommodate different embodiments and instruction styles.~A shared evaluation setting is therefore needed to study language-conditioned world modeling across domains.

We address this need with \textbf{Tiresias}, a framework named after the blind seer of Greek mythology, towards language-guided planning with world models that foresee the consequences of actions through imagination. It comprises a shared task, a benchmark, and a reference agent (Fig.~\ref{fig:task}). The task, language-conditioned world modeling, maps an initial agent-view observation and a language instruction to an action sequence that fulfills it (Fig.~\ref{fig:task}a). The benchmark, \textbf{Tiresias-Bench}, integrates 15 datasets across navigation, robotics, 3D gaming, and driving under a shared action parameterization, pairing 142{,}987 episodes with 428{,}961 instructions in three styles (Fig.~\ref{fig:task}b), and evaluates both planning and imagination on seen and unseen splits (Fig.~\ref{fig:task}c).

We consider three representative paradigms for language-guided planning (Fig.~\ref{fig:task}a): policy-only models~\citep{shah2022gnm,sridhar2024nomad}, single unified models that jointly predict actions and observations~\citep{cen2025worldvla,dong2026towards}, and world models paired with a single-rollout policy~\citep{dong2026language}. These methods commit to one plan without comparing alternatives by their imagined consequences under the instruction. \textbf{Tiresias-Agent} instead plans via selective imagination: a language-conditioned diffusion policy proposes candidate trajectories without observing the scene, a trajectory ranker shortlists them, a video-pretrained world model imagines the shortlist's consequences from the initial observation, and a visual ranker scores these rollouts; trajectory and visual scores jointly select the final plan.

On \bench{}, we benchmark representative methods from all four paradigms. Tiresias-Agent attains overall scores of 63 and 51 on the seen and unseen splits, and its rankers improve other proposal methods without further fine-tuning. Under our evaluation setting, language-only conditioning yields competitive proposal performance among the tested variants, language complements scene features in visual prediction, and landmark-based instructions support planning better than concise or intricate descriptions.

In summary, our main contributions are:

\noindent\textbf{(1)} We introduce \bench{}, bringing together 15 datasets across four domains under a shared action interface, with three instruction styles and a standardized protocol for evaluating language-guided planning and imagination.

\noindent\textbf{(2)} We propose Tiresias-Agent, which couples a language-conditioned diffusion policy with a video-pretrained world model through trajectory and visual rankers, enabling planning via selective imagination.

\noindent\textbf{(3)} We benchmark representative methods from four planning paradigms on \bench{}, and study the role of language in both imagination and planning.
\section{Related Work}
\label{sec-related-work}

\noindent \textbf{World Modeling.}~World models predict environment dynamics under agent actions, enabling outcomes to be simulated before execution~\citep{ha2018world}. The field has progressed from recurrent latent dynamics~\citep{ha2018world,hafner2024masteringdiversedomainsworld} and predictive Transformers~\citep{assran2023self,baldassarre2025back} to generative backbones acting as interactive simulators~\citep{bruce2024genie,bar2025navigation,agarwal2025cosmos,wan2025wan,xiong2026actworld}. Language increasingly complements action conditioning: actions specify local motion, while language provides a shared interface for expressing goals across embodiments. These advances now cover our four studied domains, including text-guided video plans for manipulation~\citep{du2023learning,zhou2024robodreamer}, prompt-enriched driving generation and planning~\citep{zhao2025drivedreamer,xiong2026unidrive,liu2026universe, liu2026driveva}, free-text game simulation~\citep{xiang2024pandora,yu2025gamefactory}, and instruction-grounded navigation futures~\citep{dong2026language}. These developments motivate studying how language-conditioned world models can support planning across domains with heterogeneous action spaces. We investigate this through selective imagination: Tiresias-Agent generates visual rollouts for shortlisted candidates and combines trajectory and visual scores to select a plan that fits the instruction.

\noindent \textbf{Benchmarks for World Models.}~Early evaluation followed video-generation protocols that score visual fidelity and prompt alignment~\citep{huang2023vbenchcomprehensivebenchmarksuite,duan2025worldscore}. As world models became interactive simulators, benchmarks turned to spatiotemporal consistency and physical plausibility~\citep{kwon2025stableworldmodelsmeasuring,lian2025toward}, action following~\citep{wang2024drivingdojo,yue2025ewmbench}, and functional utility such as closed-loop policy evaluation and procedural planning~\citep{zhang2025world,fan2026wow}. Most target a single domain such as driving, manipulation, or gaming. Existing multi-domain suites~\citep{wu2026omniworldbenchcomprehensiveinteractioncentricevaluation}
primarily evaluate visual rollouts conditioned on supplied actions or camera controls,
rather than the selection of actions to fulfill language instructions. Language-conditioned benchmarks include policy evaluation in manipulation, such as CALVIN~\citep{mees2022calvinbenchmarklanguageconditionedpolicy}, and world-model evaluation in navigation, such as LCVN~\citep{dong2026language}. \bench{} integrates 15 datasets across four domains with three instruction styles to jointly evaluate language-guided planning and visual prediction under a shared action interface and evaluation protocol.

\begin{figure*}[!t]
\setlength{\abovecaptionskip}{0pt}
\setlength{\belowcaptionskip}{0pt}
    \centering
    \includegraphics[width=1\linewidth]{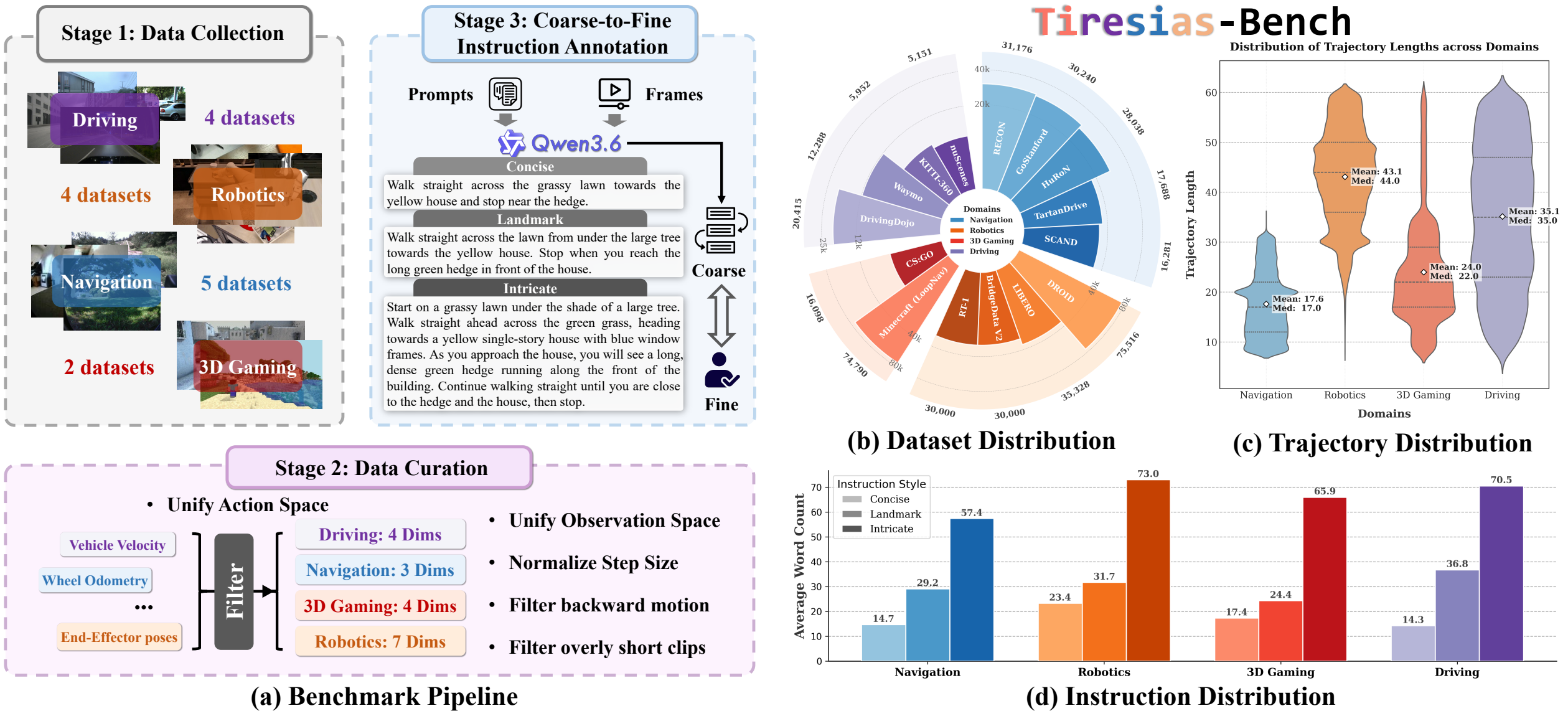}
   \caption{\textbf{Construction and statistics of \bench{}.}
    \textbf{(a)}~Dataset collection, action and observation standardization, and instruction annotation (Sec.~\ref{sec-construction}).
    \textbf{(b)}~Number of samples (episode--instruction pairs) contributed by each dataset, grouped by domain.
    \textbf{(c)}~Trajectory-length distributions (in steps) for each of the four domains.
    \textbf{(d)}~Average instruction length (in words) for each of the three instruction styles in every domain (Sec.~\ref{sec-stats}).}

    \label{fig-benchmark}
    \vspace{-0.1cm}
\end{figure*}
\section{\bench{}}
\label{sec-benchmark}

\subsection{Task Formulation}
\label{sec-task}
We formulate language-conditioned world modeling as follows.
Given an initial agent-view RGB observation $o_s$ and a natural-language
instruction $I$, an agent must predict an action sequence
$\hat{\mathcal{A}}_T$ that fulfills the instruction. We denote the
instruction and action sequence by:
\begin{equation}
    I=\langle w_1,\ldots,w_n\rangle,
    \qquad
    \hat{\mathcal{A}}_T=\langle\hat{a}_1,\ldots,\hat{a}_T\rangle.
\end{equation}
Each action $\hat{a}_t$ is a control command with a base parameterization $(u_t,\phi_t)$ shared across all domains, where $u_t\in\mathbb{R}^2$ is a planar displacement covering forward, backward, and lateral motions, and $\phi_t\in\mathbb{R}$ is a yaw rotation. Actions are executed sequentially, with a null action signaling episode termination. This task spans four domains,  $\mathcal{D}=\{\textit{Navigation},\textit{Robotics},\textit{3D Gaming},\textit{Driving}\}$, which share the base parameterization but incorporate additional domain-specific control channels: 3D gaming adds a vertical displacement $z_t\in\mathbb{R}$; driving introduces a scalar velocity $v_t\in\mathbb{R}$; and robotics, controlling an end-effector, appends the vertical displacement $z_t$  alongside roll, pitch, and a gripper command $(\rho_t,\vartheta_t,g_t)$. The raw controls from all sources are normalized into this shared parameterization (Sec.~\ref{sec-construction}).

\subsection{Benchmark Construction}

\label{sec-construction}

Instantiating this task requires episodes with a shared action interface across diverse domains. We build \bench{} in three stages (Fig.~\ref{fig-benchmark}a). 

\noindent \textbf{Dataset Collection.}~\bench{} integrates 15 datasets across four domains, mixing real and simulated sources for scene diversity and metric actions (Fig.~\ref{fig-benchmark}b). \textit{Navigation}: Go Stanford~\citep{hirose2018gonet}, RECON~\citep{shah2021rapid}, SCAND~\citep{karnan2022socially}, HuRoN~\citep{hirose2023sacson}, and TartanDrive~\citep{triest2022tartandrive} cover indoor, outdoor, social, and off-road ground-robot traversal. \textit{Robotics}: BridgeData~V2~\citep{walke2023bridgedata}, DROID~\citep{khazatsky2024droid}, RT-1~\citep{brohan2022rt}, and LIBERO~\citep{liu2023libero} provide 6-DoF gripper control across household and office scenes. \textit{3D Gaming}: Minecraft~\citep{lian2025toward} contributes open worlds, while CS:GO~\citep{pearce2021csgo} adds fast-paced human matches. \textit{Driving}: nuScenes~\citep{caesar2020nuscenes}, KITTI-360~\citep{liao2022kitti}, Waymo~\citep{sun2020scalability}, and Driving-Dojo~\citep{wang2024drivingdojo} span urban to suburban traffic, representing the benchmark's highest speeds. 
\begin{figure*}[!t] 
\setlength{\abovecaptionskip}{3pt}
\setlength{\belowcaptionskip}{0pt}
     \centering 
     \includegraphics[width=1\linewidth]{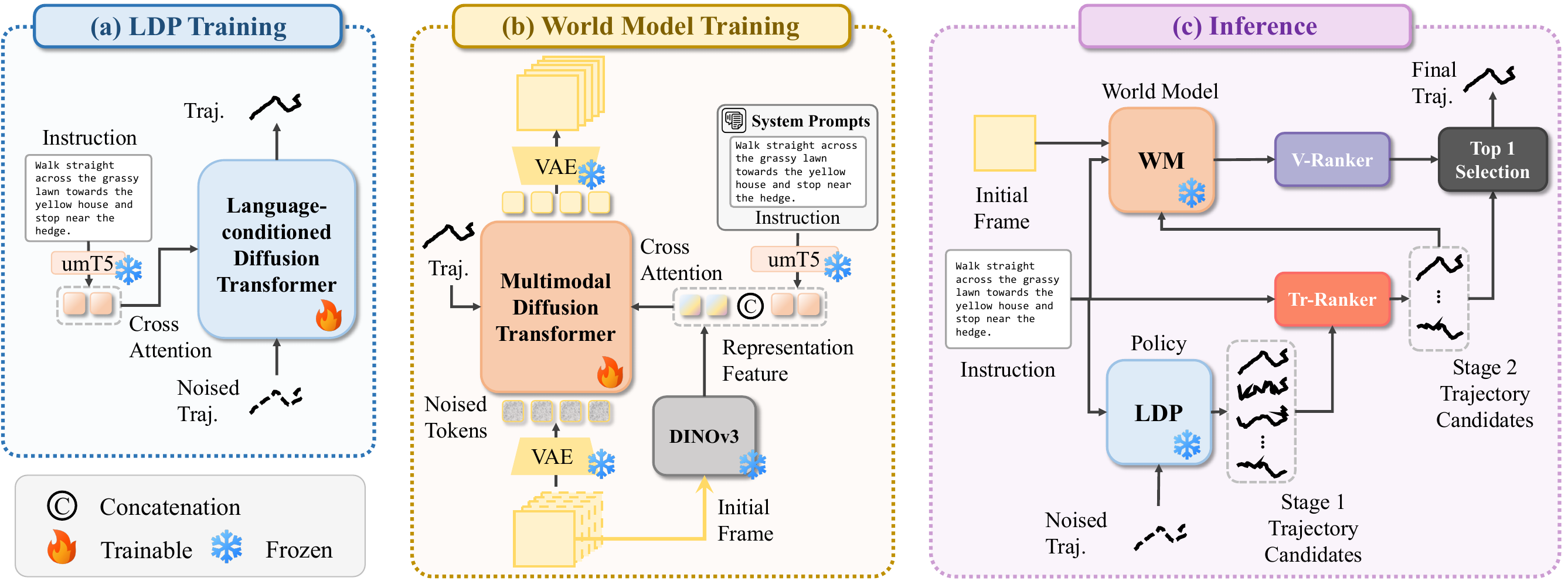} 
     \caption{\textbf{Overview of Tiresias-Agent.} \textbf{(a)}~The LDP learns to denoise action sequences given instructions. \textbf{(b)}~The Wan2.2-based world model predicts future observations given the initial frame, trajectory, and instruction. \textbf{(c)}~At inference, the Tr-Ranker shortlists LDP proposals for imagination; the fused trajectory and visual scores then select the final (Top 1 score) plan from the shortlist.}

     \label{fig:model} 
     \vspace{-0.1cm}
 \end{figure*}

\noindent \textbf{Shared Action and Observation Spaces.}~We standardize heterogeneous raw controls (e.g., wheel odometry, end-effector poses) via a shared pipeline. Episodes are converted into per-step motion increments in the local frame ($x$ forward, $y$ left, yaw counter-clockwise, $z$ up), derived from consecutive pose differences when explicit actions are absent. Following \citep{dong2026language}, displacements are normalized by each source's average step size to unify magnitudes. Agent-view frames are resized to $256\times256$. We then filter out backward, near-static, or overly short clips, segmenting the remainder into semantically coherent episodes following~\citep{bar2025navigation, dong2026language}.

\noindent \textbf{Instruction Styles.}~We augment each episode with language instructions in three styles, probing grounding at varying depths: \textbf{Concise} (essential directional cues); \textbf{Intricate} (rich descriptions of objects, layouts, and motion); and \textbf{Landmark-based} (anchoring maneuvers to salient landmarks, e.g., ``pass the parked van, then stop at the white door''). Instructions are generated via a coarse-to-fine pipeline. Across all three styles, instructions provide qualitative motion cues and landmark references without specifying exact distances, rotation angles, or action-step counts. In the coarse stage, Qwen3.6-27B~\citep{qwen3.6-27b} generates all three styles from sampled frames, supplemented by trajectory metadata and (for manipulation) native annotations. In the fine stage, these machine-generated instructions are manually verified against the videos for correctness and naturalness.~Each episode yields one episode–instruction pair per style.

\noindent \textbf{Evaluation Protocol and Metrics.}~Each domain is partitioned at the source level into train, val seen (held-out episodes from training sources), val unseen (sources completely excluded from training), and test (a mixture of both for future leaderboards). For instance, in navigation, TartanDrive is reserved for val unseen to evaluate generalization against significant visual domain shifts. We evaluate reference-trajectory agreement using four complementary metrics: SR~\citep{dong2026language}, nDTW~\citep{ilharco2019general}, ATE, and RPE~\citep{sturm2012evaluating}. These metrics quantify endpoint agreement and trajectory fidelity under a unified offline protocol, rather than directly measuring task completion through execution: SR (Success Rate) requires the terminal position error to be within one average step~\citep{dong2026language}, plus a matching terminal gripper state for robotics; nDTW measures path fidelity via order-preserving soft alignment; ATE and RPE compute the RMSE of absolute and relative position errors to capture global drift and local accuracy, respectively. The overall \bench{} score is the macro-average of SR and nDTW across the four domains.

\subsection{Benchmark Statistics}
\label{sec-stats}
Figs.~\ref{fig-benchmark}b--d summarize \bench{}. The benchmark comprises 142{,}987 episodes ($\sim$4.4M frames) from 15 datasets (12 real, 3 simulated), which the three-style annotation expands into 428{,}961 episode–instruction pairs, partitioned into training (300{,}276), val seen (38{,}607), val unseen (34{,}317), and test (55{,}761). Robotics contributes the largest share of samples (40\%), followed by navigation (29\%), gaming (21\%), and driving (10\%), and the split proportions are preserved within every domain. The domains complement one another in temporal structure (Fig.~\ref{fig-benchmark}c): robotics episodes are the longest (43.1 steps on average) while navigation ones are the shortest (17.6), and the three instruction styles occupy separated verbosity bands in every domain (Fig.~\ref{fig-benchmark}d), from 14--23 words for Concise to 57--73 for Intricate. 

Due to space limits, kindly refer to \textbf{Appendix} for more benchmark quality control and data splits details.

\section{Tiresias-Agent:~Planning with a Language-Conditioned World Model}

\label{sec-method}
\begin{figure*}[!t]
\setlength{\abovecaptionskip}{3pt}
\setlength{\belowcaptionskip}{3pt}
    \centering
    \includegraphics[width=1\linewidth]{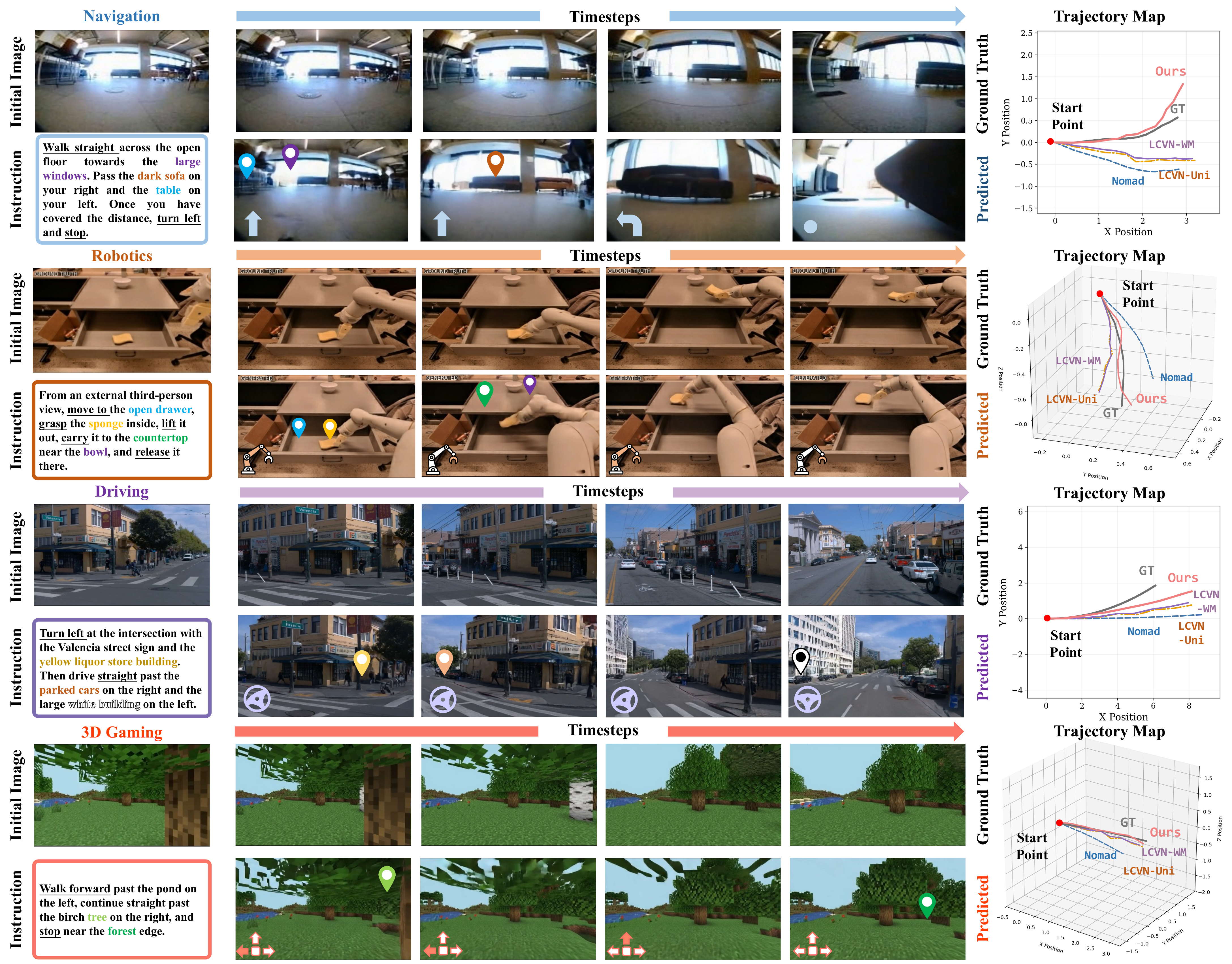}
    \caption{\textbf{Qualitative Results on Different Domains.} Given the initial observation and instruction,  we show the imagined rollout of the selected plan against the ground truth, and a comparison between the predicted trajectory and the corresponding reference trajectory.}
    \label{fig-visual_res-all}
    \vspace{-0.1cm}
\end{figure*}

We propose Tiresias-Agent for language-conditioned world modeling through selective imagination (Fig.~\ref{fig:model}). A language-conditioned diffusion policy (\textbf{LDP}) proposes candidate action sequences, and a trajectory ranker (\textbf{Tr-Ranker}) shortlists them using language and motion alone. A language-conditioned world model imagines the visual consequences of the retained candidates, which a visual ranker (\textbf{V-Ranker}) evaluates using scene and instruction context. The trajectory and visual scores jointly determine the final action sequence.

\subsection{Language-Conditioned Diffusion Policy}

\label{sec-policy}
The LDP (Fig.~\ref{fig:model}a) is a DiT-based~\citep{peebles2023scalable} diffusion policy~\citep{chi2025diffusion} that generates action sequences conditioned on language. Let $T_I=\mathrm{umT5}(I)$ denote instruction token embeddings from a frozen umT5 encoder~\citep{chung2023unimax}. Each DiT block attends to $T_I$ through cross-attention; apart from the diffusion timestep, no other conditioning is used.

For an expert action sequence $\mathcal{A}$, let $\mathbf{A}$ denote its standardized continuous controls. At diffusion step $j$, we construct $\mathbf{A}_j=\sqrt{\bar{\alpha}_j}\mathbf{A}+\sqrt{1-\bar{\alpha}_j}\epsilon$, where $\epsilon$ is standard Gaussian noise and $\bar{\alpha}_j$ is the cumulative signal-retention coefficient. Denoiser $\epsilon_\psi$ predicts the injected noise by minimizing
\begin{equation}
\mathcal{L}_{\mathrm{diff}}(\psi)
=
\mathbb{E}_{\mathcal{A},I,j,\epsilon}
\left[
\operatorname{MSE}_{\mathrm{valid}}
\big(\epsilon_\psi(\mathbf{A}_j,j,T_I),\epsilon\big)
\right],
\label{eq:ldp}
\end{equation}
where $\operatorname{MSE}_{\mathrm{valid}}$ averages over valid action steps and continuous control dimensions, excluding padding. For robotics, a separate gripper head is trained with binary cross-entropy. At inference, sampled continuous controls are mapped back to the task action representation and combined with gripper predictions where applicable to form candidate sequences.

\subsection{Language-Conditioned World Model}

\label{sec-wm}

The world model (Fig.~\ref{fig:model}b) adapts the pretrained Wan2.2 video backbone~\citep{wan2025wan} to predict future observations conditioned on actions, scene, and language. A frozen VAE encodes video frames into latents. An MLP embeds the candidate action sequence to modulate the diffusion Transformer, while cross-attention receives concatenated scene and language tokens. Scene tokens are extracted from the initial observation using DINOv3~\citep{simeoni2025dinov3}; language tokens encode the instruction preceded by a domain-specific prompt that specifies the embodiment context. The model generates
\begin{equation}
\hat{X} = (\hat{x}_{1},\dots,\hat{x}_{H}) \sim p_{\theta}\big(\cdot \mid x_{s},\, [f_{s};\, e_{I}],\, \hat{\mathcal{A}}\big),
\label{eq:wm}
\end{equation}
where $x_s$ and $f_s$ are the VAE latent and DINOv3 features of the initial observation $o_s$, respectively, and $e_I$ denotes prompted umT5 embeddings, distinct from the unprompted $T_I$. $[\,;\,]$ denotes token concatenation, and $H$ is the temporal length of the future latent sequence corresponding to $\hat{\mathcal{A}}$.
Training uses ground-truth future latents $X$ and expert actions $\mathcal{A}$, with conditions $c=(x_s,[f_s;e_I],\mathcal{A})$. We optimize:
\begin{equation}
\mathcal{L}_{\mathrm{wm}}(\theta) = \mathbb{E}_{X,\,c,\,\epsilon,\,\tau}\, \big\| u_{\theta}(X_{\tau}, \tau, c) - (\epsilon - X) \big\|_{2}^{2},
\label{eq:fm}
\end{equation}
where $\epsilon$ is standard Gaussian noise, $\tau\in[0,1]$ is the flow time, and $X_{\tau}=(1-\tau)X+\tau\epsilon$. The network $u_{\theta}$ learns the velocity along this data-to-noise interpolation.
\begin{table*}[t]
\setlength{\abovecaptionskip}{1pt}
\setlength{\belowcaptionskip}{2pt}
\caption{\textbf{Planning comparison} from different paradigms on val seen and val unseen splits of \bench{}. The score is the mean of SR and nDTW macro-averaged over four domains; higher is better for Score, SR, and nDTW, lower for ATE and RPE. Best results are in bold.}
\footnotesize
\centering
\setlength{\tabcolsep}{2pt}
\renewcommand{\arraystretch}{1.2}
\resizebox{\linewidth}{!}{%
\begin{tabular}{clIcIIcI*{4}{c}I*{4}{c}I*{4}{c}I*{4}{c}}
\hline\thickhline
\rowcolor{gray!20}
& & & \textbf{%
  \textcolor[HTML]{F9746B}{Ti}%
  \textcolor[HTML]{762A9D}{re}%
  \textcolor[HTML]{3F73B3}{si}%
  \textcolor[HTML]{F1B287}{as}%
  \textcolor{black}{-Bench}%
}
& \multicolumn{4}{cI}{\textbf{Navigation}}
& \multicolumn{4}{cI}{\textbf{Robotics}}
& \multicolumn{4}{cI}{\textbf{3D Gaming}}
& \multicolumn{4}{c}{\textbf{Driving}} \\
\cline{5-8}
\cline{9-12}
\cline{13-16}
\cline{17-20}
\rowcolor{gray!20}
& \multirow{-2}{*}{\textbf{Method}}
& \multirow{-2}{*}{\textbf{Venue}}
& \textbf{Score} (\%)
& SR & nDTW & ATE & RPE
& SR & nDTW & ATE & RPE
& SR & nDTW & ATE & RPE
& SR & nDTW & ATE & RPE \\
\hline\hline

& GNM (lang)~\citep{shah2022gnm}
& ICRA'23
& 36
& 0.24 & 0.44 & 1.10 & 0.34
& 0.25 & 0.42 & 0.74 & 0.33
& 0.25 & 0.39 & 3.18 & 0.48
& 0.36 & 0.52 & 1.29 & 0.27 \\

\rowcolor{gray!10}\cellcolor{white}& NoMaD (lang)~\citep{sridhar2024nomad}
& ICRA'24
& 40
& 0.28 & 0.48 & 0.95 & 0.29
& 0.29 & 0.45 & 0.68 & 0.31
& 0.27 & 0.41 & 3.05 & 0.45
& 0.42 & 0.57 & 1.14 & 0.24 \\

& WorldVLA~\citep{cen2025worldvla}
& NeurIPS'26
& 45
& 0.34 & 0.54 & 0.58 & 0.17
& 0.37 & 0.49 & 0.65 & 0.28
& 0.41 & 0.52 & 2.04 & 0.32
& 0.42 & 0.53 & 1.16 & 0.21 \\

\rowcolor{gray!10}\cellcolor{white}& LCVN-Uni~\citep{dong2026language}
& NeurIPS'26
& 53
& 0.43 & 0.58 & 0.39 & 0.11
& 0.43 & 0.56 & 0.50 & 0.21
& 0.49 & 0.57 & 1.96 & 0.28
& 0.52 & 0.63 & 0.89 & 0.16 \\

& Diamond~\citep{alonso2024diffusion}
& NeurIPS'24
& 33
& 0.19 & 0.40 & 1.31 & 0.40
& 0.20 & 0.37 & 0.82 & 0.36
& 0.31 & 0.45 & 2.88 & 0.42
& 0.29 & 0.46 & 1.48 & 0.30 \\

\rowcolor{gray!10}\cellcolor{white}& NWM~\citep{bar2025navigation}
& CVPR'25
& 45
& 0.31 & 0.50 & 0.72 & 0.23
& 0.33 & 0.49 & 0.64 & 0.29
& 0.40 & 0.53 & 2.30 & 0.35
& 0.46 & 0.60 & 1.04 & 0.22 \\

& RAE-NWM~\citep{zhang2026rae}
& ECCV'26
& 47
& 0.34 & 0.53 & 0.65 & 0.24
& 0.37 & 0.51 & 0.56 & 0.24
& 0.41 & 0.54 & 2.08 & 0.34
& 0.48 & 0.61 & 0.93 & 0.21 \\

\rowcolor{gray!10}\cellcolor{white}& NWM (lang)~\citep{bar2025navigation}
& CVPR'25
& 48
& 0.33 & 0.52 & 0.69 & 0.25
& 0.35 & 0.51 & 0.59 & 0.27
& 0.43 & 0.55 & 2.14 & 0.33
& 0.49 & 0.63 & 0.95 & 0.20 \\

& LCVN-WM~\citep{dong2026language}
& NeurIPS'26
& 54
& 0.44 & 0.59 & 0.37 & 0.12
& 0.44 & 0.57 & 0.47 & 0.22
& 0.51 & 0.58 & 1.88 & 0.29
& 0.54 & 0.65 & 0.83 & 0.17 \\
\cline{2-20}
\rowcolor{myblue!40}\cellcolor{white}
& Tiresias-Agent (RAE-NWM)
& --
& 56
& 0.53 & 0.61 & 0.44 & 0.15
& 0.48 & 0.56 & 0.45 & 0.19
& 0.55 & 0.57 & 1.87 & 0.29
& 0.55 & 0.63 & 0.82 & 0.18 \\

\rowcolor{myblue!60}\cellcolor{white}
& Tiresias-Agent (MPC)
& --
& 59
& 0.56 & 0.64 & 0.36 & 0.10
& 0.51 & 0.59 & 0.34 & 0.16
& 0.58 & 0.60 & 1.74 & 0.26
& 0.58 & 0.66 & 0.73 & 0.14 \\

\rowcolor{myblue!80}\cellcolor{white}\multirow{-12}{*}{\rotatebox[origin=c]{90}{\emph{\textbf{Val Seen}}}}
& \textbf{Tiresias-Agent (ours)}
& --
& \textbf{63}
& \textbf{0.60} & \textbf{0.67} & \textbf{0.35} & \textbf{0.09}
& \textbf{0.55} & \textbf{0.62} & \textbf{0.24} & \textbf{0.10}
& \textbf{0.63} & \textbf{0.64} & \textbf{1.49} & \textbf{0.19}
& \textbf{0.62} & \textbf{0.70} & \textbf{0.66} & \textbf{0.11} \\

\hline

& GNM (lang)~\citep{shah2022gnm}
& ICRA'23
& 28
& 0.12 & 0.33 & 2.45 & 0.81
& 0.18 & 0.34 & 0.93 & 0.42
& 0.19 & 0.33 & 3.58 & 0.55
& 0.27 & 0.45 & 1.69 & 0.35 \\

\rowcolor{gray!10}\cellcolor{white}& NoMaD (lang)~\citep{sridhar2024nomad}
& ICRA'24
& 31
& 0.14 & 0.36 & 2.28 & 0.74
& 0.21 & 0.38 & 0.86 & 0.39
& 0.20 & 0.35 & 3.46 & 0.52
& 0.31 & 0.50 & 1.52 & 0.32 \\

& WorldVLA~\citep{cen2025worldvla}
& NeurIPS'26
& 39
& 0.26 & 0.45 & 1.47 & 0.49
& 0.26 & 0.42 & 0.84 & 0.33
& 0.38 & 0.49 & 2.21 & 0.34
& 0.34 & 0.51 & 1.36 & 0.28 \\

\rowcolor{gray!10}\cellcolor{white}& LCVN-Uni~\citep{dong2026language}
&  NeurIPS'26
& 42
& 0.28 & 0.49 & 1.18 & 0.45
& 0.31 & 0.47 & 0.63 & 0.29
& 0.42 & 0.52 & 2.08 & 0.33
& 0.36 & 0.53 & 1.25 & 0.26 \\

& Diamond~\citep{alonso2024diffusion}
&  NeurIPS'24
& 24
& 0.10 & 0.31 & 2.76 & 0.91
& 0.12 & 0.28 & 1.01 & 0.45
& 0.22 & 0.37 & 3.35 & 0.50
& 0.16 & 0.33 & 1.95 & 0.41 \\

\rowcolor{gray!10}\cellcolor{white}& NWM~\citep{bar2025navigation}
& CVPR'25
& 33
& 0.15 & 0.37 & 2.10 & 0.70
& 0.22 & 0.39 & 0.82 & 0.38
& 0.30 & 0.45 & 2.62 & 0.41
& 0.29 & 0.46 & 1.47 & 0.31 \\

& RAE-NWM~\citep{zhang2026rae}
& ECCV'26
& 35
& 0.19 & 0.41 & 1.98 & 0.67
& 0.24 & 0.40 & 0.79 & 0.35
& 0.33 & 0.45 & 2.39 & 0.37
& 0.30 & 0.47 & 1.41 & 0.31 \\

\rowcolor{gray!10}\cellcolor{white}& NWM (lang)~\citep{bar2025navigation}
&  CVPR'25
& 35
& 0.16 & 0.39 & 2.13 & 0.68
& 0.24 & 0.41 & 0.76 & 0.35
& 0.33 & 0.48 & 2.44 & 0.38
& 0.32 & 0.50 & 1.35 & 0.29 \\

& LCVN-WM~\citep{dong2026language}
& NeurIPS'26
& 38
& 0.20 & 0.43 & 1.47 & 0.54
& 0.28 & 0.44 & 0.67 & 0.31
& 0.38 & 0.49 & 2.18 & 0.35
& 0.33 & 0.50 & 1.34 & 0.28 \\
\cline{2-20}
\rowcolor{myblue!40}\cellcolor{white}
& Tiresias-Agent (RAE-NWM)
& --
& 43
& 0.31 & 0.51 & 1.28 & 0.33
& 0.33 & 0.47 & 0.58 & 0.27
& 0.43 & 0.54 & 2.09 & 0.33
& 0.35 & 0.52 & 1.29 & 0.29 \\

\rowcolor{myblue!60}\cellcolor{white}
& Tiresias-Agent (MPC)
& --
& 46
& 0.38 & 0.56 & 0.94 & 0.18
& 0.35 & 0.49 & 0.43 & 0.20
& 0.48 & 0.55 & 1.84 & 0.28
& 0.37 & 0.53 & 1.15 & 0.24 \\

\rowcolor{myblue!80}\cellcolor{white}
\multirow{-12}{*}{\rotatebox[origin=c]{90}{\emph{\textbf{Val Unseen}}}}& \textbf{Tiresias-Agent (ours)}
& --
& \textbf{51}
& \textbf{0.41} & \textbf{0.61} & \textbf{0.89} & \textbf{0.17}
& \textbf{0.39} & \textbf{0.55} & \textbf{0.31} & \textbf{0.14}
& \textbf{0.54} & \textbf{0.61} & \textbf{1.61} & \textbf{0.25}
& \textbf{0.40} & \textbf{0.56} & \textbf{1.07} & \textbf{0.22} \\

\hline
\end{tabular}}
\label{tab:sota_plan}

\end{table*}

\subsection{Trajectory and Visual Rankers}
\label{sec-rankers}

The \textbf{Tr-Ranker} evaluates each candidate $\hat{\mathcal{A}}^{(m)}$ through scene-agnostic metric prediction~\citep{li2024hydra}. Action tokens attend to $T_I$, and a temporal Transformer aggregates their features for four heads predicting SR, nDTW, ATE, and RPE. Targets are computed offline by comparing LDP candidates with ground-truth trajectories. Training uses binary cross-entropy for SR and Huber losses~\citep{huber1992robust} for standardized continuous metrics. The predictions are combined on a common scale into $S^{(m)}_{\mathrm{tr}}$, rewarding success and path fidelity while penalizing trajectory errors.

The \textbf{V-Ranker} provides an additional ranking signal from imagined rollouts, conditioned on language and scene context~\citep{alakuijala2025video}. For each shortlisted candidate, Eq.~\ref{eq:wm} produces future latents $\hat{X}^{(m)}$. An MLP combines rollout statistics with pooled instruction features from $e_I$, scene features $f_s$, and a learned domain embedding. The episode domain determines both this embedding and the prompt used by the world model. Candidate-specific information enters the V-Ranker only through the imagined rollout, without direct access to the action sequence. An overall score and three auxiliary scores for language--motion consistency, path fidelity, and motion accuracy are combined into $S^{(m)}_{\mathrm{v}}$. 
Training uses trajectory-derived targets, combining listwise and pairwise margin-ranking losses \citep{joachims2002optimizing} on fused scores with MSE losses on fused and auxiliary predictions. 

\subsection{Inference with Selective Imagination}

\label{sec-inference}
At inference (Fig.~\ref{fig:model}c), all four independently trained modules remain frozen. The LDP samples $N$ candidates $\{\hat{\mathcal{A}}^{(m)}\}_{m=1}^{N}$ using independent noise seeds. The Tr-Ranker selects the $k$ highest-scoring candidates, forming the shortlist $\mathcal{S}\subseteq\{1,\dots,N\}$. Only these candidates are imagined by the world model and evaluated by the V-Ranker. Let $\widetilde{S}^{(m)}_{\mathrm{tr}}$ and $\widetilde{S}^{(m)}_{\mathrm{v}}$ denote the respective scores normalized within $\mathcal{S}$. We select the final action sequence through weighted score fusion:
\begin{equation}
\setlength\abovedisplayskip{-2pt}
\setlength\belowdisplayskip{0pt}
m^{\ast} = \arg\max_{m\in\mathcal{S}}\;
\lambda_{\mathrm{tr}}\widetilde{S}^{(m)}_{\mathrm{tr}}
+\lambda_{\mathrm{v}}\widetilde{S}^{(m)}_{\mathrm{v}},
\label{eq:select}
\end{equation}
\begin{equation}
\setlength\abovedisplayskip{0pt}
\setlength\belowdisplayskip{-3pt}
\hat{\mathcal{A}}^{\ast} = \hat{\mathcal{A}}^{(m^{\ast})},
\label{eq:select_tr}
\vspace{-0.1cm}
\end{equation}
where $\lambda_{\mathrm{tr}}$ and $\lambda_{\mathrm{v}}$ control the contributions of trajectory and visual scores. This design restricts visual generation to promising candidates while incorporating scene-grounded evidence into the final selection.

\section{Experiments}
\label{sec-exp}

\subsection{Experimental Setup} 
\begin{table*}[t]
\setlength{\abovecaptionskip}{3pt}
\setlength{\belowcaptionskip}{3pt}
\centering
\footnotesize
\caption{\textbf{Imagination comparison} from different paradigms on the val seen and val unseen splits of \bench{}, averaged over the four domains. $@n$ metrics score $n$-step open-loop rollouts under ground-truth actions against the real frames. Best results are in bold.}
\setlength{\tabcolsep}{3pt}
\renewcommand{\arraystretch}{1.2}
\resizebox{1\linewidth}{!}{%
\begin{tabular}{clIcIIcI*{4}{c}I*{4}{c}}
\hline\thickhline
\rowcolor{gray!20}
& & & &
\multicolumn{4}{cI}{One-step ($@1$)} &
\multicolumn{4}{c}{Rollout ($@10$)} \\
\cline{5-8} \cline{9-12}
\rowcolor{gray!20}
& \multirow{-2}{*}{\textbf{Method}}
& \multirow{-2}{*}{\textbf{Venue}}
& \multirow{-2}{*}{FVD$@32\downarrow$}
& SSIM$\uparrow$ & PSNR$\uparrow$ & LPIPS$\downarrow$ & DreamSim$\downarrow$
& SSIM$\uparrow$ & PSNR$\uparrow$ & LPIPS$\downarrow$ & DreamSim$\downarrow$ \\
\hline\hline

& WorldVLA~\citep{cen2025worldvla}
& NeurIPS'26
& 425.4 & 0.379 & 13.284 & 0.357 & 0.125
& 0.207 & 6.026 & 0.607 & 0.234 \\

\rowcolor{gray!10}\cellcolor{white}& LCVN-Uni~\citep{dong2026language}
& NeurIPS'26
& 344.8 & 0.438 & 14.018 & 0.287 & 0.096
& 0.224 & 7.562 & 0.492 & 0.214 \\

& Diamond~\citep{alonso2024diffusion}
& NeurIPS'24
& 648.7 & 0.336 & 10.642 & 0.401 & 0.128
& 0.118 & 4.731 & 0.668 & 0.301 \\

\rowcolor{gray!10}\cellcolor{white}& NWM~\citep{bar2025navigation}
& CVPR'25
& 382.1 & 0.382 & 12.364 & 0.326 & 0.111
& 0.184 & 6.218 & 0.621 & 0.242 \\

& RAE-NWM~\citep{zhang2026rae}
& ECCV'26
& 362.7 & 0.406 & 12.736 & 0.304 & 0.103
& 0.201 & 6.730 & 0.564 & 0.239 \\

\rowcolor{gray!10}\cellcolor{white}& NWM (lang)~\citep{bar2025navigation}
& CVPR'25
& 365.3 & 0.396 & 12.578 & 0.319 & 0.108
& 0.178 & 6.176 & 0.598 & 0.236 \\

& LCVN-WM~\citep{dong2026language}
& NeurIPS'26
& 321.6 & 0.452 & 18.724 & 0.268 & 0.101
& 0.289 & 11.204 & 0.471 & 0.219 \\

\rowcolor{myblue!80}\cellcolor{white}\multirow{-8}{*}{\rotatebox[origin=c]{90}{\emph{\textbf{Val Seen}}}}
& \textbf{Tiresias-Agent (ours)}
& --
& \textbf{274.0} & \textbf{0.674} & \textbf{20.351} & \textbf{0.242} & \textbf{0.089}
& \textbf{0.546} & \textbf{14.732} & \textbf{0.454} & \textbf{0.197} \\

\cline{2-12}
& WorldVLA~\citep{cen2025worldvla}
& NeurIPS'26
& 1816.3 & 0.348 & 11.102 & 0.392 & 0.149
& 0.127 & 5.075 & 0.649 & 0.302 \\

\rowcolor{gray!10}\cellcolor{white}& LCVN-Uni~\citep{dong2026language}
& NeurIPS'26
& 1563.7 & 0.366 & 12.384 & 0.342 & 0.112
& 0.174 & 6.614 & 0.548 & 0.238 \\

& Diamond~\citep{alonso2024diffusion}
& NeurIPS'24
& 2453.8 & 0.272 & 9.341 & 0.468 & 0.173
& 0.083 & 3.914 & 0.721 & 0.344 \\

\rowcolor{gray!10}\cellcolor{white}& NWM~\citep{bar2025navigation}
& CVPR'25
& 1869.3 & 0.318 & 10.982 & 0.381 & 0.142
& 0.137 & 5.421 & 0.681 & 0.286 \\

& RAE-NWM~\citep{zhang2026rae}
& ECCV'26
& 1712.9 & 0.354 & 11.256 & 0.378 & 0.143
& 0.151 & 5.802 & 0.631 & 0.275 \\

\rowcolor{gray!10}\cellcolor{white}& NWM (lang)~\citep{bar2025navigation}
& CVPR'25
& 1748.6 & 0.331 & 11.164 & 0.374 & 0.137
& 0.132 & 5.368 & 0.652 & 0.276 \\

& LCVN-WM~\citep{dong2026language}
& NeurIPS'26
& 1328.6 & 0.386 & 16.452 & 0.319 & 0.119
& 0.238 & 9.843 & 0.515 & 0.246 \\

\rowcolor{myblue!80}\cellcolor{white}\multirow{-8}{*}{\rotatebox[origin=c]{90}{\emph{\textbf{Val Unseen}}}}
& \textbf{Tiresias-Agent (ours)}
& --
& \textbf{744.9} & \textbf{0.550} & \textbf{19.695} & \textbf{0.303} & \textbf{0.088}
& \textbf{0.448} & \textbf{14.571} & \textbf{0.504} & \textbf{0.199} \\

\hline
\end{tabular}}
\label{tab:sota_imagine}

\end{table*}

\noindent \textbf{Baselines.}~We evaluate baselines spanning the three existing planning paradigms in Fig.~\ref{fig:task}.
For policy-only models, we adapt GNM~\citep{shah2022gnm} and NoMaD~\citep{sridhar2024nomad} to language goals by replacing their image-goal encoders with a frozen CLIP text encoder and projecting its outputs into the goal-token space.
For unified models, we include WorldVLA~\citep{cen2025worldvla} and LCVN-Uni~\citep{dong2026language}, which integrate action and observation prediction within a single autoregressive model.
For policy--world-model pairs, we couple the latent actor-critic LCVN-AC with five world models: Diamond~\citep{alonso2024diffusion} (UNet-based diffusion), NWM~\citep{bar2025navigation} (CDiT-based diffusion), RAE-NWM~\citep{zhang2026rae}, NWM (lang), which incorporates CLIP-encoded instructions through the CDiT AdaLN block, and LCVN-WM~\citep{dong2026language}.
We additionally evaluate two variants of our framework.
Tiresias-Agent (RAE-NWM) replaces our world model with RAE-NWM while retaining the LDP and two-stage ranking pipeline.
Tiresias-Agent (MPC) replaces the LDP with sampling-based MPC while retaining our world model and the same ranking pipeline, assessing the contribution of learned language-conditioned proposals.
All methods are trained on the same \bench{} training split and evaluated under a shared protocol. These comparisons aim to investigate the role of language in planning and imagination across different modeling paradigms, rather than to demonstrate advantages of any particular method or architecture.

\noindent \textbf{Evaluation Metrics.}~Planning is evaluated using \bench{} metrics detailed in Sec.~\ref{sec-benchmark}: SR, nDTW, ATE, RPE, and the overall score. We also assess world-model imagination quality: $@1$ metrics compare the predicted next frame against the ground truth, while $@n$ metrics evaluate $n$-step open-loop generation conditioned on ground-truth actions. We adopt SSIM, PSNR, LPIPS~\citep{zhang2018unreasonable}, and DreamSim~\citep{fu2023dreamsim} for $@1$ and $@10$ evaluations, alongside FVD~\citep{unterthiner2019fvd} at $@32$.

\noindent \textbf{Implementation Details.}~The world model is initialized from Wan2.2-TI2V-5B and LDP is trained from scratch; umT5, DINOv3, and VAE remain frozen throughout training. Visual encoders process all observations at a uniform $256\times256$ resolution, and scene features are extracted using a frozen DINOv3 ViT-L/16 distilled encoder (300M parameters). The world model is trained for 10 epochs on \bench{}, utilizing 8$\times$NVIDIA A100 GPUs (80GB). At inference, the LDP generates $N{=}128$ candidates using a fixed set of independent noise seeds. The Tr-Ranker retains top $k{=}8$ candidates for imagination, and final selection combines trajectory and visual scores with weights $\lambda_{\mathrm{tr}}{:}\lambda_{\mathrm{v}}=2{:}1$.
Due to space limits, kindly refer to \textbf{Appendix} for more implementation and experimental details.
\subsection{Performance Comparison on \bench{}}
\begin{table*}[!t]
\setlength{\abovecaptionskip}{4pt}
\setlength{\belowcaptionskip}{2pt}
\centering

\begin{minipage}[t]{0.49\textwidth}
\vspace{0pt}
\centering
\caption{Ablation on \textbf{cross-attention conditioning}
of the WM on val seen; system prompt
retained in all variants.}
\label{tab:abl_cond_scaled}

\small
\renewcommand{\arraystretch}{1.16}
\setlength{\tabcolsep}{3pt}
\resizebox{\linewidth}{!}{%
\begin{tabular}{lIIccIcc}
\hline\thickhline
\rowcolor{gray!20}
& \multicolumn{2}{cI}{\textbf{Planning}}
& \multicolumn{2}{c}{\textbf{Imagination}} \\
\cline{2-3}\cline{4-5}
\rowcolor{gray!20}
\multirow{-2}{*}{\textbf{WM Conditioning}}
& SR$\uparrow$ & nDTW$\uparrow$
& SSIM$@1\uparrow$ & FVD$@32\downarrow$ \\
\hline\hline
Instruction only
& 0.56 & 0.65 & 0.61 & 328.7 \\
DINOv3 only
& 0.54 & 0.64 & 0.58 & 367.4 \\
\rowcolor{myblue!80}
\textbf{Instruction + DINOv3}
& \textbf{0.60} & \textbf{0.66}
& \textbf{0.67} & \textbf{274.0} \\
\hline
\end{tabular}%
}
\end{minipage}
\hfill
\begin{minipage}[t]{0.49\textwidth}
\vspace{0pt}
\centering
\caption{Ablation on \textbf{LDP conditioning}:
random candidate execution versus two-stage
selection ($128 \rightarrow 8$).}
\label{tab:abl_ldp_cond}

\small
\renewcommand{\arraystretch}{1.1}
\setlength{\tabcolsep}{3pt}
\resizebox{\linewidth}{!}{%
\begin{tabular}{lIIccIcc}
\hline\thickhline
\rowcolor{gray!20}
& \multicolumn{2}{cI}{\textbf{Unranked exec.}}
& \multicolumn{2}{c}{\textbf{Full pipeline}} \\
\cline{2-3}\cline{4-5}
\rowcolor{gray!20}
\multirow{-2}{*}{\textbf{LDP Conditioning}}
& SR$\uparrow$ & nDTW$\uparrow$
& SR$\uparrow$ & nDTW$\uparrow$ \\
\hline\hline
Observation only
& 0.06 & 0.26 & 0.41 & 0.54 \\
\rowcolor{myblue!80}
Language only (default)
& \textbf{0.13} & \textbf{0.34}
& \textbf{0.60} & \textbf{0.66} \\
Language + Observation
& 0.11 & 0.31 & 0.56 & 0.63 \\
\hline
\end{tabular}%
}
\end{minipage}
\end{table*}

\begin{figure*}[!t]
\setlength{\abovecaptionskip}{3pt}
\setlength{\belowcaptionskip}{0pt}
\centering

\begin{minipage}[b]{0.57\textwidth}
\vspace{0pt}
    \centering
    \includegraphics[width=1\linewidth]{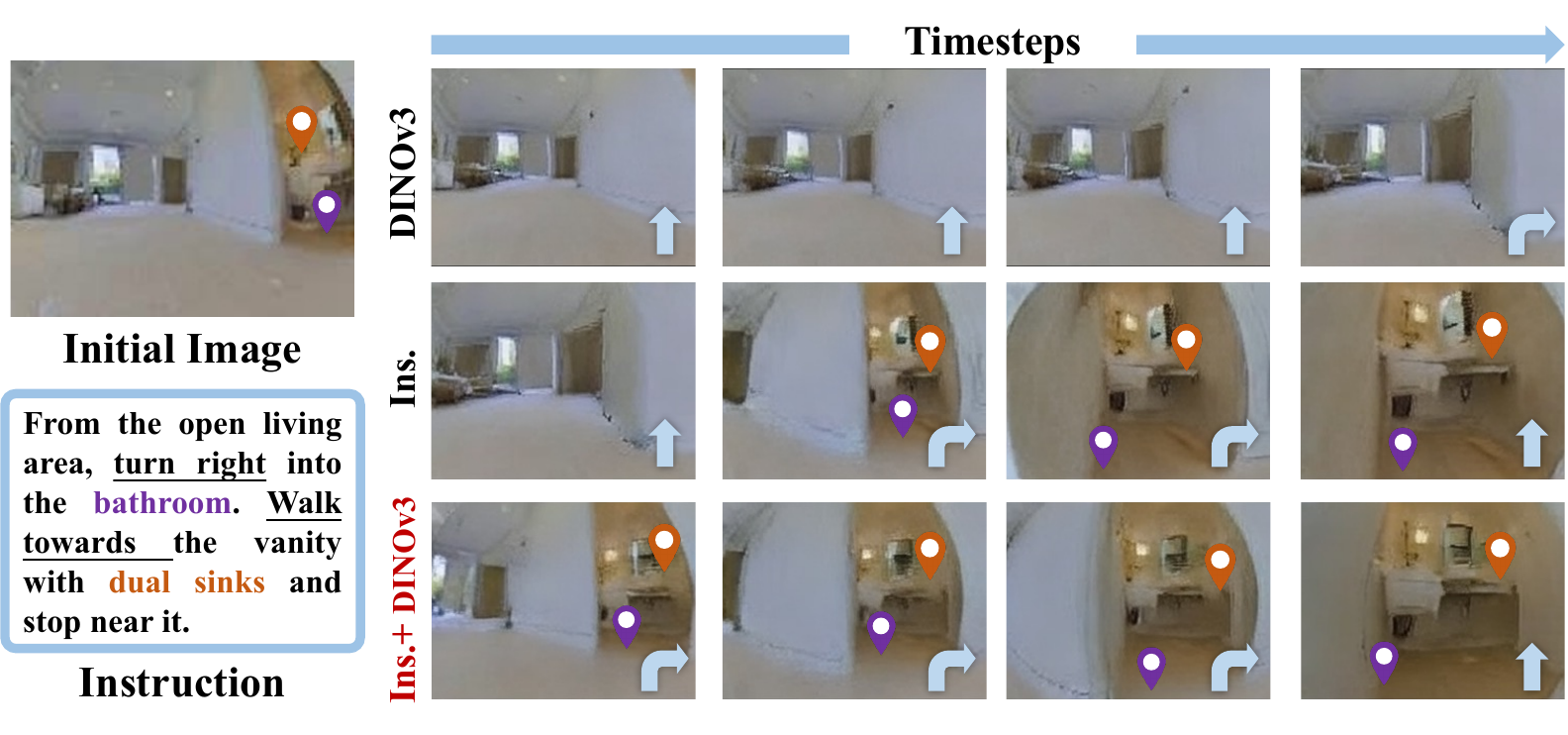}
    \par
\end{minipage}
\hfill
\begin{minipage}[b]{0.42\textwidth}
\vspace{0pt}
    \centering
    \begingroup
    \pgfkeys{/pgf/number format/assume math mode=true}

    \resizebox{0.9\linewidth}{!}{%
    \begin{tikzpicture}[
        x=1.65cm,
        y=5.2cm,
        text=TextGray,
        ticklabel/.style={
            font=\chartfont\fontsize{9}{9}\selectfont\bfseries
        },
        metriclabel/.style={
            font=\chartfont\fontsize{9}{9}\selectfont\bfseries
        },
        valuelabel/.style={
            font=\chartfont\tiny\bfseries,
            inner sep=0.5pt,
            anchor=south
        },
        legendlabel/.style={
            font=\chartfont\fontsize{9}{9}\selectfont\bfseries,
            anchor=west,
            inner sep=2pt
        }
    ]

    \def\BarHalf{0.14}
    \def\StyleOffset{0.28}
    \def\PlotLeft{0.10}
    \def\PlotRight{3.80}

    \foreach \v in {0,0.2,0.4,0.6,0.8}{
        \draw[draw=GridGray,line width=0.3pt]
            (\PlotLeft,\v) -- (\PlotRight,\v);

        \node[
            ticklabel,
            anchor=east,
            inner sep=2pt
        ] at (\PlotLeft,\v) {%
            \pgfmathprintnumber[
                fixed,
                precision=1,
                fixed zerofill
            ]{\v}%
        };
    }

    \draw[draw=gray!65,line width=0.4pt]
        (\PlotLeft,0.84)
        -- (\PlotLeft,0)
        -- (\PlotRight,0);

    \node[
        font=\chartfont\fontsize{9}{9}\selectfont\bfseries,
        rotate=90,
        anchor=south,
        inner sep=1pt
    ] at (-0.30,0.40) {Metric Value};

    \foreach \ylabel/\xx/\vc/\vl/\vi/\mcolor in {%
        {SR$\uparrow$}/0.60/0.59/0.65/0.56/MetricPurple,
        {nDTW$\uparrow$}/1.50/0.65/0.69/0.64/MetricBlue,
        {ATE$\downarrow$}/2.40/0.72/0.60/0.75/MetricOrange,
        {RPE$\downarrow$}/3.30/0.13/0.10/0.14/MetricRed%
    }{
        \pgfmathsetmacro{\XC}{\xx-\StyleOffset}
        \pgfmathsetmacro{\XI}{\xx+\StyleOffset}

        \fill[fill=\mcolor,fill opacity=0.25]
            ({\XC-\BarHalf},0)
            rectangle ({\XC+\BarHalf},\vc);

        \fill[fill=\mcolor,fill opacity=0.90]
            ({\xx-\BarHalf},0)
            rectangle ({\xx+\BarHalf},\vl);

        \fill[fill=\mcolor,fill opacity=0.60]
            ({\XI-\BarHalf},0)
            rectangle ({\XI+\BarHalf},\vi);

        \node[valuelabel]
            at (\XC,{\vc+0.012}) {%
                \pgfmathprintnumber[
                    fixed,
                    precision=2,
                    fixed zerofill
                ]{\vc}%
            };

        \node[
            valuelabel,
            font=\chartfont\fontsize{5}{5}\selectfont\bfseries
        ] at (\xx,{\vl+0.012}) {%
            \pgfmathprintnumber[
                fixed,
                precision=2,
                fixed zerofill
            ]{\vl}%
        };

        \node[valuelabel]
            at (\XI,{\vi+0.012}) {%
                \pgfmathprintnumber[
                    fixed,
                    precision=2,
                    fixed zerofill
                ]{\vi}%
            };

        \node[
            metriclabel,
            anchor=north,
            inner sep=2pt
        ] at (\xx,-0.025) {\ylabel};
    }

    \fill[white]
        (2.94,0.425) rectangle (3.79,0.815);

    \foreach \name/\alpha/\ly in {%
        Concise/0.25/0.75,
        Landmark/0.90/0.62,
        Intricate/0.60/0.49%
    }{
        \fill[fill=LegendGray,fill opacity=\alpha]
            (3.00,{\ly-0.025})
            rectangle (3.10,{\ly+0.025});

        \node[legendlabel]
            at (3.14,\ly) {\name};
    }

    \end{tikzpicture}%
    }
    \endgroup
    \par
\end{minipage}

\par\nointerlineskip
\begin{minipage}[b]{0.57\textwidth}
    \caption{\textbf{Visual Ablation} on cross-attention conditioning. The specified bathroom and sinks appear only with language.}
    \label{fig:abls}
\end{minipage}
\hfill
\begin{minipage}[b]{0.42\textwidth}
    \caption{Planning performance across three different
    \textbf{instruction styles} on the val seen split.}
    \label{fig:instruction_style}
\end{minipage}

\end{figure*}

Tabs.~\ref{tab:sota_plan} and~\ref{tab:sota_imagine} compare planning and visual prediction on \bench{} across four domains. Tiresias-Agent provides a competitive reference, achieving planning scores of 63 and 51 on val seen and val unseen, respectively. Its MPC variant scores 59 and 46 with the same world model and ranking pipeline, supporting the benefit of learned language-conditioned proposals within this framework. Fig.~\ref{fig-visual_res-all} provides qualitative examples across the four domains, showing imagined task progression such as transferring the sponge and turning at an intersection. Within the policy--world-model paradigm, adding language conditioning to NWM increases its planning scores from 45 to 48 on val seen and from 33 to 35 on val unseen, and also lowers FVD.

Generalization also remains challenging for both planning and imagination: all evaluated agents obtain lower planning scores on val unseen, while all evaluated world models exhibit higher FVD. Together, these results illustrate how \bench{} supports exploring the role of language within planning paradigms, relationship between planning and imagination, and generalization to unseen sources.

\begin{figure*}[t]
\centering
\setlength{\abPlotWidth}{.185\linewidth}
\setlength{\abPlotGap}{.060\linewidth}
\setlength{\abPlotHeight}{25mm} 
\begin{tikzpicture}
\begin{groupplot}[
    abAxis,
    group style={
        group size=4 by 1,
        horizontal sep=\abPlotGap
    },
    tick label style={
        font=\fontsize{7}{8}\selectfont,
        color=abText
    },
    label style={
        font=\fontsize{8}{9}\selectfont,
        color=abText
    },
    xlabel style={
        at={(axis description cs:.5,-.14)},
        anchor=north,
        font=\fontsize{8}{9}\selectfont
    },
    legend style={
        font=\fontsize{6.5}{7.5}\selectfont,
        draw=none,
        fill=none,
        inner sep=1pt,
        row sep=-1pt,
        column sep=1pt
    }
]

\nextgroupplot[
    xmode=log,
    log basis x=2,
    xmin=15,xmax=590,
    xtick={16,32,64,128,256,512},
    xticklabels={16,32,64,128,256,512},
    ymin=.08,ymax=.88,
    ytick={.1,.3,.5,.7},
    xlabel={(a) Pool size $N$},
    ylabel={SR $\uparrow$},
    legend columns=1,
    legend style={
        at={(.98,.03)},
        anchor=south east
    }
]

\addplot[
    name path=abBottom,
    draw=none,
    forget plot
] coordinates {
    (16,.13)(512,.13)
};

\addplot[
    name path=abTop,
    draw=none,
    forget plot
] coordinates {
    (16,.56)(32,.56)(64,.58)
    (128,.60)(256,.61)(512,.61)
};

\addplot[
    abOurs,
    fill opacity=.12,
    draw=none,
    forget plot
] fill between[of=abBottom and abTop];

\addplot[abRandom] coordinates {
    (15,.13)(590,.13)
};
\addlegendentry{Random}

\addplot[abOurs,mark=*] coordinates {
    (16,.56)(32,.56)(64,.58)
    (128,.60)(256,.61)(512,.61)
};
\addlegendentry{Tiresias (Tr+V)}

\addplot[abTr,mark=triangle*] coordinates {
    (16,.508)(32,.516)(64,.524)
    (128,.530)(256,.534)(512,.535)
};
\addlegendentry{Tr Only}

\addplot[abGray,dashed] coordinates {
    (16,.58)(32,.65)(64,.71)
    (128,.78)(256,.82)(512,.83)
};
\addlegendentry{Oracle-of-$N$}

\nextgroupplot[
    xmode=log,
    log basis x=2,
    xmin=.8,xmax=82,
    xtick={1,4,8,16,32,64},
    xticklabels={1,4,8,16,32,64},
    ymin=48,ymax=65,
    ytick={50,55,60,65},
    xlabel={(b) Shortlist size $k$},
    ylabel={SR (\%) $\uparrow$},
    legend columns=1,
    legend style={
        at={(.98,-.03)},
        anchor=south east
    }
]

\addplot[
    name path=ab16Lo,
    draw=none,
    forget plot
] coordinates {
    (1,48.6000)(4,52.9000)(8,52.8000)
};

\addplot[
    name path=ab16Hi,
    draw=none,
    forget plot
] coordinates {
    (1,53.0000)(4,56.1000)(8,59.2000)
};

\addplot[
    abGray,
    fill opacity=.12,
    draw=none,
    forget plot
] fill between[of=ab16Lo and ab16Hi];

\addplot[abGray,mark=diamond*] coordinates {
    (1,50.8000)(4,54.5000)(8,56.0000)
};
\addlegendentry{$N=16$}

\addplot[
    name path=ab32Lo,
    draw=none,
    forget plot
] coordinates {
    (1,51.4000)(4,54.7000)
    (8,55.4000)(16,55.8000)
};

\addplot[
    name path=ab32Hi,
    draw=none,
    forget plot
] coordinates {
    (1,51.8000)(4,55.7000)
    (8,57.6000)(16,58.2000)
};

\addplot[
    abBlue,
    fill opacity=.12,
    draw=none,
    forget plot
] fill between[of=ab32Lo and ab32Hi];

\addplot[abBlue,mark=*] coordinates {
    (1,51.6000)(4,55.2000)
    (8,56.5000)(16,57.0000)
};
\addlegendentry{$N=32$}

\addplot[
    name path=ab64Lo,
    draw=none,
    forget plot
] coordinates {
    (1,51.9000)(4,55.1000)(8,56.1000)
    (16,57.4000)(32,57.6000)
};

\addplot[
    name path=ab64Hi,
    draw=none,
    forget plot
] coordinates {
    (1,52.9000)(4,56.1000)(8,59.5000)
    (16,59.4000)(32,60.0000)
};

\addplot[
    abOrange,
    fill opacity=.12,
    draw=none,
    forget plot
] fill between[of=ab64Lo and ab64Hi];

\addplot[abOrange,mark=triangle*] coordinates {
    (1,52.4000)(4,55.6000)(8,57.8000)
    (16,58.4000)(32,58.8000)
};
\addlegendentry{$N=64$}

\addplot[
    name path=ab128Lo,
    draw=none,
    forget plot
] coordinates {
    (1,51.0000)(4,56.1000)(8,58.9000)
    (16,59.8000)(32,58.0000)(64,58.1000)
};

\addplot[
    name path=ab128Hi,
    draw=none,
    forget plot
] coordinates {
    (1,55.0000)(4,59.9000)(8,61.5000)
    (16,61.8000)(32,63.2000)(64,62.5000)
};

\addplot[
    abPurple,
    fill opacity=.12,
    draw=none,
    forget plot
] fill between[of=ab128Lo and ab128Hi];

\addplot[abPurple,mark=square*] coordinates {
    (1,53.0000)(4,58.0000)(8,60.2000)
    (16,60.8000)(32,60.6000)(64,60.3000)
};
\addlegendentry{$N=128$}

\nextgroupplot[
    xmin=-.55,xmax=4.55,
    ymin=.20,ymax=.78,
    ytick={.2,.3,.4,.5,.6,.7},
    xtick={0,1,2,3,4},
    xticklabel=\empty,
xticklabel style={opacity=0},
    xlabel={(c) $\lambda_{\mathrm{tr}}:\lambda_{\mathrm{v}}$},
    ylabel={Metric value},
    legend columns=2,
    legend style={
        at={(.02,.98)},
        anchor=north west
    }
]

\path[fill=abHighlight]
    (axis cs:.53,.20) rectangle (axis cs:1.47,.78);

\addplot[
    ybar,
    area legend,
    bar width=6.1pt,
    bar shift=0pt,
    fill=abBlue,
    fill opacity=.8,
    draw=white,
    line width=.25pt,
    abError
] coordinates {
    (-.17,.5300) +- (0,.0200)
    (.83,.6020)  +- (0,.0130)
    (1.83,.5700) +- (0,.0180)
    (2.83,.5200) +- (0,.0200)
    (3.83,.4700) +- (0,.0110)
};
\addlegendentry{SR $\uparrow$}

\addplot[
    ybar,
    area legend,
    bar width=6.1pt,
    bar shift=0pt,
    fill=abOurs,
    fill opacity=.8,
    draw=white,
    line width=.25pt,
    abError
] coordinates {
    (.17,.6400)  +- (0,.0040)
    (1.17,.6580) +- (0,.0040)
    (2.17,.6500) +- (0,.0040)
    (3.17,.6400) +- (0,.0050)
    (4.17,.6100) +- (0,.0060)
};
\addlegendentry{nDTW $\uparrow$}

\coordinate (abFusionTickA) at (axis cs:0,.20);
\coordinate (abFusionTickB) at (axis cs:1,.20);
\coordinate (abFusionTickC) at (axis cs:2,.20);
\coordinate (abFusionTickD) at (axis cs:3,.20);
\coordinate (abFusionTickE) at (axis cs:4,.20);

\nextgroupplot[
    xmin=-.55,xmax=3.55,
    ymin=.20,ymax=.78,
    ytick={.2,.3,.4,.5,.6,.7},
    xtick={0,1,2,3},
    xticklabel=\empty,
xticklabel style={opacity=0},
    xlabel={(d) Selector / Imag. Budget},
    ylabel={Metric value},
    xlabel style={
        at={(axis description cs:.5,-.14)},
        anchor=north,
        font=\fontsize{8}{9}\selectfont
    },
    legend columns=2,
    legend style={
        at={(.02,.98)},
        anchor=north west
    }
]

\path[fill=abHighlight]
    (axis cs:2.53,.20) rectangle (axis cs:3.47,.78);

\addplot[
    ybar,
    area legend,
    bar width=7.7pt,
    bar shift=0pt,
    fill=abBlue,
    fill opacity=.8,
    draw=white,
    line width=.25pt,
    abError
] coordinates {
    (-.17,.2700) +- (0,.0250)
    (.83,.3900)  +- (0,.0160)
    (1.83,.4600) +- (0,.0130)
    (2.83,.6020) +- (0,.0130)
};
\addlegendentry{SR $\uparrow$}

\addplot[
    ybar,
    area legend,
    bar width=7.7pt,
    bar shift=0pt,
    fill=abOurs,
    fill opacity=.8,
    draw=white,
    line width=.25pt,
    abError
] coordinates {
    (.17,.5500)  +- (0,.0020)
    (1.17,.5900) +- (0,.0030)
    (2.17,.6200) +- (0,.0020)
    (3.17,.6580) +- (0,.0040)
};
\addlegendentry{nDTW $\uparrow$}

\coordinate (abSelectorTickA) at (axis cs:0,.20);
\coordinate (abSelectorTickB) at (axis cs:1,.20);
\coordinate (abSelectorTickC) at (axis cs:2,.20);
\coordinate (abSelectorTickD) at (axis cs:3,.20);

\end{groupplot}

\begin{scope}[
    every node/.style={
        anchor=north,
        inner sep=0pt,
        yshift=-4pt,
        text=abText,
        font=\fontsize{7}{8}\selectfont
    }
]
    \node at (abFusionTickA) {$1{:}0$};
    \node[text=abPurple] at (abFusionTickB)
        {$\mathbf{2{:}1}$};
    \node at (abFusionTickC) {$1{:}1$};
    \node at (abFusionTickD) {$1{:}2$};
    \node at (abFusionTickE) {$0{:}1$};
\end{scope}

\begin{scope}[
    every node/.style={
        anchor=north,
        inner sep=0pt,
        yshift=-3pt,
        text=abText,
        font=\fontsize{6}{7}\selectfont
    }
]
    \node at (abSelectorTickA)
        {$\mathrm{R}\mathord{\to}4$};
    \node at (abSelectorTickB)
        {$\mathrm{R}\mathord{\to}8$};
    \node at (abSelectorTickC)
        {$\mathrm{R}\mathord{\to}16$};
    \node[text=abPurple] at (abSelectorTickD)
        {$\mathbf{Tr}\mathord{\to}\mathbf{8}$};
\end{scope}

\end{tikzpicture}
\caption{\textbf{Ablations on candidate sampling and selection} on val seen. \textbf{(a)}~Test-time scaling with pool size $N$ at fixed $k{=}8$; Oracle-of-$N$ provides the SR upper bound over each candidate pool. \textbf{(b)}~Effect of shortlist size $k$ for different pool sizes. \textbf{(c)}~Ranker fusion weights $\lambda_{\mathrm{tr}}:\lambda_{\mathrm{v}}$ under the default $128 \rightarrow 8$ funnel. \textbf{(d)}~Random versus Tr-Ranker shortlisting with different imagination budgets at $N{=}128$. Shaded bands and error bars in (b--d) indicate $\pm$ one standard deviation. R denotes random selection.}
\label{fig:abls_data}
\end{figure*}
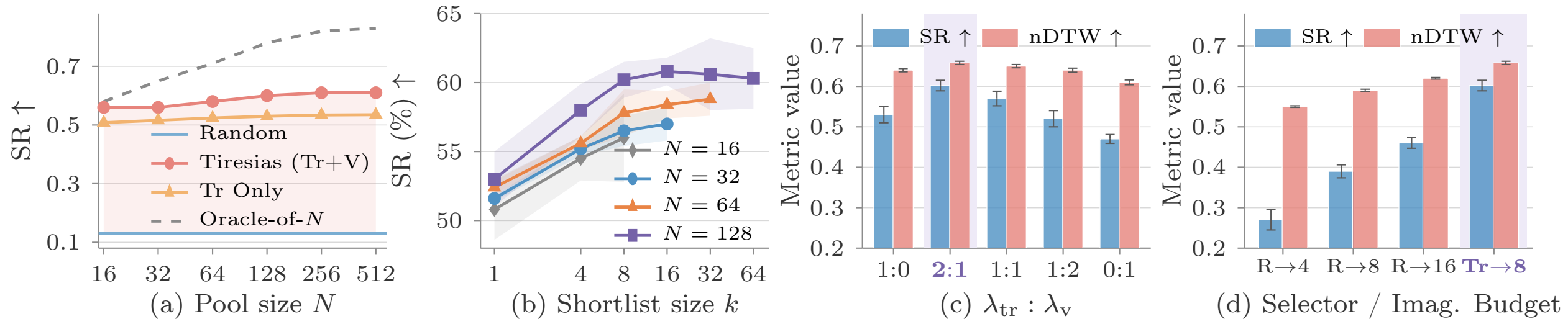

\vspace{-0.1cm}
\subsection{Ablation and Analysis}
\vspace{-0.1cm}
\noindent \textbf{What role does language play in planning and imagination?}~
Tab.~\ref{tab:abl_cond_scaled} varies the world model's cross-attention inputs while retaining the initial-frame VAE latent and domain prompt. Combining instructions with DINOv3 features performs best, and removing instructions causes the larger drop, highlighting the role of language in guiding task-relevant visual predictions. Fig.~\ref{fig:abls} illustrates this effect: the specified bathroom and sinks appear only in language-conditioned rollouts. For LDP, language alone yields the best results both before and after ranking (Tab.~\ref{tab:abl_ldp_cond}). These support our design of using language-only conditioning for trajectory proposals and combining language with scene features in world model.

\noindent \textbf{Which instruction style best supports planning?}~
We compare concise, landmark-based, and intricate instructions in Fig.~\ref{fig:instruction_style}. Landmark-based instructions perform best, while intricate descriptions underperform concise ones despite providing more detail. This suggests that language guidance benefits more from clear references linking motions to observable targets than from additional descriptive content.



\begin{table}[!t]
\setlength{\abovecaptionskip}{2pt}
\setlength{\belowcaptionskip}{0pt}
\centering
        \setlength{\tabcolsep}{6pt}
        \tiny
        \captionof{table}{
            \textbf{Selector comparison} on val seen using the same
            eight randomly sampled LDP candidates. Imag.\ denotes
            methods requiring world-model rollouts. Best results are in bold. 
        }
        \renewcommand{\arraystretch}{1.1}

        \resizebox{0.8\linewidth}{!}{%
        \begin{tabular}{lIcIIcccc}
            \hline\thickhline
            \rowcolor{gray!20}\textbf{Selector}
            & \textbf{Imag.}
            & SR$\uparrow$
            & nDTW$\uparrow$
            & ATE$\downarrow$
            & RPE$\downarrow$ \\
            \hline\hline

            Random
            & $\times$
            & 0.13 & 0.34 & 2.35 & 0.29 \\

            Consensus (mean traj.)
            & $\times$
            & 0.24 & 0.47 & 1.43 & 0.20 \\

            Cluster-vote (medoid)
            & $\times$
            & 0.18 & 0.40 & 1.91 & 0.25 \\

            Tr-Ranker + Direct
            & $\times$
            & 0.31 & 0.54 & 1.10 & 0.17 \\

            Qwen3.6 verifier
            & \checkmark
            & 0.26 & 0.49 & 1.28 & 0.19 \\

            \rowcolor{myblue!40} Tr-Ranker
            & $\times$
            & 0.30 & 0.53 & 1.14 & 0.17 \\

            \rowcolor{myblue!50} V-Ranker
            & \checkmark
            & 0.28 & 0.51 & 1.20 & 0.19 \\

            \rowcolor{myblue!80} \textbf{Tr + V (fused)}
            & \checkmark
            & \textbf{0.39}
            & \textbf{0.59}
            & \textbf{0.98}
            & \textbf{0.14} \\
\hdashline

            \textcolor{gray}{Oracle-of-8}
            & \textcolor{gray}{--}
            & \textcolor{gray}{0.52}
            & \textcolor{gray}{0.66}
            & \textcolor{gray}{0.85}
            & \textcolor{gray}{0.12} \\
\hline
            
        \end{tabular}%
        }

        \label{tab:abl_rank_scaled}
\end{table}
\begin{table}[!t]
\setlength{\abovecaptionskip}{2pt}
\setlength{\belowcaptionskip}{0pt}
\centering
\tiny
\setlength{\tabcolsep}{12pt}
\renewcommand{\arraystretch}{1.1}
\captionof{table}{
\textbf{Stage-one shortlist quality} on val seen ($N{=}128$).
Retention is ratio of shortlist oracle SR to full-pool oracle SR.
Hit@$k$ is percentage of episodes whose oracle-best candidate
is retained.
}
\resizebox{0.8\linewidth}{!}{%
\begin{tabular}{lIIcccc}
\hline\thickhline
\rowcolor{gray!20}\textbf{Metric}
& $k{=}1$
& $k{=}4$
& $k{=}8$
& $k{=}16$ \\
\hline\hline

Retention$\uparrow$
& 0.68
& 0.80
&\cellcolor{myblue!80} \textbf{0.85}
& 0.86 \\

Final SR$_{\text{Tr}}$$\uparrow$
& 0.53
& 0.58
& \cellcolor{myblue!80} \textbf{0.60}
& 0.61 \\

Final SR$_{\text{Random}}$$\uparrow$
& 0.13
& 0.27
& \cellcolor{myblue!80} \textbf{0.39}
& 0.46 \\

Hit@$k$ (\%)$\uparrow$
& 5.39
& 14.77
& \cellcolor{myblue!80} \textbf{22.34}
& 34.22 \\

\hline
\end{tabular}}
\label{tab:shortlist_oracle}
\vspace{-0.1cm}
\end{table}

\noindent \textbf{How does imagination contribute to plan selection?}~Tab.~\ref{tab:abl_rank_scaled} compares selectors on the same eight randomly sampled LDP candidates. Both learned rankers outperform the consensus heuristics, while score fusion raises SR from 0.30 for the Tr-Ranker and 0.28 for the V-Ranker to 0.39, demonstrating their complementary value. Adding a scene-aware direct scorer that skips imagination yields only
0.31 SR, suggesting that the gain stems from imagined outcomes rather
than scene access alone. We also compare against a Qwen3.6 verifier that applies MVoT-inspired reasoning~\citep{li2025imagine} to the same imagined rollouts. The V-Ranker performs better on SR, nDTW, and ATE, with the same RPE. These results support the role of imagination in plan selection and highlight the importance of how imagined outcomes are scored.



\noindent \textbf{How much imagination does planning need?}~
We vary the proposal pool and shortlist sizes in Figs.~\ref{fig:abls_data}a and b. With $k{=}8$ fixed, final SR largely plateaus around $N{=}128$ despite continued gains in Oracle-of-$N$, showing that broader candidate coverage does not automatically improve the selected plan. Enlarging the shortlist likewise yields diminishing returns. Tab.~\ref{tab:shortlist_oracle} shows that a shortlist of 8 out of 128 candidates preserves 85\% of the full-pool oracle SR, yet contains the oracle-best candidate in only 22.34\% of episodes. This indicates that the shortlist often preserves successful alternatives even when it misses the best candidate. Consistent with this, Tr-Ranker shortlisting outperforms random shortlisting (Fig.~\ref{fig:abls_data}d). Within the shortlist, a $2{:}1$ trajectory-to-visual score ratio performs best (Fig.~\ref{fig:abls_data}c), balancing motion agreement with visual evidence. We therefore adopt the $128 \rightarrow 8$ funnel with $2{:}1$ score fusion.



\noindent \textbf{Do the rankers generalize beyond LDP proposals?}~
We replace the LDP with NoMaD (lang) or sampling-based MPC, keeping the $128\rightarrow8$ pipeline and reusing both rankers without further fine-tuning on val seen. Both alternatives benefit from trajectory ranking and improve further with visual verification (Tab.~\ref{tab:plug_play}), demonstrating the applicability of our selection pipeline beyond LDP proposals.

\begin{table}[!t]
\setlength{\abovecaptionskip}{1pt}
\setlength{\belowcaptionskip}{0pt}
\centering
\tiny
\setlength{\tabcolsep}{10pt}
\renewcommand{\arraystretch}{1.1}
\captionof{table}{
\textbf{Plug-and-play trajectory ranking}.~The trained rankers are applied to NoMaD (lang) and MPC without further fine-tuning. Best results are in bold.
}
\resizebox{0.8\linewidth}{!}{%
\begin{tabular}{lIlIIcccc}
\hline\thickhline
\rowcolor{gray!20}\textbf{Policy} & \textbf{Selector} & SR$\uparrow$ & nDTW$\uparrow$ & ATE$\downarrow$ & RPE$\downarrow$ \\
\hline\hline
\multirow{3}{*}{MPC}
 & Unranked    & 0.11 & 0.32 & 2.62 & 0.31 \\ 
 & Tr-Ranker      & 0.49 & 0.60 & 0.92 & 0.20 \\ 
 & Tr + V  & 0.56 & 0.62 & 0.79 & 0.17 \\ 
\hline
\multirow{3}{*}{NoMaD (lang)}
 & Unranked    & 0.32 & 0.48 & 1.46 & 0.32 \\ 
 & Tr-Ranker      & 0.42 & 0.58 & 1.05 & 0.21 \\ 
 & Tr + V  & 0.47 & 0.60 & 0.92 & 0.18 \\ 
\hline
\multirow{3}{*}{LDP (ours)}
& Unranked    & 0.13 & 0.34 & 2.35 & 0.29 \\ 
  & Tr-Ranker      & 0.53 & 0.64 & 0.77 & 0.13 \\ 
  & Tr + V \cellcolor{myblue!80}  & \textbf{0.60} \cellcolor{myblue!80} & \textbf{0.66}\cellcolor{myblue!80}  & \textbf{0.69} \cellcolor{myblue!80} & \textbf{0.12}\cellcolor{myblue!80}  \\ 
\hline
\end{tabular}}
\label{tab:plug_play}
\end{table}


\section{Conclusion}

We introduce a task and benchmark for language-conditioned world modeling. \bench{} unifies 15 datasets across four domains under a shared action interface and evaluation protocol, with three instruction styles. We further propose Tiresias-Agent, which combines language-conditioned trajectory proposals with selective imagination and achieves competitive planning and imagination performance across the evaluated domains. Ablations support the complementary roles of trajectory and visual ranking, reveal the trade-off between candidate coverage and imagination cost, and show that the trained rankers also benefit alternative candidate generators. Together, \bench{} and Tiresias-Agent lay the groundwork for broader studies of language-guided planning, with execution-based validation in simulators and on physical robots as a natural direction for future work.

\section*{In Memoriam}

We dedicate this work to the memory of Professor Alexander G.~Hauptmann,
a cherished mentor and friend. Alex mentored Zhi-Qi Cheng throughout
his seven years at Carnegie Mellon University. After Zhi-Qi joined the
University of Washington, Alex continued to support him, his
students, and his growing team. Even near the end of his life, Alex
remained engaged in our research on world models and continued to
encourage us.

Alex was deeply curious about how people perceive the world. When
Zhi-Qi told him about this research, Alex asked whether world models
might help us understand human perception. That question has stayed
with us. We miss his kindness, his curiosity, and the trust he placed
in his students. We hope to honor his memory through the questions we
pursue and the students we mentor.

\section*{Acknowledgments}
This work was supported in part by faculty startup funding from the University of Washington and by the U.S. Department of Transportation through the Pacific Northwest Transportation Consortium (PacTrans), including PacTrans seed funding and the Small Research Projects program. We also gratefully acknowledge graduate student awards from the Vicky L. Carwein and William B. Andrews Endowment for Graduate Programs and the University of Washington Graduate School Top Scholar program. Zhi-Qi Cheng acknowledges faculty development support through the Undergraduate Research Faculty Fellows program, supported by the Henry Luce Foundation and the campus Office of Academic Affairs, and through Engineering Unleashed and the Kern Entrepreneurial Engineering Network (KEEN), generously supported by The Kern Family Foundation.

\newpage

\appendix
\section*{Appendix}
\addcontentsline{toc}{section}{Appendix}  

\setcounter{table}{0}  
\setcounter{figure}{0}

\setcounter{equation}{0}
\renewcommand{\thetable}{A\arabic{table}}
\renewcommand{\thefigure}{A\arabic{figure}}
\renewcommand{\thelisting}{A\arabic{listing}}

\renewcommand{\theequation}{A\arabic{equation}}

\noindent This supplementary material provides comprehensive technical documentation, extended dataset analyses, and additional empirical results to support the main text. This supplementary material is organized into three primary sections: 
First, Sec.~\ref{appx:benchmark} details the construction of the \bench{} benchmark, encompassing trajectory preprocessing, the automated annotation pipeline, human evaluation protocols, and extended dataset statistics. 
Second, Sec.~\ref{appx:method} formalizes the Tiresias-Agent, providing more details for the constituent modules (LDP, World Model, Tr-Ranker, and V-Ranker), alongside domain-specific system prompts and inference pseudo-code. 
Finally, Sec.~\ref{appx:experiments} outlines the extended experimental setups, including rigorous definitions of the evaluation metrics, training implementation details, and more experimental results.

\section{\bench{} Details}
\label{appx:benchmark}
\subsection{Comparison with Existing Benchmarks}
\begin{table*}[h]
\setlength{\abovecaptionskip}{2pt}
\setlength{\belowcaptionskip}{0pt}
\centering

\caption{\textbf{Comparison of world-model and related policy benchmarks.}
Our benchmark combines four control domains, language-guided planning
and visual evaluation, a shared action interface, and three instruction styles.}
\label{tab:benchmark_comparison}

\begingroup
\tiny
\setlength{\tabcolsep}{2pt}
\renewcommand{\arraystretch}{1.2}
\providecommand{\bcy}{\ensuremath{\checkmark}}
\providecommand{\bcn}{\ensuremath{\times}}

\resizebox{\textwidth}{!}{%
\begin{tabular}{l||cccc|c|cccc}
\thickhline
\rowcolor{gray!20}
& \multicolumn{4}{c|}{\textbf{Domain coverage}}
& & \multicolumn{4}{c}{\textbf{Evaluation / interface}} \\
\cline{2-5}\cline{7-10}
\rowcolor{gray!20}
\multirow{-2}{*}{\textbf{Benchmark}}
& Nav. & Rob. & Game & Drive
& \multirow{-2}{*}{\textbf{Reported scale}}
& Lang. plan. & Visual & Shared action & Multi style \\
\hline\hline

World Stability~\citep{kwon2025stableworldmodelsmeasuring}
& \bcn & \bcn & \bcy & \bcn
& 2 environments
& \bcn & \bcy & \bcn & \bcn \\

LoopNav~\citep{lian2025toward}
& \bcn & \bcn & \bcy & \bcn
& 250 h / 20M frames
& \bcn & \bcy & \bcn & \bcn \\

DrivingDojo~\citep{wang2024drivingdojo}
& \bcn & \bcn & \bcn & \bcy
& 17.8K clips
& \bcn & \bcy & \bcn & \bcn \\

EWMBench~\citep{yue2025ewmbench}
& \bcn & \bcy & \bcn & \bcn
& 100 eval. episodes
& \bcn & \bcy & \bcn & \bcn \\

WoW-World-Eval~\citep{fan2026wow}
& \bcn & \bcy & \bcn & \bcn
& 609 eval. samples
& \bcy & \bcy & \bcn & \bcy \\

Omni-WorldBench~\citep{wu2026omniworldbenchcomprehensiveinteractioncentricevaluation}
& \bcn & \bcy & \bcy & \bcy
& 1,068 eval. prompts
& \bcn & \bcy & \bcn & \bcn \\


World-in-World~\citep{zhang2025world}
& \bcy & \bcy & \bcn & \bcn
& 1,079 eval. instances
& \bcy & \bcy & \bcy & \bcn \\

WorldArena~\citep{shang2026worldarena}
& \bcn & \bcy & \bcn & \bcn
& 2,500 videos / 50 tasks
& \bcy & \bcy & \bcn & \bcn \\

CALVIN~\citep{mees2022calvinbenchmarklanguageconditionedpolicy}
& \bcn & \bcy & \bcn & \bcn
& 24 h / 1,000 eval. chains
& \bcy & \bcn & \bcn & \bcn \\

LIBERO~\citep{liu2023libero}
& \bcn & \bcy & \bcn & \bcn
& 130 tasks
& \bcy & \bcn & \bcn & \bcn \\

LCVN~\citep{dong2026language}
& \bcy & \bcn & \bcn & \bcn
& 39,016 traj. / 117,048 pairs
& \bcy & \bcy & \bcn & \bcy \\

\rowcolor{myblue!80}
\textbf{\bench{} (Ours)}
& \bcy & \bcy & \bcy & \bcy
& \textbf{142,987 episodes / 428,961 pairs}
& \bcy & \bcy & \bcy & \bcy \\
\hline
\end{tabular}%
}

\par\vspace{2pt}
\noindent
\begin{minipage}{\textwidth}
\tiny
\setlength{\parindent}{0pt}
\textbf{Definitions.}
\bcy/\bcn: included/not included in the reported protocol.
Nav.: physical-world navigation; Rob.: manipulation;
Game: 3D gaming (including Minecraft and DMLab).
\textit{Lang. plan.}: language-guided action planning/control evaluation,
including video-to-action execution.
\textit{Visual}: generated-observation quality, consistency,
or plausibility evaluation.
\textit{Shared action}: a common action interface across multiple domains.
\textit{Multi style}: explicit instruction-style comparison
(two styles in WoW-World-Eval; three in LCVN and ours).
\textbf{Scale.}
Units and corpus/evaluation scope differ; counts are not directly comparable.
DrivingDojo follows the published paper
(17.8K; project page: 18.2K).
World-in-World sums 551 recognition, 144 navigation,
184 QA, and 200 manipulation instances.
WorldArena includes 500 held-out videos.
Our 15-source corpus spans all splits;
pairs denote episode--instruction pairs.
\end{minipage}
\par
\endgroup
\end{table*}

Table~\ref{tab:benchmark_comparison} compares existing benchmarks in terms of domain coverage, data scale, and evaluation scope. World Stability~\citep{kwon2025stableworldmodelsmeasuring}, LoopNav~\citep{lian2025toward}, and DrivingDojo~\citep{wang2024drivingdojo} primarily assess visual consistency or action-conditioned prediction, whereas CALVIN~\citep{mees2022calvinbenchmarklanguageconditionedpolicy} and LIBERO~\citep{liu2023libero} focus on language-conditioned manipulation policies. Recent suites, including World-in-World~\citep{zhang2025world} and WorldArena~\citep{shang2026worldarena}, additionally evaluate functional utility through planning or downstream control. WoW-World-Eval~\citep{fan2026wow} and LCVN~\citep{dong2026language} also examine different instruction styles. These capabilities are represented individually across prior benchmarks, but their domain coverage and evaluation interfaces differ. \bench{} brings them together across navigation, robotics, 3D gaming, and driving under a shared action interface and evaluation protocol. It integrates 15 datasets into 142,987 episodes and 428,961 episode--instruction pairs, supporting both language-guided planning and visual prediction with three instruction styles per episode. Since existing works report different units and corpus scopes, the scale column provides context rather than a size ranking.

\subsection{Trajectory Sources and Preprocessing} 

Navigation spans five ground-robot datasets covering indoor corridors, unstructured open-world scenes, human-populated social spaces, and off-road terrains. Across all datasets, the navigation action $(u_t,\phi_t)$ is derived from consecutive relative poses, with displacements normalized by the per-source average step size.

\noindent \textbf{Go Stanford}~\citep{hirose2018gonet}.~This dataset comprises approximately 24 hours of indoor video (3\,Hz) collected by a teleoperated TurtleBot2 equipped with a Ricoh THETA S fisheye camera across 25 buildings. The trajectories primarily feature corridors, doorways, and staircases, providing tight, structurally repetitive indoor layouts. The forward-facing view is rectified to a perspective crop, resized to $256\times256$, and combined with wheel odometry to compute per-step displacements.

\noindent \textbf{RECON}~\citep{shah2021rapid}.~This dataset contains over 5{,}000 autonomous trajectories collected by a Clearpath Jackal via time-correlated random walks. Recorded over 18 months across nine outdoor environments (e.g., parking lots, farmlands, residential areas), the data includes RGB observations captured by a forward-facing fisheye camera with a $170^\circ$ FOV.

\noindent \textbf{SCAND}~\citep{karnan2022socially}.~SCAND provides 8.7 hours (40\,km) of human-teleoperated, socially compliant demonstrations recorded indoors and outdoors. It features two morphologically distinct platforms: a wheeled Clearpath Jackal and a legged Boston Dynamics Spot. The trajectories capture naturally occurring social interactions, such as street crossing and passing pedestrians. Actions are derived from wheel odometry, paired with the front-facing RGB stream.

\noindent \textbf{HuRoN}~\citep{hirose2023sacson}.~HuRoN comprises 75 hours (58.7\,km) of robot navigation across five office environments, capturing over 4{,}000 human-robot interactions. Unlike SCAND, HuRoN's autonomous data collection policy actively approaches pedestrians, resulting in episodes dense with dynamic obstacles and frequent instances of yielding behavior.

\noindent \textbf{TartanDrive}~\citep{triest2022tartandrive}.~This dataset includes approximately 200{,}000 off-road interactions (5 hours at 10\,Hz) recorded by a Yamaha Viking ATV traversing unstructured terrains (e.g., rocks, mud, steep slopes). Lacking indoor or urban structures and featuring visible robot parts in the camera view, it presents significant visual and dynamic disparities compared to other sources. Thus, TartanDrive is strictly excluded from training and reserved for val unseen and test splits to evaluate generalization under severe domain shifts.

The robotics domain involves 6-DoF gripper manipulation and constitutes the largest subset of \bench{} ($40\%$), encompassing household kitchens, offices, tabletop lab setups, and a procedurally generated simulator. Because all four datasets record absolute end-effector poses rather than displacement commands, each episode is converted into per-step increments within the local end-effector frame, defining a 7-dimensional action space $(u_t, z_t, \phi_t, \rho_t, \vartheta_t, g_t)$. Translations are normalized by the per-source average step size. The gripper state is maintained as a binary open/close command and explicitly verified at the terminal step.

\noindent \textbf{DROID}~\citep{khazatsky2024droid}.~As the largest robotics dataset in our split, DROID contains 76k demonstration trajectories (350 hours) collected across 564 scenes, 52 buildings, and 86 tasks. Demonstrations used identical Franka Panda hardware in environments ranging from research labs to real households.

\noindent \textbf{BridgeData~V2}~\citep{walke2023bridgedata}.~This dataset comprises 60{,}096 trajectories collected using a WidowX arm across 24 environments. It features 50{,}365 human demonstrations spanning 13 skills (e.g., pick-and-place, pushing, folding) and over 100 distinct objects. Scenes are designed to support multiple feasible tasks simultaneously, providing diverse instruction conditioning from identical initial observations.

\noindent \textbf{RT-1}~\citep{brohan2022rt}.~The RT-1 corpus includes approximately 130k episodes collected by 13 mobile manipulators operating in office kitchen environments. The data covers over 700 manipulation tasks, including picking, placing, drawer articulation, and object rearrangement.

\noindent \textbf{LIBERO}~\citep{liu2023libero}.~As the sole simulated dataset in the robotics split, LIBERO provides 130 procedurally generated manipulation tasks categorized into four suites, varying by object types, spatial layouts, and task goals. Similar to TartanDrive, LIBERO is strictly excluded from training and reserved for the val unseen and test splits to evaluate generalization under visual and structural distribution shifts.

\begin{listing*}[!t]
\caption{Annotation prompt used to generate navigation instructions.}
\label{appx.tab:annot-navigation}

\begin{annotationprompt}[NavigationColor]
\small
\setlength{\parindent}{0pt}
\setlength{\parskip}{2pt}

\promptfield{Role}
You are a visual navigation data annotation assistant.

\promptfield{Inputs}
A sequence of evenly sampled first-person images, in temporal order;
a 2D top-down trajectory map (color from dark to light indicates time;
the red arrow indicates yaw heading); trajectory metadata and an
angle summary (for reference).

\promptfield{Task}
Combine the images and the 2D trajectory geometry to produce three
English navigation instructions. The 2D trajectory map is for judging
direction changes only. Every instruction must end with a sentence
that includes \texttt{stop} (not a standalone ``Stop'').
Do not output your reasoning; output JSON only.

\promptsection{Output styles}

\begin{promptstyle}{concise}{10}
Consists solely of a minimal sequence of action commands centered on
the core movements, direction changes, and stopping location.
A small number of concise positional cues may be included to help
situate an action, but these should serve the path description rather
than expand into descriptions of scenery, objects, or the environment
along the way. A phrase that mainly reports what is visible during
travel, rather than how to move or where to stop, should generally be
omitted unless it is the sole positional information required for a
turn or a stop. The overall style should be short and direct, as clean
as a command list.

\end{promptstyle}

\begin{promptstyle}{intricate}{4}
The most detailed style, as detailed as possible: the sequence of
actions plus extensive visible visual information and scene detail
(objects, people, environmental features, atmosphere, etc.).
\end{promptstyle}

\begin{promptstyle}{landmark}{10}
Organizes the path description around concrete, recognizable
landmarks, stating explicitly which action is performed at which
landmark; landmarks should mark the key nodes of the path.
\end{promptstyle}

\promptsection{Shared constraints}

\begin{promptconstraint}
Do not invent details that are not visible.

Each of the three styles must be no more than 100 words.
\end{promptconstraint}

\promptsection{Reference example}

\textit{One of several examples provided in context.}

\begin{promptexample}{concise}{10}
``Walk straight along the stone-paved walkway and then stop at the end.''
\end{promptexample}

\begin{promptexample}{intricate}{4}
``Proceed along the open, stone-paved walkway, following the blue
tactile paving strip that runs through its center. A black manhole
cover lies on the left, with two pedestrians in dark clothing farther
ahead on that side and another group approaching from the right.
Dense trees line both sides of the walkway, and sunlight filtering
through the canopy casts dappled shadows across the path.
Continue forward and stop near the pedestrians ahead.''
\end{promptexample}

\begin{promptexample}{landmark}{10}
``Follow the blue tactile paving strip along the stone-paved walkway.
Pass the black manhole cover on the left, continue toward the
pedestrians ahead, and stop before reaching them.''
\end{promptexample}

\end{annotationprompt}
\end{listing*}

\begin{listing*}[!t]
\caption{Annotation prompt used to generate robotics instructions.}
\label{appx.tab:annot-robotics}

\begin{annotationprompt}[RoboticsColor]
\small
\setlength{\parindent}{0pt}
\setlength{\parskip}{2pt}

\promptfield{Role}
You are a robot-manipulation visual instruction generation assistant.

\promptfield{Inputs}
A single storyboard image formed by concatenating the robot-arm images
in temporal order (either a wrist-mounted first-person view or an
external third-person view; the model must determine which), plus the
per-frame 7-DoF end-effector action
($\Delta x,\Delta y,\Delta z,\Delta\text{roll},
\Delta\text{pitch},\Delta\text{yaw},\text{gripper}$).
Combine the image sequence and the action sequence to produce three
styles of instruction. Do not output your reasoning; output JSON only.

\promptfield{Analyze the manipulation}
Determine whether the viewpoint is wrist (first-person wrist camera)
or agentview (fixed third-person camera); identify the key objects
visible in the images, their appearance, and their spatial relations;
identify the main phases of the action sequence, but treat the action
sequence as intent/attempt only, not proof of success; pay particular
attention to keyframe roles such as
\texttt{near\_gripper\_close\_attempt} and
\texttt{near\_gripper\_open\_release},
as well as the event windows supplied by
\texttt{event\_metadata}; merge consecutive similar actions and
ignore minor perturbations.

\promptfield{Organize the styles}
Ensure that the three styles are clearly and visibly distinct.

\promptsection{Output styles}

\begin{promptstyle}{concise}{10}
The shortest and most abstract style, retaining only the core sequence
of action verbs (e.g., move forward, grasp, lift, place, rotate, or
close drawer), like a pure command list. If evidence of success is
insufficient, write attempt, not confirmed, or failed rather than
completing it into a successful sequence.
\end{promptstyle}

\begin{promptstyle}{intricate}{4}
A vivid narrative style describing object appearance (color, material,
and shape), the gripper's motion trajectory and state changes, the
surrounding workspace, and the visual feedback during execution.
It should be clearly richer and more vivid than landmark, like
narrating a visualizable manipulation story. It must faithfully
reflect whether the manipulation succeeded, explicitly stating
uncertainty or failure when evidence is insufficient.
\end{promptstyle}

\begin{promptstyle}{landmark}{10}
Structures the whole action sequence around the clearly visible,
recognizable key objects in the image as spatial and manipulation
anchors, with each major step explicitly tied to a specific object
(e.g., ``move to object X and grasp it'' or ``place it beside
object X''). It should be more concise and structured than intricate,
foregrounding objects as reference points. It must not describe a
placement as successful when evidence is insufficient.
\end{promptstyle}

\promptsection{Shared constraints}

\begin{promptconstraint}
Do not invent any details that do not appear in the images.

Each of the three styles must be no more than 100 words.

Every instruction must begin by explicitly stating the viewpoint type:

~~~~~~~ $\cdot$ Wrist viewpoint:
    ``From a wrist-mounted first-person view, \ldots''

~~~~~~~  $\cdot$ Agentview viewpoint:
    ``From an external third-person view, \ldots''

\end{promptconstraint}

\promptsection{Reference example}

\textit{One of several examples provided in context.}

\begin{promptexample}{concise}{10}
``From a wrist-mounted first-person view, pull open the drawer,
grasp the black bowl, and place it inside.''
\end{promptexample}

\begin{promptexample}{intricate}{4}
``From a wrist-mounted first-person view, a white bowl, a black bowl,
a blue box, and a dark bottle rest on the wooden table beside a black
drawer. Lower the black robotic gripper toward the gray drawer handle
and pull the drawer open. Then raise the gripper, move toward the
black bowl positioned above the white bowl, grasp it, lift it from
the table, and place it inside the open drawer.''
\end{promptexample}

\begin{promptexample}{landmark}{10}
``From a wrist-mounted first-person view, move to the gray handle of
the black drawer and pull it open. Then move toward the black bowl
positioned above the white bowl, grasp and lift it, and place it
inside the open drawer.''
\end{promptexample}

\end{annotationprompt}
\vspace{0.15cm}
\end{listing*}

\begin{listing*}[!t]
\caption{Annotation prompt used to generate 3D gaming instructions
(Minecraft source).}
\label{appx.tab:annot-gaming-mc}

\begin{annotationprompt}[GamingColor]
\small
\setlength{\parindent}{0pt}
\setlength{\parskip}{2pt}

\promptfield{Role}
You are a Minecraft first-person visual navigation data annotation
assistant.

\promptfield{Inputs}
Minecraft first-person screenshots in temporal order; a 2D top-down
trajectory map, trajectory metadata, horizontal travel distance,
elevation change, and a turn summary.

\promptfield{Task}
Combine the scene visible in the screenshots, their temporal order,
and the 2D trajectory geometry to produce three clearly distinct
English navigation instructions for the same Minecraft movement
trajectory. The 2D trajectory map is for judging direction changes
only. Do not output your reasoning; output JSON only.

\promptfield{Coordinates}
Each position frame is $[x,y,z]$, where $x,y$ is the horizontal map
plane and $z$ is elevation; each delta frame is
$[\Delta x,\Delta y,\Delta z,\Delta\text{yaw}]$, with yaw in radians.
The trajectory map uses the $(x,y)$ components of position.

\promptfield{Turn rule}
Ignore negligible perturbations.

\promptsection{Output styles}

\begin{promptstyle}{concise}{10}
The shortest and most abstract style, like a tight chain of action
commands. It retains only the core movement, left/right turns,
necessary key positions, and the final stop. It does not list the
blocks, trees, animals, or environmental details seen along the way,
and avoids lengthy passing/past process descriptions.
\end{promptstyle}

\begin{promptstyle}{intricate}{4}
The most detailed of the three styles, clearly longer than concise
and landmark. It describes how the action unfolds strictly in temporal
order, drawing as much as possible on Minecraft details genuinely
visible in the screenshots (terrain, block textures, roads or paths,
trees, vegetation, water, buildings, fences, bridges, torches,
farmland, animals, sky, and lighting), tying this visual information
to the action nodes.
\end{promptstyle}

\begin{promptstyle}{landmark}{10}
The length and level of detail lie between concise and intricate.
The instruction is anchored on the most recognizable Minecraft
landmarks or path nodes (a distinctive building, stone tower, bridge,
village entrance, fence, torch, farmland, pool, clear forest edge,
or distinctive terrain), explicitly stating which action is performed
on reaching each landmark. It is more structured and less exhaustive
about roadside details than intricate, but provides more
landmark-grounding information than concise.
\end{promptstyle}

\promptsection{Shared constraints}

\begin{promptconstraint}
Each of the three styles must be no more than 100 words.
\end{promptconstraint}

\promptsection{Reference example}

\textit{One of several examples provided in context.}

\begin{promptexample}{concise}{10}
``Walk straight, turn left at the stone tower, continue straight,
and stop beside the oak fence.''
\end{promptexample}

\begin{promptexample}{intricate}{4}
``Move forward across the grassy clearing toward the stone tower,
keeping the oak trees on your left and the water pool on your right.
Continue along the dirt path until the tower fills the center of
view, turn left around its corner, and follow the torch-lit path
past the wooden fence and small wheat farm. Keep walking toward
the bridge, cross it, and stop beside the birch fence at the far end.''
\end{promptexample}

\begin{promptexample}{landmark}{10}
``Head toward the stone tower, turn left at its corner, follow the
torch-lit path to the bridge, and stop beside the birch fence.''
\end{promptexample}

\end{annotationprompt}
\end{listing*}

\begin{listing*}[!t]
\caption{Annotation prompt used to generate 3D gaming instructions
(CS:GO source).}
\label{appx.tab:annot-gaming-csgo}

\begin{annotationprompt}[GamingColor]
\small
\setlength{\parindent}{0pt}
\setlength{\parskip}{2pt}

\promptfield{Role}
You are a Counter-Strike: Global Offensive first-person visual
navigation data annotation assistant.

\promptfield{Inputs}
CS:GO first-person screenshots in temporal order; a 2D top-down
trajectory map, with color from dark to light indicating time, a green
point marking the start, a red point marking the end, and a red arrow
indicating yaw heading; trajectory metadata, travel distance,
elevation change, and a turn summary.

\promptfield{Task}
Combine the scene genuinely visible in the screenshots, their temporal
order, and the trajectory geometry to produce three clearly distinct
English navigation instructions for the same CS:GO movement trajectory.
The player's arms are typically visible at the bottom of the frame,
a crosshair is typically visible at the center, and some segments
also show reloading, holding a knife-shaped prop, teammates, or other
in-game states. Do not output your reasoning; output JSON only.

\promptfield{CS:GO context requirement}
All three instructions must preserve the first-person FPS context
and include at least one CS:GO feature genuinely visible in the
screenshots. CS:GO features should support the movement instruction,
not replace the description of visible map structures such as roads,
walls, crates, doors, and corridors.

\promptfield{Coordinates}
Each position frame is $[x,y,z]$, where $x,y$ is the horizontal map
plane and $z$ is elevation; each delta frame is
$[\Delta x,\Delta y,\Delta z,\Delta\text{yaw}]$, with yaw in radians.

\promptfield{Turn rule}
The red yaw arrows in the trajectory map are the primary direction
signal; the position curve is only for judging the path and stopping
location. Do not treat every arrow as a separate turn.

\promptsection{Output styles}

\begin{promptstyle}{concise}{10}
Minimal and command-like, retaining only the core movement action,
direction change, at most one turn/stop positional cue when necessary,
and the final stop. A very short CS:GO context phrase may be added,
but the scene must not be expanded.

\end{promptstyle}

\begin{promptstyle}{intricate}{4}
The most detailed of the three styles, clearly narrating in temporal
order how the action unfolds, combining the crosshair, teammates,
and CS:GO scene details genuinely visible in the screenshots
(roads, walls, crates, containers, doors, windows, stairs, corners,
corridors, railings, vehicles, vegetation, building edges, lighting,
and sky). It states where movement continues, where a turn occurs,
how movement continues after the turn, and where it stops. It should
make full use of rich screen content but never fabricate anything
that is not visible.
\end{promptstyle}

\begin{promptstyle}{landmark}{10}
Organizes the instruction around the most recognizable CS:GO
landmarks or key path nodes visible in the screenshots (crates,
a building entrance, a long corridor, stairs, a door, a corner,
a wall, a vehicle, or a road intersection), explicitly stating
which action is performed on reaching each landmark, while retaining
one natural crosshair or teammate context cue. It is more specific
than concise and more structured and concise than intricate.
\end{promptstyle}

\promptsection{Shared constraints}

\begin{promptconstraint}
All three instructions must describe the same real trajectory and
must not contradict one another in action order or main turns.
The instructions should read like movement orders a CS:GO player
would receive rather than generic robot-navigation sentences.
Invisible target coordinates, metadata values, or trajectory-map text
must not be written directly into the instructions.
Elevation change should be mentioned only when stairs, a ramp,
a platform, or a clear height difference is visible.
If a landmark is unclear, fall back on a visible wall, road, crate,
door, or corridor as a conservative reference rather than inventing
map structure.
Each of the three styles must be no more than 100 words.
\end{promptconstraint}

\promptsection{Reference example}

\textit{One of several examples provided in context.}

\begin{promptexample}{concise}{10}
``Walk up the stairs, turn right at the top, continue straight,
and stop.''
\end{promptexample}

\begin{promptexample}{intricate}{4}
``Keep the gun-shaped prop raised while steadily ascending the
concrete stairs. At the top, continue alongside the gray textured
wall and turn right onto the open platform. Move forward past the
crouching player and another character on the right, then stop near
the gray box ahead.''
\end{promptexample}

\begin{promptexample}{landmark}{10}
``Ascend the stairs, pass the gray wall at the top, and turn right
onto the open platform. Continue straight and stop near the gray
box ahead.''
\end{promptexample}

\end{annotationprompt}
\vspace{0.2cm}
\end{listing*}


\begin{listing*}[ht!]
\caption{Annotation prompt used to generate driving instructions.}
\label{appx.tab:annot-driving}

\begin{annotationprompt}[DrivingColor]
\fontsize{8pt}{9pt}\selectfont
\setlength{\parindent}{0pt}
\setlength{\parskip}{2pt}

\promptfield{Role}
You are an autonomous-driving visual navigation instruction
generation assistant.

\promptfield{Inputs}
A sequence of evenly sampled, temporally ordered first-person driving
images; a 2D top-down trajectory map (color from dark to light
indicates time; a red arrow indicates the vehicle's yaw heading);
trajectory metadata and a speed summary (for reference and must not
be ignored).

\promptfield{Task}
Combine the images, the 2D trajectory geometry, and the speed
information to produce three English autonomous-driving navigation
instructions. The 2D trajectory map is for judging direction changes
only. Do not output your reasoning; output JSON only.

\promptfield{Rules}
Focus on the overall geometric shape of the trajectory line, clear
turns, intersection turns, continuous curves, lane changes,
entering or leaving a main road, and the associated speed adjustments
(accelerating, decelerating, or holding a steady speed).

If the trajectory contains a clear stopping segment (multiple
consecutive frames with speed $<0.5$ m/s and negligible displacement),
all three styles must mention the stop. If the vehicle resumes travel
after stopping, this must be described as continuing or proceeding
after stopping. If the speed approaches zero and displacement stops
increasing toward the end of the trajectory, all three styles must
end with stop or an equivalent expression.

\promptsection{Output styles}

\begin{promptstyle}{concise}{10}
The shortest and most abstract style, retaining only the core driving
action, direction change, and speed action when necessary. It reads
like a concise chain of driving commands, foregrounding core events
such as go straight, turn left, turn right, merge, change lane,
or stop. It does not expand on roadside scene details unless a
positional cue is necessary for a turn or a stop.

\end{promptstyle}

\begin{promptstyle}{intricate}{4}
The most detailed style, narrating clearly in temporal order how
the action unfolds and describing as richly as possible the vehicles,
pedestrians, road infrastructure, buildings, vegetation, sky,
road surface, and lane environment visible in the images.
Speed changes should be woven naturally into the narration
(e.g., proceeding smoothly, decelerating while approaching a turn,
accelerating through an intersection, or holding a moderate speed
along the main road). Stopping or stop-and-go behavior should be
described with a natural cause (waiting at a red light, yielding,
or traffic congestion), together with the process of resuming travel.
\end{promptstyle}

\begin{promptstyle}{landmark}{10}
Organizes the path around the most salient key landmarks or road
nodes (a traffic light, intersection, lane split, crosswalk,
construction barrier, gas station, distinctive building, or cluster
of parked vehicles), explicitly stating which action is performed
at each landmark. It should be more concise and structured than
intricate. A stop at a key landmark must be explicitly stated.
\end{promptstyle}

\promptsection{Shared constraints}

\begin{promptconstraint}
Do not invent any details that do not appear in the images.

Each of the three styles must be no more than 100 words.
\end{promptconstraint}

\promptsection{Reference example: no stop}

\textit{One of several examples provided in context.}

\begin{promptexample}{concise}{10}
``Drive straight ahead, turn right at the intersection, and continue
forward.''
\end{promptexample}

\begin{promptexample}{intricate}{4}
``Proceed straight along the busy multi-lane road at a steady speed,
passing a row of parked cars and storefronts on your right, while
several vehicles travel in the adjacent lanes and pedestrians remain
visible on the sidewalks. As you approach the large signalized
intersection where multiple cars are waiting, decelerate smoothly,
turn right onto the side street, and accelerate gently as you
continue past the houses and trees.''
\end{promptexample}

\begin{promptexample}{landmark}{10}
``Move forward down the main avenue at a steady speed, passing the
gas station on your right. At the signalized intersection beside
the tall glass building, decelerate, turn right, and accelerate
gently along the tree-lined street.''
\end{promptexample}

\promptsection{Reference example: with stop}

\textit{One of several examples provided in context.}

\begin{promptexample}{concise}{10}
``Continue toward the traffic light ahead, stop there, then proceed
through the intersection and turn right.''
\end{promptexample}

\begin{promptexample}{intricate}{4}
``A two-lane residential road extends ahead, with buildings and
parked vehicles lining both sides. Continue at a moderate speed
toward the intersection, where a red traffic light and a line of
waiting cars come into view. Decelerate gradually and stop behind
the queued vehicles. When the light turns green, accelerate gently,
turn right onto the side street, and continue forward past a row
of trees and a construction barrier on your left.''
\end{promptexample}

\begin{promptexample}{landmark}{10}
``Approach the traffic light at the upcoming intersection, stop
behind the queued vehicles, then turn right once the signal changes
and proceed past the construction barrier.''
\end{promptexample}

\end{annotationprompt}
\vspace{0.3cm}
\end{listing*}

The 3D gaming domain constitutes $21\%$ of \bench{} and features two distinct embodiments: a first-person avatar in competitive maps and a player character in procedurally generated open worlds. Both incorporate a vertical displacement $z_t$ into the base parameterization, defining a 4-dimensional action space that accounts for jumping, falling, and terrain-following, alongside planar movement and yaw. 3D gaming is the only exception to the source-level partition used in the other three domains: it is partitioned by map rather than by dataset. Because excluding an entire dataset would eliminate one of the two unique embodiments from the training distribution, specific maps and locations of both sources are instead excluded from training entirely and reserved for val unseen. These held-out maps play the role that held-out sources play in the other domains, preserving embodiment diversity during training while evaluating generalization to novel scene layouts and visual conditions.

\noindent \textbf{CS:GO}~\citep{pearce2021csgo}.~This dataset comprises 5.5 million frames ($\sim$ 95 hours) of \textit{Counter-Strike: Global Offensive} gameplay. The data was aggregated from public servers via spectator mode with inferred actions, supplemented by demonstrations. Trajectories exhibit high-speed, reactive dynamics such as strafing, sprinting, and abrupt view rotations, requiring instructions to be grounded in salient architectural references rather than slowly evolving scene content.

\noindent \textbf{Minecraft}~\citep{lian2025toward}.~Derived from the LoopNav dataset, this source provides 250 hours (approximately 20 million frames) of loop-based navigation with recorded actions, collected across 147 diverse locations within the open-world environment of \textit{Minecraft}.

The driving domain involves ego-vehicle motion on public roads and features the highest speeds in \bench{}. In addition to the base parameterization, a scalar velocity $v_t$ is introduced, which is critical for disambiguating maneuvers where identical planar displacements result from significantly different vehicle speeds. Across all datasets, multi-view recordings are filtered to retain only the front-facing camera as the agent-centric observation. Native resolutions, varying from $1408\times376$ (KITTI-360) to $1920\times1280$ (Waymo), are uniformly resized to $256\times256$. Because raw driving recordings are continuous and substantially exceed the duration of typical instructable maneuvers, scenes are segmented into clips of 8--60 frames, with segments containing reverse motion discarded.

\noindent \textbf{DrivingDojo}~\citep{wang2024drivingdojo}.~As the largest dataset in the driving split, DrivingDojo distills approximately 7{,}500 hours of daily driving video using planning-and-control commands, manually defined rules, and model-assisted labeling. The dataset specifically targets maneuver completeness, encompassing longitudinal actions (e.g., acceleration, emergency braking, stop-and-go) and lateral actions (e.g., U-turns, overtaking, lane changes), alongside dense multi-agent interactions such as cut-ins and merging~\citep{chi2026driver,chi2026pv}. Planar positions are derived from per-frame camera-to-world extrinsics, yaw is extracted from the horizontal projection of the camera axis, and world-frame displacements are projected onto the current heading to compute the signed forward velocity of the vehicle at each time step.

\noindent \textbf{Waymo}~\citep{sun2020scalability}.~The Waymo Open Dataset provides 1{,}150 scenes (20 seconds each) recorded with a synchronized multi-sensor suite across San Francisco, Phoenix, and Mountain View. Planar position and yaw are extracted directly from the provided vehicle-pose matrices. The onboard sensor-reported vehicle velocity is rotated into the ego frame to isolate the forward component.

\noindent \textbf{KITTI-360}~\citep{liao2022kitti}.~KITTI-360 comprises over 300k images of suburban driving sequences, featuring precise georegistered vehicle and camera poses obtained via GNSS/IMU fusion. As the only dataset providing pre-computed positions, headings, per-step increments, and velocities, its released increments are verified against consecutive pose differences and adopted directly.

\noindent \textbf{nuScenes}~\citep{caesar2020nuscenes}.~nuScenes consists of 1{,}000 scenes (20 seconds each) recorded in Boston and Singapore. Notably, it introduces severe environmental variability, including nighttime and rainy conditions, as well as left-hand traffic conventions. Scene boundaries and frame orderings are reconstructed from the released metadata. Translations and rotation quaternions from annotated keyframes are converted into yaw angles, with velocities derived via finite differences. Given its significant disparities in appearance, geography (pairing Asian and North American cities), and driving conventions (where left-hand traffic geometrically inverts turn instructions), nuScenes is excluded from training and reserved for the val unseen and test partitions to assess robust generalization to these held-out driving environments.

\subsection{Annotation Pipeline} 
\label{appx-annotation}

To initialize the automated annotation pipeline, we manually annotated all three instruction styles for a seed set of 150 trajectories, spanning representative scenes across all domains. These expert-curated triplets serve as few-shot exemplars within the domain-specific prompts~\citep{liu2025grl}. During generation, the model is conditioned on multiple contextually relevant references drawn from structurally similar scenes~\citep{wang2025we}, extending beyond the single reference example shown in each of Listings A1--A5.

Conditioned on these exemplars, Qwen3.6-27B~\citep{qwen3.6-27b} generates all three instruction styles simultaneously for each remaining episode (prompts for each domain are provided in Listings A1--A5). The model input integrates evenly sampled first-person frames with a compact geometric trajectory summary. Specifically, navigation, driving, and gaming domains utilize a 2D top-down path map paired with distance, heading-change, and speed statistics. For robotics, the input comprises a storyboard of concatenated frames alongside the 7-DoF action sequence and gripper states, facilitating continuous object tracking during the manipulation. Every generated instruction undergoes rigorous human verification against the corresponding video. Annotators manually refine the text to ensure factual correctness and linguistic naturalness, finalizing the three instruction variants released per episode. Furthermore, a random subset of these finalized instructions is independently evaluated by human raters, as detailed in Sec.~\ref{appx-human-eval}.

\begin{table*}[!t]
\small
\centering
\setlength{\tabcolsep}{4pt}
\caption{\textbf{Multi-style instruction samples from \bench{}.} One episode per domain, each paired with all three instruction styles. The styles share the same trajectory and differ only in the level of specificity and the degree of visual anchoring, enabling a controlled evaluation of language grounding across domains.}
\begin{tabular}{@{}cl p{\dimexpr\textwidth-4\tabcolsep-1.5em-5em\relax}@{}}
\toprule[1.5pt]
& \makebox[5em][l]{Style} & Instruction \\
\midrule
\multirow{10}{*}{\makebox[1.5em][c]{\rotatebox[origin=c]{90}{\emph{\textbf{\textcolor[HTML]{3F73B3}{Navigation}}}}}}
& \makebox[5em][l]{Concise} & Move forward, turn left and continue straight until you reach the open area, then stop. \\
\cmidrule(l){2-3}
& Intricate & Start by moving forward through the open space with a yellow archway on the left and a blue seating area ahead. Follow the path as it curves gently to the left, then continue past the blue seating area and a white pillar. The floor transitions from carpet to a shiny white surface. Proceed into the larger open area with tables and chairs visible in the distance, then stop. \\
\cmidrule(l){2-3}
& Landmark & Walk forward past the yellow archway and blue seating area. Follow the path as it curves left, then continue past the white pillar onto the shiny white floor. Proceed into the open area with tables and stop. \\
\midrule
\multirow{11}{*}{\makebox[1.5em][c]{\rotatebox[origin=c]{90}{\emph{\textbf{\textcolor[HTML]{F1B287}{Robotics}}}}}}
& Concise & From an external third-person view, grasp and lift one corner of the blue cloth, then fold it diagonally. \\
\cmidrule(l){2-3}
& Intricate & From an external third-person view, the robot arm approaches a light blue cloth spread flat on a wooden table near several colorful toy vegetables. The gripper targets the bottom-right corner, closes around the fabric, and lifts it upward. The arm then moves diagonally across the table, drawing the corner toward the top-left side. As the cloth is pulled, it folds over itself into a triangular shape. The gripper releases the fabric, leaving the cloth partially folded on the table. \\
\cmidrule(l){2-3}
& Landmark & From an external third-person view, move to the bottom-right corner of the blue cloth on the wooden table, grasp it, and pull it diagonally toward the top-left corner to fold the fabric. \\
\midrule
\multirow{10}{*}{\makebox[1.5em][c]{\rotatebox[origin=c]{90}{\emph{\textbf{\textcolor[HTML]{F9746B}{3D Gaming}}}}}}
& Concise & Move straight along the jungle path and stop when the foliage opens up. \\
\cmidrule(l){2-3}
& Intricate & Begin by moving forward through the thick, green jungle canopy, navigating past the large, vine-covered tree trunk immediately on your left. Keep walking straight along the narrow, leafy path, maintaining a steady pace as the dense foliage surrounds you. As you progress, the path remains clear directly ahead, with hanging vines and lush vegetation on both sides. Continue forward until the immediate path ahead becomes slightly more open, then stop in the clearing. \\
\cmidrule(l){2-3}
& Landmark & Walk forward past the large vine-covered tree on the left, continue straight through the dense jungle path, and stop when the foliage opens up slightly ahead. \\
\midrule
\multirow{10}{*}{\makebox[1.5em][c]{\rotatebox[origin=c]{90}{\emph{\textbf{\textcolor[HTML]{762A9D}{Driving}}}}}}
& Concise & Drive straight ahead, accelerating gently through the intersection and continuing up the hill. \\
\cmidrule(l){2-3}
& Intricate & Start by driving straight through the wide intersection under green traffic lights, accelerating gently behind a black sedan. Continue uphill along the straight, tree-lined residential street, with parked cars and buildings lining both sides. Maintain a steady, moderate speed and a safe distance from the sedan as you proceed under the clear blue sky. \\
\cmidrule(l){2-3}
& Landmark & Proceed straight through the intersection beneath the green traffic lights, then continue up the steep hill between rows of parked cars and residential buildings at a steady pace. \\
\bottomrule[1.5pt]
\end{tabular}
\label{appx.tab:instruction-examples}
\end{table*}

\begin{figure*}[!ht]
    \centering
    \includegraphics[width=1\linewidth]{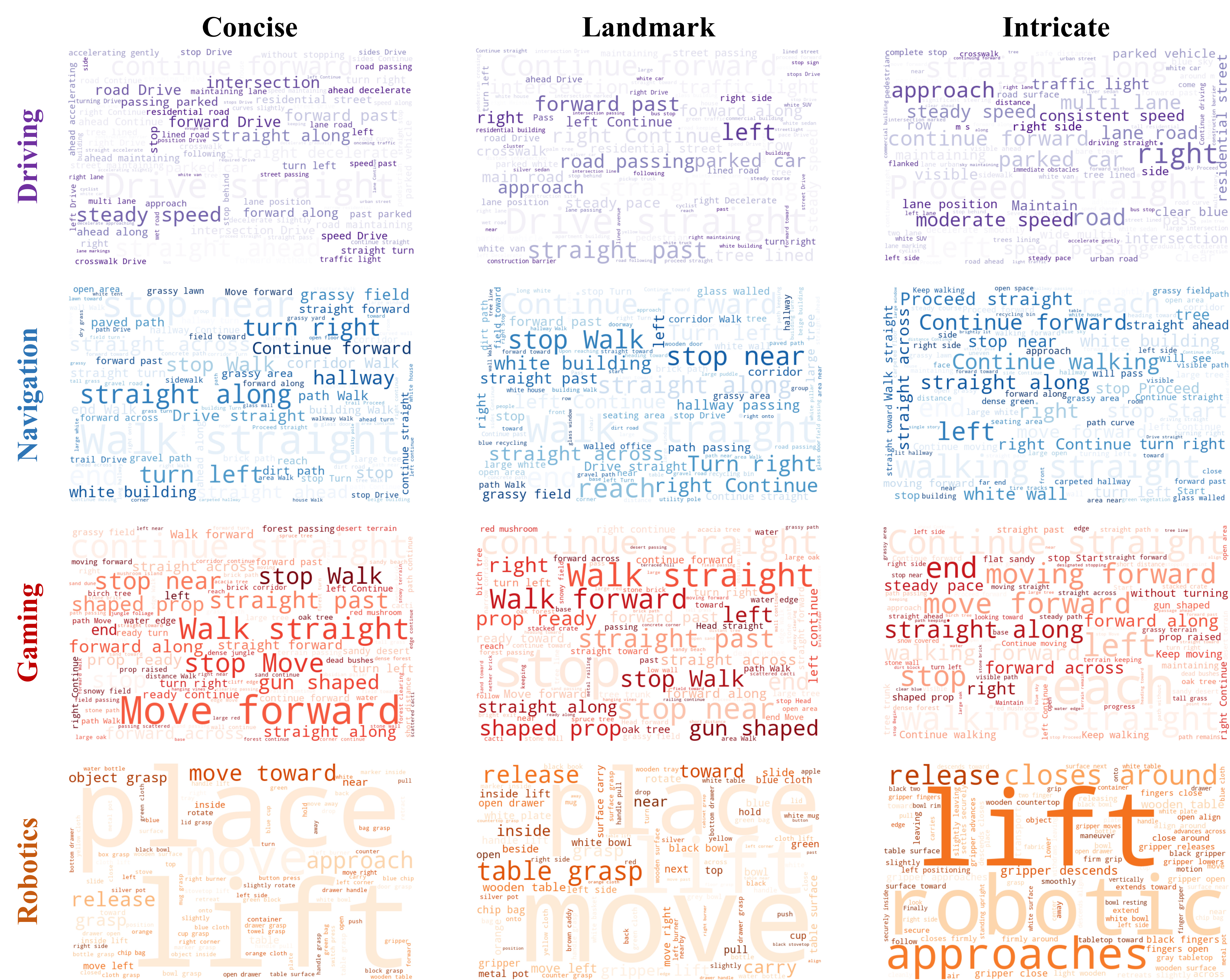}
    \caption{\textbf{Word clouds of the released instructions, by domain and instruction style.} Word size reflects frequency within the corresponding style-domain subset of \bench{}.}
    \label{appx.fig:word-cloud}
\end{figure*}

\subsection{Human Evaluation of Instruction Quality}
\label{appx-human-eval}

To quantify the quality of the released instructions, we conduct a human study on a random subset of \bench{}. We draw 1{,}200 episodes, approximately balanced across the four domains and the three instruction styles, and recruit volunteer raters from our research group who took no part in the annotation pipeline. For each item, a rater watches the agent-view video, inspects the accompanying 2D or 3D trajectory map, and scores the paired instruction on five dimensions along a six-point scale, where the top score denotes an instruction indistinguishable from one written by human annotators. The dimensions are: (1)~\emph{Trajectory--Instruction Alignment}, whether the described path matches the executed one in its turns, distances, and stopping point; (2)~\emph{Style Adherence}, whether the instruction stays within its assigned style instead of drifting toward another; (3)~\emph{Geometric Correctness}, whether directional references and spatial relations hold from the agent's own viewpoint; (4)~\emph{Landmark Accuracy}, whether the referenced visual landmarks exist in the scene and are correctly placed; and (5)~\emph{Naturalness \& Fluency}, whether the wording reads like language a human guide would actually use. Every episode--instruction pair receives three independent ratings, which we average over raters and dimensions. The mean over all 1{,}200 samples is 5.37 out of 6.00, indicating that the released instructions sit close to human-written quality.

\subsection{Multi-Style Instruction Examples}
\label{appx-ins-example}
Tab.~\ref{appx.tab:instruction-examples} presents one representative episode per domain, each annotated in all three styles. Because the three styles describe the same episode, they differ only in specificity and emphasis, never in the underlying trajectory. Concise keeps only what is needed to reproduce the path; Landmark-based anchors each maneuver to a named visual feature, such as the yellow archway in navigation or the green traffic lights in driving; and Intricate embeds the same path in extended scene description, much of which is not required for execution. The three styles are not arbitrary partitions of a single distribution but correspond to distinct grounding demands, and they follow how people actually give directions, which range from terse imperatives to landmark-anchored guidance to running commentary depending on the speaker and on how familiar the listener is assumed to be with the scene~\citep{kolagar2024aligning, kong2024controllable, wu2025govig}. Accordingly, Concise probes whether a model can follow minimal directional language, Landmark-based whether it can ground words in specific visual features, and Intricate whether it can recover control-relevant content from verbose description.

\subsection{Data Splits}
\label{appx-splits}
\bench{} is split at the episode level, so the three instruction styles of an episode always fall into the same split. Its 142{,}987 episodes are divided into 100{,}092 for training, 12{,}869 for val seen, 11{,}439 for val unseen, and 18{,}587 for test, corresponding to 300{,}276, 38{,}607, 34{,}317, and 55{,}761 episode--instruction pairs, and every domain follows the same proportions of roughly 70\%, 9\%, 8\%, and 13\%. In navigation, robotics, and driving, the partition is made at the source level (Sec.~3.2 of the main paper). The held-out sources TartanDrive, LIBERO, and nuScenes contribute no episodes to training or val seen; they supply the entire val unseen split of their domains (3{,}291, 4{,}556, and 1{,}168 episodes, respectively) and place their remaining episodes in test (2{,}605, 7{,}220, and 549). All other sources are divided among training, val seen, and test, so that val seen consists of held-out episodes from sources observed during training. In 3D gaming, where the partition is made at the map level, both sources appear in every split: Minecraft and CS:GO contribute 17{,}593 and 3{,}614 episodes to training, 2{,}262 and 465 to val seen, 1{,}887 and 537 episodes from held-out maps to val unseen, and 3{,}188 and 750 to test. The test split therefore mixes seen and unseen data.

\subsection{More Benchmark Statistics}

Tab.~\ref{tab:datasets} summarizes the full composition of \bench{}: the source dataset, real-vs-simulated status, action dimensionality, and episode/instruction counts for each of the 15 datasets across the four domains.
Fig.~\ref{appx.fig:word-cloud} aggregates the released instructions into per-style word clouds for each domain, complementing the single-episode examples of Tab.~\ref{appx.tab:instruction-examples} with a corpus-level view of the same three-way separation. Across all four domains, Concise instructions are dominated by the motion and control verbs that name a maneuver directly, such as \emph{turn right}, \emph{move forward}, and \emph{grasp}, \emph{lift}, \emph{release} in robotics: the vocabulary needed to reproduce the trajectory and little else. Landmark-based instructions carry the same verbs but pair them with frequent, concrete object nouns, such as \emph{white building}, \emph{parked car}, and \emph{wooden table}, reflecting the explicit tie between each maneuver and a recognizable anchor that the style is designed to enforce. Intricate instructions retain this vocabulary as well, but surround it with descriptive modifiers and scene qualifiers that are largely absent from the other two styles, such as \emph{consistent speed}, \emph{carpeted hallway}, and \emph{gripper closes around}, consistent with the extended descriptions of surrounding layout and appearance. This pattern shows that the three styles differ not only in length, as Fig.~2d already reports, but in the composition of vocabulary.

The domain-specific vocabulary within each style further reflects the action space of Sec.~3.1 together with each domain's typical scene content. Driving instructions cluster around traffic elements such as intersections, crosswalks, and parked cars, alongside speed-related terms tied to the velocity channel unique to that domain. Navigation vocabulary is architectural, dominated by terms like hallway, corridor, and building. Gaming instructions reference terrain and vegetation, such as grassy field, oak tree, and birch tree. Robotics instructions stand apart from the other three domains at the vocabulary level: they are dominated by contact verbs tied to gripper, such as grasp and release.

\begin{table*}[t]
\setlength{\abovecaptionskip}{2pt}
\setlength{\belowcaptionskip}{0pt}
\centering
\tiny
\caption{Composition of \bench{}. Action variant follows the base parameterization $(u_t,\phi_t)$ of Sec.~3.1, with domain-specific channels ($z_t$, $v_t$, or $\rho_t,\vartheta_t,g_t$) appended where applicable. Every episode is paired with three instruction styles (concise, intricate, landmark), yielding three instruction-conditioned samples per episode.}
\resizebox{\linewidth}{!}{%
\begin{tabular}{l||lllcrr}
\thickhline
\rowcolor{gray!20}\textbf{Domain} & \textbf{Dataset} & \textbf{Source} &\textbf{Action variant} & \textbf{Dim.} & \textbf{\#Episodes}& \textbf{\#Instructions} \\
\hline\hline
\textcolor[HTML]{3F73B3}{Navigation} & Go Stanford & Real & $(u,\phi)$ & 3 & 10{,}080 & 30{,}240 \\
 & RECON & Real & $(u,\phi)$ & 3 & 10{,}392 & 31{,}176 \\
 & SCAND & Real & $(u,\phi)$ & 3 & 5{,}427 & 16{,}281 \\
 & HuRoN & Real & $(u,\phi)$ & 3 & 9{,}346 & 28{,}038 \\
 & TartanDrive & Real & $(u,\phi)$ & 3 & 5{,}896 & 17{,}688 \\
\hline
\textcolor[HTML]{F1B287}{Robotics} & BridgeData V2 & Real & $+\,z,\rho,\vartheta,g$ & 7 & 10{,}000 & 30{,}000 \\
 & DROID & Real & $+\,z,\rho,\vartheta,g$ & 7 & 25{,}172 & 75{,}516 \\
 & LIBERO & Sim & $+\,z,\rho,\vartheta,g$ & 7 & 11{,}776 & 35{,}328 \\
 & RT-1 & Real & $+\,z,\rho,\vartheta,g$ & 7 & 10{,}000 & 30{,}000 \\
\hline
\textcolor[HTML]{F9746B}{3D Gaming} & Minecraft (LoopNav) & Sim & $+\,z$ & 4 & 24{,}930 & 74{,}790 \\
 & CS:GO & Sim & $+\,z$ & 4 & 5{,}366 & 16{,}098 \\
\hline
\textcolor[HTML]{762A9D}{Driving} & DrivingDojo & Real & $+\,v$ & 4 & 6{,}805 & 20{,}415 \\
 & KITTI-360 & Real & $+\,v$ & 4 & 1{,}984 & 5{,}952 \\
 & Waymo & Real & $+\,v$ & 4 & 4{,}096 & 12{,}288 \\
 & nuScenes & Real & $+\,v$ & 4 & 1{,}717 & 5{,}151 \\
\hline
Total & 15 datasets & 12 real / 3 sim & -- & 3--7 & 142{,}987 & 428{,}961 \\
\hline
\end{tabular}}
\label{tab:datasets}
\end{table*}

\section{Tiresias-Agent Details}

\label{appx:method}
\begin{figure*}[!t]
\setlength{\abovecaptionskip}{2pt}
\setlength{\belowcaptionskip}{0pt}
    \centering
    \includegraphics[width=0.8\linewidth]{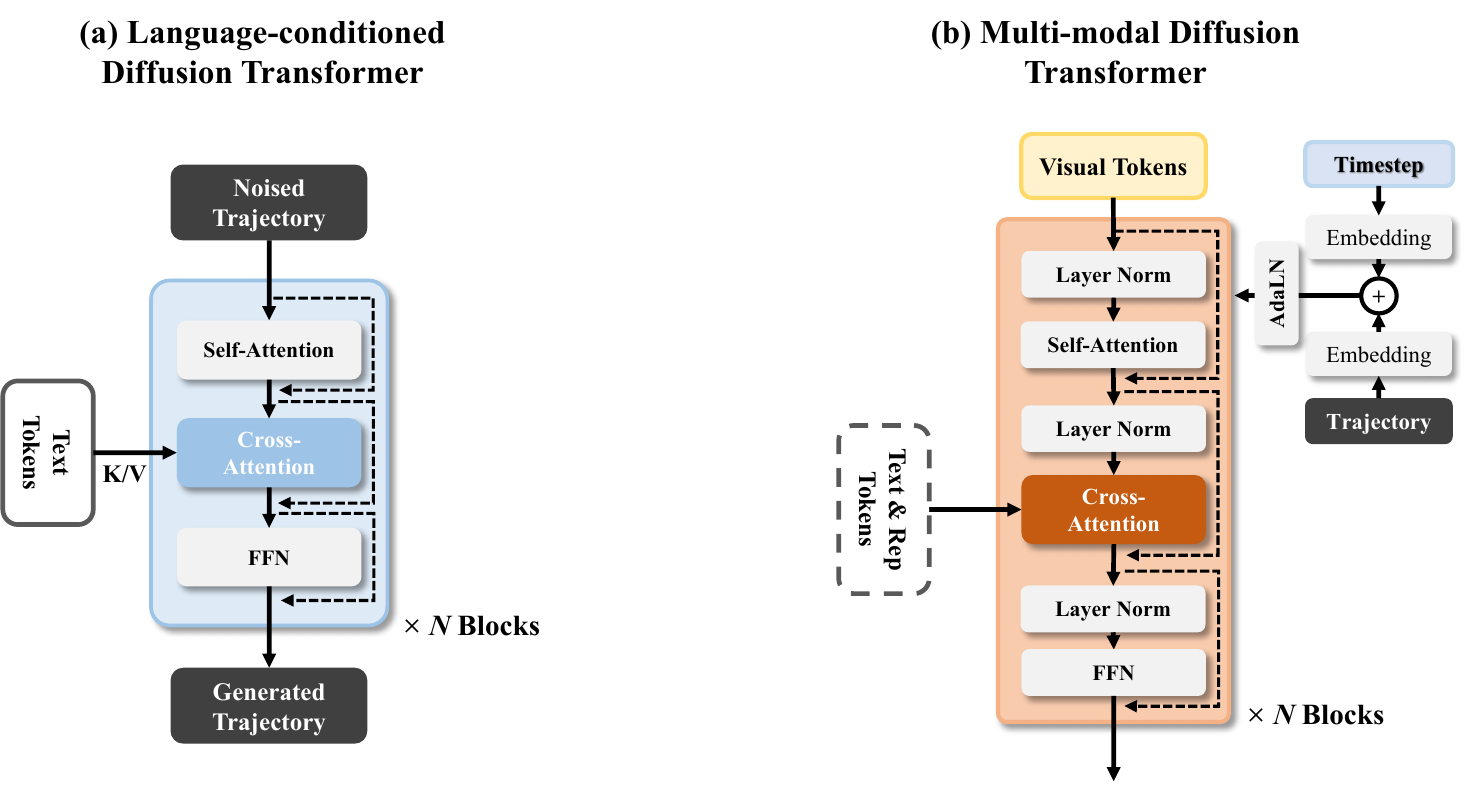}
    \caption{\textbf{Architectures of two trained modules of Tiresias-Agent.} \textbf{(a)}~LDP's language-conditioned diffusion Transformer (Sec.~\ref{appx-ldp}).~\textbf{(b)}~The world model's multimodal diffusion Transformer (Sec.~\ref{appx-wm}). }
    \label{fig:detail-struc}
\end{figure*}

\subsection{System Prompts}
\label{appx-sys-prompt}
The world model is conditioned on a domain-specific system prompt prepended to the episode instruction. Following the template in Tab.~\ref{appx.tab:system-prompts}, the concatenated text sequence is encoded by the frozen umT5 encoder and injected into the world model via cross-attention as the instruction embedding. This prompt is fixed per domain across all episodes and remains strictly identical during training and inference.

\begin{table*}[!t]
\small
\centering
\caption{\textbf{System prompts used to condition the language-conditioned world model.} For each domain, the system prompt and the episode instruction are concatenated into one text sequence, encoded by the frozen umT5 encoder, and supplied to the world model. Prompts are fixed per domain and identical at training and inference time.}
\renewcommand{\arraystretch}{1.4}
\begin{tabular}{p{6em}p{40em}}
\toprule[1.5pt]
\textbf{\textcolor[HTML]{3F73B3}{Navigation}} & \textbf{System}: You are a language-conditioned navigation world model. One initial observation is provided together with the route instruction and the full action sequence. Generate the complete future agent-view video in one shot. Treat the instruction as the route semantics and the action sequence as transition control for timing, direction, and motion scale. Keep motion physically plausible, temporally continuous, and faithful to both instruction and action. \newline \textbf{Instruction}: \{language instruction\} \\
\hline
\textbf{\textcolor[HTML]{F1B287}{Robotics}} & \textbf{System}: You are a language-conditioned robot-manipulation world model. One initial observation is provided together with the manipulation instruction and the full 7-DoF end-effector action sequence. Generate the complete future video in one shot. Preserve the viewpoint stated at the beginning of the instruction: a wrist-mounted first-person view must remain wrist-mounted first-person, and an external third-person view must remain external third-person. In an external third-person view, camera extrinsics are fixed: never pan, tilt, zoom, dolly, orbit, or track the robot. In a wrist-mounted first-person view, the camera moves only because it is rigidly attached to the robot wrist. The actor is a robotic arm with a mechanical gripper; never replace it with a human hand, human arm, skin, or fingers. Preserve whether the initial scene is real or simulated rather than changing visual domain. Treat the instruction as task and contact semantics, and the action sequence as transition control for end-effector translation, rotation, and gripper opening or closing. Maintain robot identity, geometry, kinematic continuity, object permanence, contact, grasp, lift, transport, placement, and release. \newline \textbf{Instruction}: \{language instruction\} \\
\hline
\textbf{\textcolor[HTML]{F9746B}{3D Gaming}} & \textbf{System}: You are a language-conditioned first-person gaming world model. One initial observation is provided together with the gameplay instruction and the full ego-action sequence. Generate the complete future first-person gaming video in one shot. Treat the instruction as task semantics and the action sequence as transition control for forward motion, lateral strafing, yaw rotation (looking left/right), and vertical movement (jumping, falling, stairs). Keep scene geometry, player motion, lighting, and object appearance physically plausible and temporally continuous. Be faithful to both instruction and action. \newline \textbf{Instruction}: \{language instruction\} \\
\hline
\textbf{\textcolor[HTML]{762A9D}{Driving}} & \textbf{System}: You are a language-conditioned self-driving world model. One initial front-facing observation is provided together with the route instruction and the full ego-action sequence. Generate the complete future driving video in one shot. Treat the instruction as route semantics and the action sequence as transition control for speed, direction, and motion scale. Keep road geometry, ego-motion, traffic, and lighting physically plausible and temporally continuous. Be faithful to both instruction and action. \newline \textbf{Instruction}: \{language instruction\} \\
\bottomrule[1.5pt]
\end{tabular}
\label{appx.tab:system-prompts}
\vspace{0.2cm}
\end{table*}

\subsection{LDP Details} 
\label{appx-ldp} 
As illustrated in Fig.~\ref{fig:detail-struc}(a), the Language-conditioned Diffusion Policy (LDP) denoises a sampled action sequence into a valid trajectory guided solely by language. Each of its $L$ blocks sequentially applies self-attention over the action tokens, cross-attention using the frozen umT5~\citep{chung2023unimax} instruction embeddings as keys and values, and a feed-forward layer, with residual connections throughout. Notably, apart from the diffusion timestep, this cross-attention mechanism serves as the exclusive conditioning source; visual inputs are strictly excluded. This scene-agnostic design ensures that sampling the LDP with independent noise seeds generates diverse, instruction-compliant trajectories rather than observation-constrained motions. Consequently, this produces a sufficiently diverse candidate pool for the subsequent ranking stage, effectively deferring the scene-dependent evaluation to the world model and V-Ranker. LDP is trained from scratch by denoising expert trajectories sampled from the \bench{} training split.

\subsection{Language-Conditioned World Model Details} 
\label{appx-wm} 
As illustrated in Fig.~\ref{fig:detail-struc}(b), the world model adapts the Wan2.2~\citep{wan2025wan} multimodal diffusion Transformer while preserving its native interfaces. Frames are encoded into visual tokens by a frozen VAE~\citep{wan2025wan} and subsequently noised. Each transformer block sequentially applies self-attention, cross-attention, and a feed-forward layer, interleaved with layer normalization and residual connections. Only transformer parameters are updated during training.

The two conditioning signals described in Sec.~4.2 are injected via distinct mechanisms, reflecting their respective structural roles. The candidate trajectory is embedded by a lightweight MLP, added to the timestep embedding, and injected via AdaLN, enabling it to modulate all visual tokens uniformly at each denoising step. In contrast, the instruction and scene features serve as keys and values in the cross-attention layers, formed by concatenating the prompted umT5~\citep{chung2023unimax} tokens with the DINOv3~\citep{simeoni2025dinov3} representation tokens of the initial frame. 

This architectural decoupling aligns with the distinct nature of the signals: the trajectory is a low-dimensional, temporally aligned control signal that governs macroscopic ego-motion, which is effectively captured by uniform modulation. Conversely, grounding fine-grained semantics (e.g., \emph{the yellow archway}) necessitates dense, patch-level alignment between text tokens and visual features, naturally facilitated by cross-attention. The model is trained using the flow-matching objective detailed in Sec.~4.2.

\subsection{Tr-Ranker Details} 
\label{appx-tr-ranker} 
The Tr-Ranker evaluates a candidate trajectory based solely on language and motion, operating without visual rendering. It processes the frozen umT5 instruction tokens alongside a candidate action sequence generated by the LDP. The actions are first projected into the model dimension and augmented with temporal positional embeddings to preserve maneuver ordering. These action tokens then query the instruction embeddings via cross-attention. Subsequently, a temporal Transformer aggregates the outputs into a unified representation per candidate, from which four distinct heads predict the evaluation metrics: SR, nDTW, ATE, and RPE.

Training labels are generated offline by evaluating each training-time candidate against its corresponding ground-truth trajectory. The classification head (SR) is optimized using binary cross-entropy, while the three continuous metric heads employ a Huber loss to mitigate the impact of severe outliers. During inference, the predictions from these heads are normalized to a common scale and fused into the final language-side score $S^{(m)}_{\mathrm{tr}}$, where SR and nDTW are incorporated as positive rewards, and ATE and RPE as penalties.

\subsection{V-Ranker Details} 
\label{appx-v-ranker} 
The V-Ranker evaluates each candidate through the future states predicted by the frozen world model, without directly accessing the candidate action sequence. For candidate $m$, the world model generates a sequence of video latents $\hat{X}^{(m)}$. The anchor frame is excluded from the rollout descriptor so that it summarizes the imagined future. For each latent channel, we compute the temporal mean, standard deviation, and mean absolute frame-to-frame difference, summarizing appearance and temporal variation. The resulting descriptor is projected to a shared dimension together with three contextual features: the instruction embedding $e_I$, the scene semantic features, and a learned domain-specific embedding. The fusion MLP receives the concatenation of these representations, augmented with the element-wise products between the rollout descriptor and the projected instruction and scene features. Within an episode, contextual features are shared across candidates; candidate-specific evidence enters V-Ranker only through the imagined rollout.

Finally, four heads predict an overall score and three auxiliary scores for language--motion consistency, path fidelity, and motion accuracy. Their outputs are combined with weights of $1{:}0.5{:}0.5{:}0.25$, respectively, to obtain the final visual score $S^{(m)}_{\mathrm{v}}$, giving the auxiliary scores smaller coefficients than the overall score.

\subsection{Pseudo-Code for Inference} 
\label{appx-pseudocode} 
Alg.~\ref{alg:inference} outlines the selective-imagination pipeline of Sec.~4.4. Because all four modules are trained independently and kept frozen, inference is a direct composition without joint fine-tuning. The two stages process the language instruction differently. Stage one encodes the instruction alone; as the LDP and Tr-Ranker are scene-agnostic, they evaluate only whether a trajectory aligns with the text command. In contrast, stage two prepends a domain-specific system prompt to the instruction, providing the world model with the embodiment context required to simulate domain-appropriate futures. This pipeline relies on a structural asymmetry in computational cost. The Tr-Ranker, operating without visual inputs, efficiently scores the entire pool of $N$ candidates. The world model is then invoked exclusively for the top-$k$ candidates. Restricting simulation to this shortlist keeps ranked imagination computationally efficient compared to evaluating every proposal.

\begin{algorithm}[htbp]
\caption{Tiresias-Agent: Inference with Selective Imagination}
\label{alg:inference}
\begin{algorithmic}[1]
\REQUIRE observation $o_s$, instruction $I$, domain $d$
\REQUIRE frozen LDP $\pi$, world model $p_\theta$, Tr-Ranker, V-Ranker
\REQUIRE pool size $N$, shortlist size $k$, weights $\lambda_{\mathrm{tr}}, \lambda_{\mathrm{v}}$
\ENSURE selected action sequence $\hat{\mathcal{A}}^{\ast}$
\STATE $x_s \leftarrow \mathrm{VAE}(o_s)$, \quad $f_s \leftarrow \mathrm{DINOv3}(o_s)$
\STATE \COMMENT{Stage one: propose and rank from the instruction alone}
\STATE $T_I \leftarrow \mathrm{umT5}(I)$
\FOR{$m = 1$ to $N$}
    \STATE $\hat{\mathcal{A}}^{(m)} \sim \pi(\cdot \mid T_I;\, \epsilon_m)$
    \STATE $S^{(m)}_{\mathrm{tr}} \leftarrow \text{Tr-Ranker}(\hat{\mathcal{A}}^{(m)}, T_I)$
\ENDFOR
\STATE $\mathcal{S} \leftarrow$ indices of the top-$k$ scores in $\{S^{(m)}_{\mathrm{tr}}\}_{m=1}^{N}$
\STATE \COMMENT{Stage two: imagine only the shortlisted candidates}
\STATE $e_I \leftarrow \mathrm{umT5}(\mathrm{prompt}_d \oplus I)$
\FOR{$m \in \mathcal{S}$}
    \STATE $\hat{X}^{(m)} \sim p_\theta\big(\cdot \mid x_s,\, [f_s; e_I],\, \hat{\mathcal{A}}^{(m)}\big)$
    \STATE $S^{(m)}_{\mathrm{v}} \leftarrow \text{V-Ranker}(\hat{X}^{(m)}, e_I, f_s, d)$
\ENDFOR
\STATE $\widetilde{S}^{(m)}_{\mathrm{tr}},\, \widetilde{S}^{(m)}_{\mathrm{v}} \leftarrow$ normalize $S^{(m)}_{\mathrm{tr}},\, S^{(m)}_{\mathrm{v}}$ within $\mathcal{S}$
\STATE $m^{\ast} \leftarrow \arg\max_{m \in \mathcal{S}} \; \lambda_{\mathrm{tr}} \widetilde{S}^{(m)}_{\mathrm{tr}} + \lambda_{\mathrm{v}} \widetilde{S}^{(m)}_{\mathrm{v}}$
\RETURN $\hat{\mathcal{A}}^{\ast} = \hat{\mathcal{A}}^{(m^{\ast})}$
\end{algorithmic}
\end{algorithm}

\section{Experimental Details}
\label{appx:experiments}
\subsection{Evaluation Metric Details}
\label{appx-eval-metrics}

A predicted action sequence is forward-integrated into a position trajectory by composing each planar displacement with the accumulated yaw, and the reference trajectory is obtained the same way from the ground-truth actions. Navigation and driving are compared in the plane, $p_t=(x_t,y_t)\in\mathbb{R}^2$; 3D gaming and robotics are compared in space, $p_t=(x_t,y_t,z_t)\in\mathbb{R}^3$. $P=(p_1,\dots,p_{T_{\mathrm{gt}}})$ and $\hat{P}=(\hat{p}_1,\dots,\hat{p}_{\hat{T}})$ denote the reference and predicted trajectories, and $\delta_{\mathcal{D}}$ represents average step size of domain $\mathcal{D}$.

\noindent\textbf{SR}. Success Rate (SR) is a binary outcome at the terminal state~\citep{dong2025ha,dong2026language}: the episode succeeds if the Euclidean distance between the predicted and reference endpoints is within the domain's average step size. In robotics, success additionally requires the terminal gripper state to match the reference, since reaching the right position while releasing at the wrong time or place does not complete the task.

\noindent\textbf{nDTW}~\citep{ilharco2019general}. We align $\hat{P}$ and $P$ with dynamic time warping and take the minimum cumulative Euclidean cost $DTW(\hat{P},P)$ over order-preserving matches:
\begin{equation}
\mathrm{nDTW}=\exp\left(-\frac{DTW(\hat{P},P)}{\delta_{\mathcal{D}}T_{\mathrm{gt}}}\right),
\end{equation}
normalized to $1$ for a perfect match and decaying toward $0$ as the paths diverge, in units of average steps per reference step so it stays comparable across domains with very different motion scales. Because the alignment is monotonic but not one-to-one, nDTW tolerates a different speed or step count along the same route while still penalizing the wrong route.

\noindent\textbf{ATE and RPE}~\citep{sturm2012evaluating}. Both are computed over the temporally corresponding steps of the two trajectories, with no start-point, end-point, Procrustes, or DTW alignment applied beforehand. ATE is the RMSE of the position error between corresponding steps and reflects global drift; RPE is the RMSE of the error between corresponding step-to-step displacements and reflects local motion error. The two metrics are complementary: a plan with accurate per-step motion can still accumulate large ATE~\citep{liu2026streamvlo}, and reporting both separates local control error from the long-horizon drift.

\noindent\textbf{Imagination Metrics.}~To evaluate simulation performance, SSIM, PSNR, LPIPS~\citep{zhang2018unreasonable}, and DreamSim~\citep{fu2023dreamsim} are reported at two horizons: @1 compares the single predicted next frame to the ground truth, and @10 evaluates a ten-step open-loop rollout. FVD~\citep{unterthiner2019fvd} compares distributions of complete sequences and is reported at the horizon of @32.
\begin{figure*}[!t]
    \centering
    \includegraphics[width=1\linewidth]{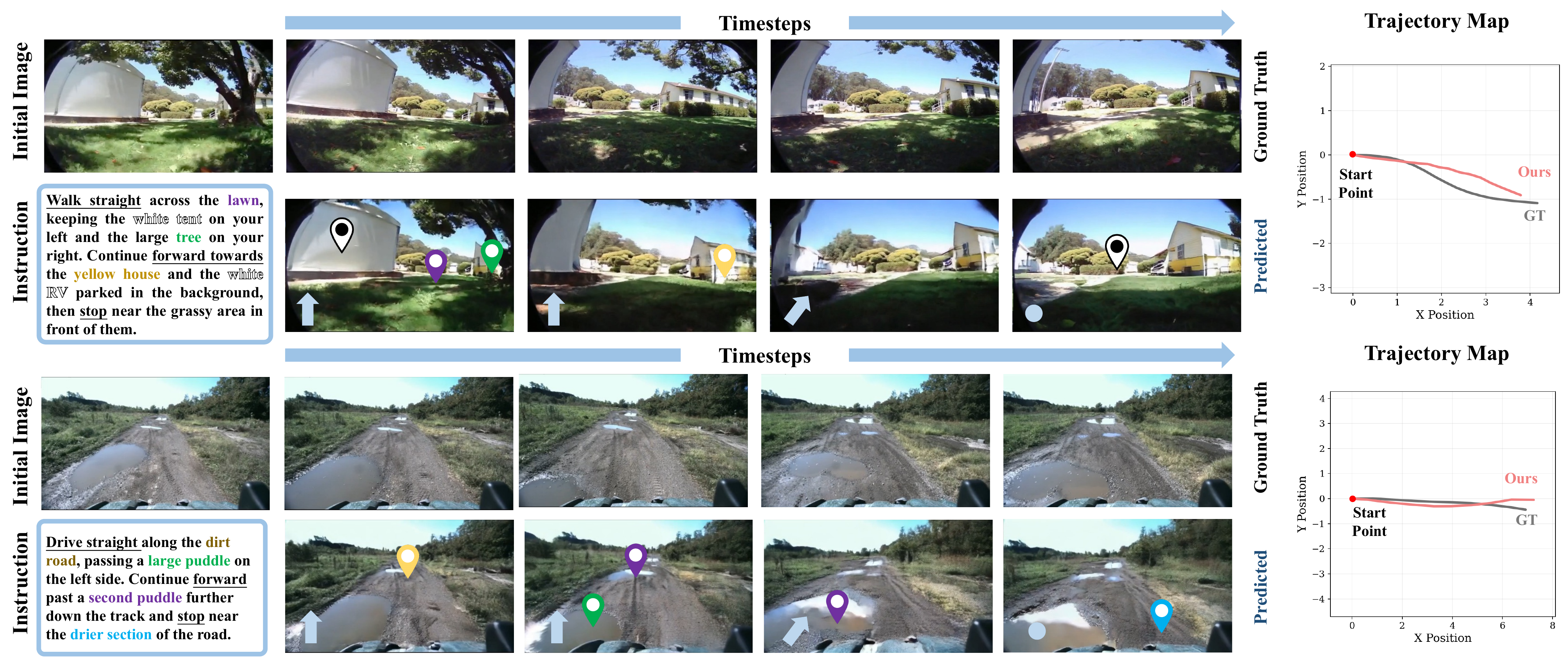}
    \caption{\textbf{More qualitative results on val seen and val unseen navigation scenarios.} For each case, we show the initial observation and instruction, the imagined rollout of the selected plan against the ground-truth future, and the predicted trajectory against the reference.}
    \label{fig-visual_nav1}
\end{figure*}
\begin{figure*}[!t]
    \centering
    \includegraphics[width=1\linewidth]{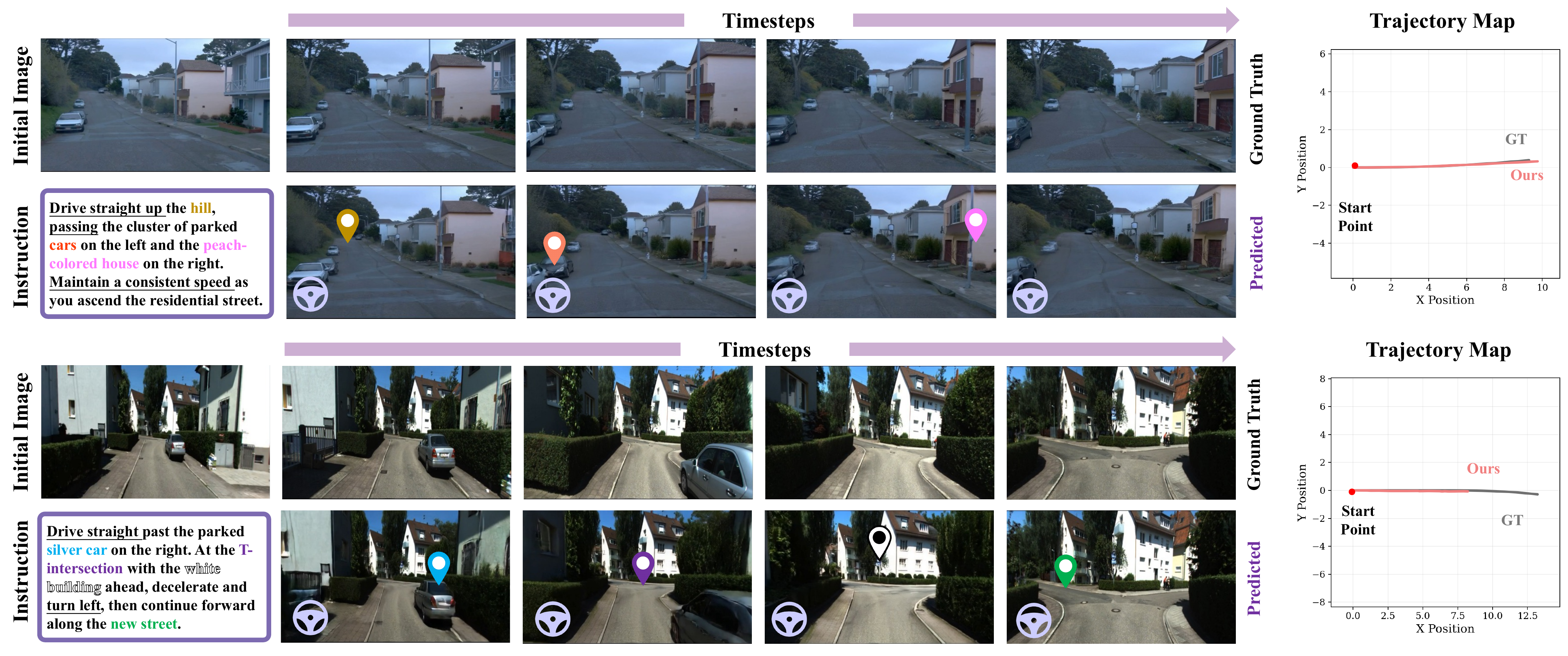}
    \caption{\textbf{More qualitative results on driving scenarios.} For each case, we show the initial observation and instruction, the imagined rollout of the selected plan against ground-truth future, and predicted trajectory against the reference.}
    \label{fig-visual_driving1}
\end{figure*}
\subsection{Implementation Details}
\label{sec:appx-impl}

\noindent\textbf{Diamond}~\citep{alonso2024diffusion}. We select Diamond as a world-model baseline representative of UNet-based diffusion architectures, adopting its best-performing configuration following the official implementation. Continuous actions are incorporated through a linear embedding layer for conditioning. To jointly support action prediction and imagination, we pair Diamond with LCVN-AC~\citep{dong2026language}, training LCVN-AC from scratch on the \bench{} train split, with Diamond as world model.

\noindent\textbf{NWM}~\citep{bar2025navigation}. We select NWM as a second world-model baseline, representative of DiT-based diffusion architectures. We adopt NWM's best reported configuration, a context length of 4 with a CDiT-XL backbone (1B parameters), training it from scratch on the \bench{} train split. At inference we follow the original paper and use 250 diffusion denoising steps. As with Diamond, we pair NWM with LCVN-AC~\citep{dong2026language} to jointly support planning and simulation, training LCVN-AC from scratch on the \bench{} train split with NWM as the world model.

\noindent\textbf{RAE-NWM}~\citep{zhang2026rae}.~We follow the official implementation, but scale the CDiT backbone from its default DiT-B setting to DiT-XL (28 blocks, hidden size 1152, 16 heads) to match the capacity of NWM. The remaining design is unchanged: a shallow-and-wide DDT head, a context window of 4 frames, and a time-driven gate that scales the action and horizon embeddings as a function of the flow time before they are injected through AdaLN. The transition model is trained on the \bench{} train split under our shared action interface with the flow-matching objective for 50 epochs, using AdamW with a batch size of 96, no weight decay, and a learning rate decayed linearly from $2\times10^{-4}$ to $2\times10^{-6}$. At inference we use an Euler ODE solver with 50 integration steps. As with NWM, we pair RAE-NWM with LCVN-AC~\citep{dong2026language}, trained from scratch on the \bench{} train split with RAE-NWM as the world model.

\noindent\textbf{NWM (lang)}. NWM (lang) shares the NWM configuration above and also injects CLIP-encoded instructions through the AdaLN layers of each CDiT block. It is paired with LCVN-AC in the same way as NWM.

\begin{figure*}[!t]
    \centering
    \includegraphics[width=1\linewidth]{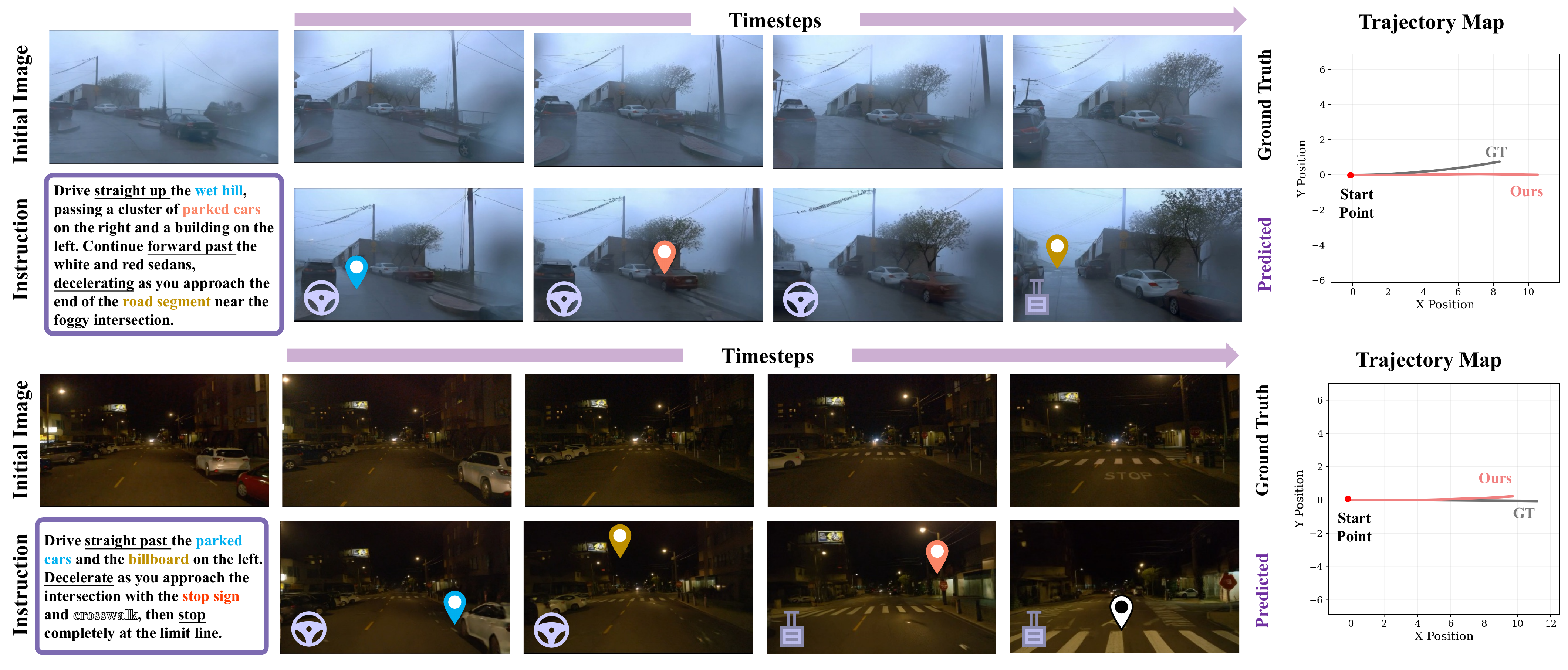}
    \caption{\textbf{More qualitative results on driving scenarios.} For each case, we show the initial observation and instruction, the imagined rollout of the selected plan against ground-truth future, and predicted trajectory against the reference.}
    \label{fig-visual_driving2}
    \vspace{-0.2cm}
\end{figure*}
\begin{figure*}[!t]
    \centering
    \includegraphics[width=1\linewidth]{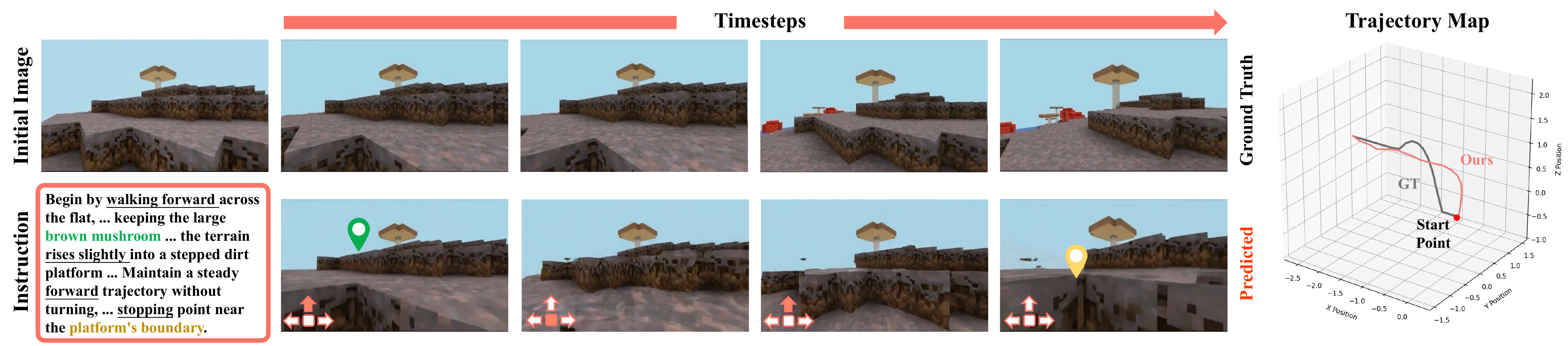}
    \caption{\textbf{More qualitative results on 3D gaming scenarios.} For each case, we show the initial observation and instruction, the imagined rollout of the selected plan against ground-truth future, and predicted trajectory against the reference.}
    \label{fig-visual_game1}
    \vspace{-0.2cm}
\end{figure*}

\begin{figure*}[!t]
    \centering
    \includegraphics[width=1\linewidth]{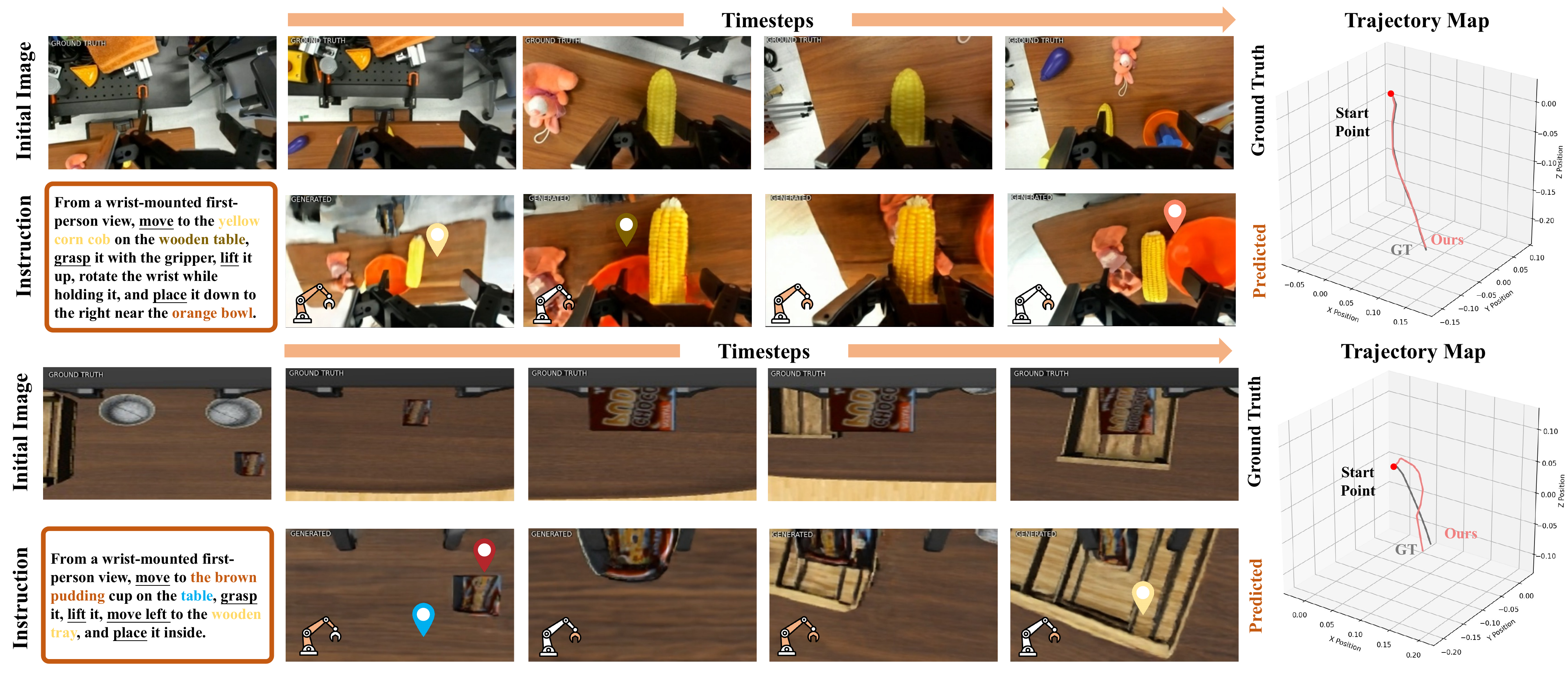}
    \caption{\textbf{More qualitative results on first-person-view robotics scenarios.} For each case, we show initial observation and instruction, imagined rollout of selected plan against ground-truth future, and predicted trajectory against reference.}
    \label{fig-visual_robot1}
    \vspace{-0.2cm}
\end{figure*}

\begin{figure*}[!t]
    \centering
    \includegraphics[width=1\linewidth]{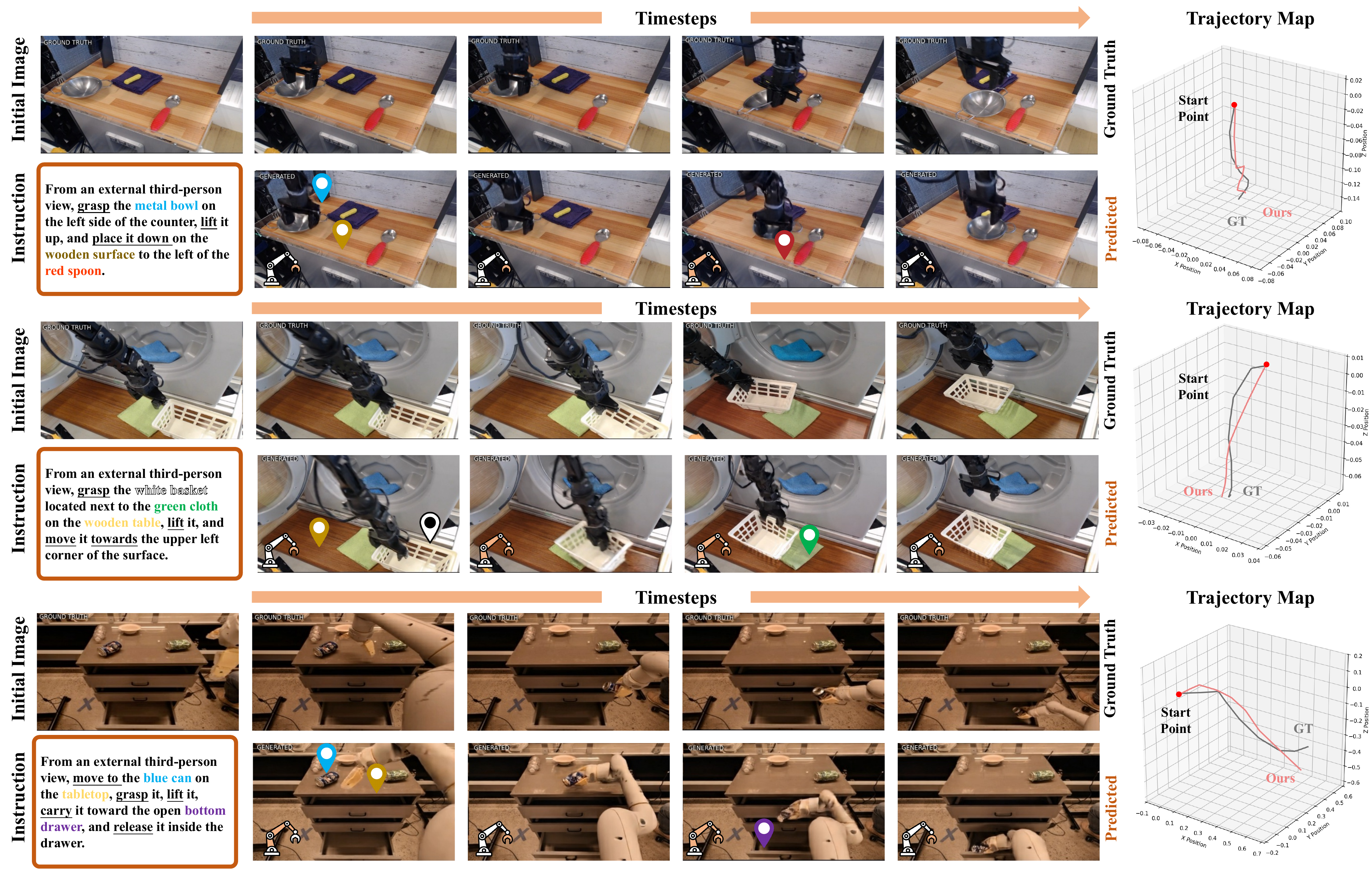}
    \caption{\textbf{More qualitative results on third-person-view robotics scenarios.} For each case, we show initial observation and instruction, imagined rollout of selected plan against ground-truth future, and predicted trajectory against reference.}
    \label{fig-visual_robot2}
    \vspace{-0.2cm}
\end{figure*}

\noindent\textbf{WorldVLA}~\citep{cen2025worldvla}.~WorldVLA is a unified autoregressive model that acts as both an action model and a world model within a single Chameleon-based backbone, representing images, text, and actions as discrete tokens. We follow its original setting, including the bin action discretization, the action attention mask, and the joint objective $\mathcal{L}=\mathcal{L}_{\mathrm{action}}+\alpha\mathcal{L}_{\mathrm{world}}$ with $\alpha=0.04$, but do not use the released checkpoints: the model is retrained on the \bench{} train split under our shared action interface.

\begin{table*}[!t]
\setlength{\abovecaptionskip}{2pt}
\setlength{\belowcaptionskip}{0pt}
\centering
\small
\setlength{\tabcolsep}{17pt}
\caption{Ablation on the \textbf{candidate funnel} on val seen. \textit{Top}: $N \rightarrow k \rightarrow 1$ samples $N$ candidates, imagines the Tr-Ranker top-$k$, and selects one trajectory. \textit{Middle}: Tr-Ranker versus random shortlisting at matched imagination budgets. \textit{Bottom}: ranker score fusion with the default $128 \rightarrow 8$ funnel. Bold marks the default configuration.}
\begin{tabular}{lcccc}
\thickhline
\rowcolor{gray!20}Setting & SR$\uparrow$ & nDTW$\uparrow$ & ATE$\downarrow$ & RPE$\downarrow$ \\
\hline\hline
\rowcolor{gray!10}\multicolumn{5}{l}{\textit{Funnel shape $N \rightarrow k \rightarrow 1$ (stage one: Tr-Ranker)}} \\
\rowcolor{white} $16 \rightarrow 1$   & \ms{0.508}{0.022} & \ms{0.604}{0.005} & \ms{0.797}{0.010} & \ms{0.155}{0.001} \\
$16 \rightarrow 4$   & \ms{0.545}{0.016} & \ms{0.622}{0.005} & \ms{0.791}{0.011} & \ms{0.152}{0.003} \\
$16 \rightarrow 8$   & \ms{0.560}{0.032} & \ms{0.636}{0.002} & \ms{0.766}{0.001} & \ms{0.146}{0.001} \\
$32 \rightarrow 1$   & \ms{0.516}{0.002} & \ms{0.610}{0.007} & \ms{0.786}{0.013} & \ms{0.153}{0.002} \\
$32 \rightarrow 4$   & \ms{0.552}{0.005} & \ms{0.628}{0.006} & \ms{0.783}{0.009} & \ms{0.150}{0.002} \\
$32 \rightarrow 8$   & \ms{0.565}{0.011} & \ms{0.641}{0.004} & \ms{0.762}{0.008} & \ms{0.145}{0.002} \\
$32 \rightarrow 16$  & \ms{0.570}{0.012} & \ms{0.646}{0.006} & \ms{0.757}{0.013} & \ms{0.143}{0.002} \\
$64 \rightarrow 1$   & \ms{0.524}{0.005} & \ms{0.620}{0.004} & \ms{0.778}{0.005} & \ms{0.148}{0.001} \\
$64 \rightarrow 4$   & \ms{0.556}{0.005} & \ms{0.641}{0.004} & \ms{0.770}{0.006} & \ms{0.145}{0.001} \\
$64 \rightarrow 8$   & \ms{0.578}{0.017} & \ms{0.648}{0.005} & \ms{0.731}{0.007} & \ms{0.139}{0.002} \\
$64 \rightarrow 16$  & \ms{0.584}{0.010} & \ms{0.652}{0.004} & \ms{0.726}{0.007} & \ms{0.136}{0.001} \\
$64 \rightarrow 32$  & \ms{0.588}{0.012} & \ms{0.654}{0.005} & \ms{0.722}{0.010} & \ms{0.134}{0.001} \\
$128 \rightarrow 1$  & \ms{0.530}{0.020} & \ms{0.640}{0.004} & \ms{0.770}{0.007} & \ms{0.132}{0.002} \\
$128 \rightarrow 4$  & \ms{0.580}{0.019} & \ms{0.648}{0.005} & \ms{0.726}{0.007} & \ms{0.131}{0.002} \\
\rowcolor{myblue!80}$\mathbf{128 \rightarrow 8}$ & \textbf{\ms{0.602}{0.013}} & \textbf{\ms{0.658}{0.004}} & \textbf{\ms{0.685}{0.005}} & \textbf{\ms{0.123}{0.001}} \\
$128 \rightarrow 16$ & \ms{0.608}{0.010} & \ms{0.660}{0.004} & \ms{0.682}{0.006} & \ms{0.121}{0.001} \\
$128 \rightarrow 32$ & \ms{0.606}{0.026} & \ms{0.659}{0.004} & \ms{0.684}{0.007} & \ms{0.122}{0.002} \\
$128 \rightarrow 64$ & \ms{0.603}{0.022} & \ms{0.658}{0.003} & \ms{0.686}{0.008} & \ms{0.123}{0.002} \\
$256 \rightarrow 1$  & \ms{0.534}{0.026} & \ms{0.645}{0.005} & \ms{0.765}{0.010} & \ms{0.130}{0.002} \\
$256 \rightarrow 8$  & \ms{0.606}{0.025} & \ms{0.662}{0.005} & \ms{0.683}{0.006} & \ms{0.120}{0.002} \\
$256 \rightarrow 16$ & \ms{0.607}{0.018} & \ms{0.666}{0.004} & \ms{0.680}{0.005} & \ms{0.117}{0.001} \\
$256 \rightarrow 32$ & \ms{0.611}{0.022} & \ms{0.667}{0.005} & \ms{0.678}{0.006} & \ms{0.117}{0.001} \\
$512 \rightarrow 1$  & \ms{0.535}{0.028} & \ms{0.646}{0.005} & \ms{0.764}{0.010} & \ms{0.129}{0.002} \\
\hline
\rowcolor{gray!10} \multicolumn{5}{l}{\textit{Stage-one selector at matched imagination budget ($N{=}128$)}} \\

\rowcolor{white} Random $\rightarrow 4$  & \ms{0.270}{0.025} & \ms{0.550}{0.002} & \ms{1.050}{0.007} & \ms{0.160}{0.001} \\
Random $\rightarrow 8$  & \ms{0.390}{0.016} & \ms{0.590}{0.003} & \ms{0.980}{0.016} & \ms{0.140}{0.001} \\
Random $\rightarrow 16$ & \ms{0.460}{0.013} & \ms{0.620}{0.002} & \ms{0.930}{0.002} & \ms{0.130}{0.001} \\
\rowcolor{myblue!80}$\mathbf{Tr \rightarrow 8\text{ (default)}}$ & \textbf{\ms{0.602}{0.013}} & \textbf{\ms{0.658}{0.004}} & \textbf{\ms{0.685}{0.005}} & \textbf{\ms{0.123}{0.001}} \\
\hline
\rowcolor{gray!10}\multicolumn{5}{l}{\textit{Fusion weights $\lambda_{\mathrm{tr}}:\lambda_{\mathrm{v}}$ ($128 \rightarrow 8$)}} \\
\rowcolor{white} $1:0$          & \ms{0.530}{0.020} & \ms{0.640}{0.004} & \ms{0.770}{0.007} & \ms{0.132}{0.002} \\
\rowcolor{myblue!80}$\mathbf{2:1}$ & \textbf{\ms{0.602}{0.013}} & \textbf{\ms{0.658}{0.004}} & \textbf{\ms{0.685}{0.005}} & \textbf{\ms{0.123}{0.001}} \\
$1:1$          & \ms{0.570}{0.018} & \ms{0.650}{0.004} & \ms{0.730}{0.006} & \ms{0.120}{0.001} \\
$1:2$          & \ms{0.520}{0.020} & \ms{0.640}{0.005} & \ms{0.810}{0.008} & \ms{0.130}{0.002} \\
$0:1$          & \ms{0.470}{0.011} & \ms{0.610}{0.006} & \ms{0.950}{0.010} & \ms{0.150}{0.003} \\
\hline
\end{tabular}
\label{tab:abl_funnel}
\vspace{-0.2cm}
\end{table*}

\noindent\textbf{LCVN-WM}~\citep{dong2026language}. We adopt the official implementation and training recipe of LCVN-WM, retraining it from scratch on the \bench{} train split under our shared action interface. Following their configuration, an ImageVAE~\citep{kingma2013auto} first compresses input images into $32\times32$ latent representations, and a DiT-XL backbone (approximately 1B parameters, hidden size 1152) is trained with diffusion forcing~\citep{chen2024diffusion} so that each frame retains an independent noise level. We use the AdamW optimizer~\citep{loshchilov2017decoupled} with linear warmup and a constant learning rate of $1\times10^{-4}$, and sample with the deterministic DDIM sampler~\citep{song2020denoising} for 50 steps, following~\citep{dong2026language}.

\noindent\textbf{LCVN-Uni}~\citep{dong2026language}. We adopt the official implementation of LCVN-Uni, which fine-tunes GAIR Anole-7B~\citep{chern2024anole} with a 4096-token context while keeping the bin, text, and image tokenizers frozen. We retrain it on the \bench{} train split under their reported configuration: input images are resized to $448\times448$ (height $\times$ width) and discretized into 784 visual tokens for context sizes of 1 or 2, and 625 tokens for a context size of 4. Following their setup, only the LoRA~\citep{hu2022lora} adapters (rank 16) in the Transformer's \textit{qkv} projections are updated~\citep{liu2023llava}. Optimization uses AdamW for 20 epochs at a learning rate of $2\times10^{-4}$, on 8$\times$NVIDIA A100 GPUs (80GB each) with a global batch size of 8.

\noindent \textbf{Tiresias-Agent}.~The world model is initialized from Wan2.2-TI2V-5B and the LDP is trained from scratch; umT5, DINOv3, and the VAE remain frozen throughout training. Visual encoders process all observations at a uniform $256\times256$ resolution, and scene features are extracted using a frozen DINOv3 ViT-L/16 distilled encoder (300M parameters). Training and evaluation run on Ubuntu 24.04.4 LTS with Python 3.10.20, PyTorch 2.4.1, and CUDA 11.8, on 8$\times$NVIDIA A100 80GB PCIe GPUs (driver 580.173.02). The world model is trained for 10 epochs on \bench{}. At inference, the LDP samples $N{=}128$ candidates using fixed noise seeds, the Tr-Ranker shortlists $k{=}8$ candidates, and the two scores are fused with $\lambda_{\mathrm{tr}}{:}\lambda_{\mathrm{v}}=2{:}1$.

\noindent\textbf{Selectors (Tab.~5 of the main paper).} All selectors operate on the same eight LDP candidates sampled at random per episode. \emph{Random} executes one candidate chosen uniformly. \emph{Consensus} returns the step-wise mean of the eight candidate action sequences. \emph{Cluster-vote} groups the candidates by agglomerative clustering on their pairwise DTW distances and returns the medoid of the largest cluster. \emph{Tr-Ranker} and \emph{V-Ranker} select the candidate with the highest $S_{\mathrm{tr}}$ and $S_{\mathrm{v}}$, respectively, and \emph{Tr + V} applies the fusion rule of Sec.~4.4 of the main paper with $\lambda_{\mathrm{tr}}{:}\lambda_{\mathrm{v}}=2{:}1$. \emph{Tr-Ranker + Direct} tests whether scene access alone explains the gain of imagination: it keeps the V-Ranker's architecture, contextual inputs (instruction embedding, scene features $f_s$, and domain embedding), training targets, and fusion with $S_{\mathrm{tr}}$, but replaces the rollout descriptor with an embedding of the candidate action sequence, so that no world-model rollout is required. The \emph{Qwen3.6 verifier} scores the same imagined rollouts as the V-Ranker: following MVoT-style reasoning~\citep{li2025imagine}, the model is given the instruction, the initial observation, and uniformly sampled frames of each decoded rollout, reasons step by step over the imagined frames, and outputs a fulfillment score; the highest-scoring candidate is selected. \emph{Oracle-of-8} selects the best candidate under the ground-truth metrics, preferring a successful candidate and breaking ties by nDTW, and serves as an upper bound. Likewise, Oracle-of-$N$ in Fig.~7a of the main paper counts an episode as successful if any of the $N$ candidates succeeds.

\subsection{More Experimental Results}

\noindent \textbf{Full Results of the Candidate Funnel Ablation.}~Tab.~\ref{tab:abl_funnel} reports the complete numerical results behind Figs.~7b--d in the main paper. The table lists all four planning metrics (SR, nDTW, ATE, and RPE) with the standard deviation over runs as a subscript. The top block enumerates the evaluated funnel shapes $N \rightarrow k \rightarrow 1$ over pool sizes $N \in \{16, \dots, 512\}$ and shortlist sizes $k \in \{1, \dots, 64\}$; the middle block compares Tr-Ranker against random shortlisting at matched imagination budgets with $N{=}128$; and the bottom block sweeps the fusion weights $\lambda_{\mathrm{tr}}{:}\lambda_{\mathrm{v}}$ under the default $128 \rightarrow 8$ funnel. The $N \rightarrow 1$ rows correspond to selection with the Tr-Ranker alone (equivalently, fusion weight $1{:}0$), and the $128 \rightarrow k$ rows together with the Random $\rightarrow k$ rows provide the Final SR columns of Tab.~6. The trends are consistent with those discussed in Sec.~5.3: performance improves with larger pools and shortlists but saturates around $N{=}128$ and $k{=}8$, Tr-Ranker shortlisting outperforms random shortlisting even at twice the imagination budget, and the $2{:}1$ fusion yields the highest SR among all weightings. Beyond SR, the additional metrics show that the improvements in goal completion are accompanied by consistent reductions in trajectory error: ATE and RPE decrease as the funnel widens, confirming that ranked imagination improves the whole executed path rather than only the terminal position.

\begin{table}[t]
\setlength{\abovecaptionskip}{2pt}
\setlength{\belowcaptionskip}{0pt}
\centering
\tiny
\setlength{\tabcolsep}{12pt}
\caption{\textbf{Planning performance and inference latency}
on val seen.
Latency measures the time to produce the final action trajectory,
including candidate generation and selection where applicable.
For Tiresias-Agent, $N \rightarrow k \rightarrow 1$ imagines
$k$ shortlisted candidates before selecting one.}
\resizebox{0.8\linewidth}{!}{%
\begin{tabular}{llcc}
\thickhline
\rowcolor{gray!20}Method & Configuration & SR$\uparrow$ & Latency (s)$\downarrow$ \\
\hline\hline
NoMaD (lang) & -- & 0.32 & 0.7 \\
LCVN-Uni & -- & 0.47 & 85 \\
\hline
Tiresias-Agent & $128 \rightarrow 8 \rightarrow 1$ (default)
& 0.60 & 36 \\
Tiresias-Agent & $128 \rightarrow 2 \rightarrow 1$
& 0.55 & 9 \\
\hline
\end{tabular}}
\label{tab:latency_tradeoff}
\vspace{-0.2cm}
\end{table}

\noindent \textbf{Performance--Latency Trade-off.}~Tab.~\ref{tab:latency_tradeoff} reports planning performance and inference latency. The default $128 \rightarrow 8 \rightarrow 1$ pipeline achieves an SR of 0.60 with a latency of 36 s. Reducing the shortlist to two candidates lowers latency to 9 s, with an SR of 0.55, illustrating the trade-off between planning performance and imagination cost. Results for NoMaD (lang) and LCVN-Uni provide additional reference points under their respective configurations.

\noindent \textbf{Text Encoder.}~To assess sensitivity to the text encoder, we replace all umT5 text encoders in Tiresias-Agent with CLIP text encoders. This variant achieves an overall \bench{} score of 62 on val seen, compared with 63 for the default configuration. The small performance gap suggests that our framework remains effective with an alternative text encoder and that its gains do not depend solely on umT5.

\subsection{More Qualitative Results}
\label{appx-more-qual}

Figs.~\ref{fig-visual_nav1}--\ref{fig-visual_robot2} provide additional cases covering all four domains, both robotics viewpoints, and both real and simulated sources. Each panel includes the initial observation and instruction, the imagined rollout against the ground-truth future, and the predicted trajectory against the reference.

\noindent\textbf{Navigation.} Fig.~\ref{fig-visual_nav1} pairs a wide-angle campus scene with an off-road track held out entirely from training (Sec.~3.2). The rollouts place named landmarks on the sides the instruction specifies and preserve the source appearance, including mud and standing water characteristic of unseen domain.

\noindent\textbf{3D Gaming.} Fig.~\ref{fig-visual_game1} shows an episode whose terrain rises into a stepped platform, exercising the vertical channel that this domain adds in Sec.~3.1. The elevation change is visible in both the rollout and the 3D trajectory, and block-level geometry stays consistent as the camera advances.

\noindent\textbf{Driving.} Figs.~\ref{fig-visual_driving1} and~\ref{fig-visual_driving2} span daylight residential streets, fog, and night. Adverse conditions are preserved rather than normalized away, with fog density, nighttime illumination, and road markings all remaining plausible across the horizon, and the speed directives in each instruction are reflected in the rollout pace.

\noindent\textbf{Robotics.} Figs.~\ref{fig-visual_robot1} and~\ref{fig-visual_robot2} separate the two viewpoints. The wrist-mounted cases show the camera moving only with the end-effector, including a commanded wrist rotation while holding the object, while the third-person cases keep the camera fixed across the horizon as only the arm moves. Both figures contain a real and a simulated episode, and in neither does the viewpoint or the visual domain drift, consistent with the robotics system prompt. The full contact sequence is realized, up to releasing an object inside an open drawer.

\section{Limitations}
Our evaluation is conducted on \bench{}, which includes data from both real and simulated environments, but does not include experiments deploying Tiresias-Agent on physical robots. Its effectiveness on real hardware therefore remains to be validated. We leave deployment and evaluation on physical robotic platforms for navigation and manipulation to future work.

\clearpage\clearpage
\bibliographystyle{assets/plainnat}
\bibliography{main}

\begin{thebibliography}{83}
\providecommand{\natexlab}[1]{#1}
\providecommand{\url}[1]{\texttt{#1}}
\expandafter\ifx\csname urlstyle\endcsname\relax
  \providecommand{\doi}[1]{doi: #1}\else
  \providecommand{\doi}{doi: \begingroup \urlstyle{rm}\Url}\fi

\bibitem[Agarwal et~al.(2025)Agarwal, Ali, Bala, Balaji, Barker, Cai, Chattopadhyay, Chen, Cui, Ding, et~al.]{agarwal2025cosmos}
Niket Agarwal, Arslan Ali, Maciej Bala, Yogesh Balaji, Erik Barker, Tiffany Cai, Prithvijit Chattopadhyay, Yongxin Chen, Yin Cui, Yifan Ding, et~al.
\newblock Cosmos world foundation model platform for physical {AI}.
\newblock \emph{arXiv preprint arXiv:2501.03575}, 2025.

\bibitem[Alakuijala et~al.(2024)Alakuijala, McLean, Woungang, Farsad, Kaski, Marttinen, and Yuan]{alakuijala2025video}
Minttu Alakuijala, Reginald McLean, Isaac Woungang, Nariman Farsad, Samuel Kaski, Pekka Marttinen, and Kai Yuan.
\newblock Video-language critic: Transferable reward functions for language-conditioned robotics.
\newblock \emph{arXiv preprint arXiv:2405.19988}, 2024.

\bibitem[Alonso et~al.(2024)Alonso, Jelley, Micheli, Kanervisto, Storkey, Pearce, and Fleuret]{alonso2024diffusion}
Eloi Alonso, Adam Jelley, Vincent Micheli, Anssi Kanervisto, Amos~J Storkey, Tim Pearce, and Fran{\c{c}}ois Fleuret.
\newblock Diffusion for world modeling: Visual details matter in {Atari}.
\newblock \emph{Advances in Neural Information Processing Systems}, 37:\penalty0 58757--58791, 2024.

\bibitem[Assran et~al.(2023)Assran, Duval, Misra, Bojanowski, Vincent, Rabbat, LeCun, and Ballas]{assran2023self}
Mahmoud Assran, Quentin Duval, Ishan Misra, Piotr Bojanowski, Pascal Vincent, Michael Rabbat, Yann LeCun, and Nicolas Ballas.
\newblock Self-supervised learning from images with a joint-embedding predictive architecture.
\newblock In \emph{Proceedings of the IEEE/CVF Conference on Computer Vision and Pattern Recognition}, pages 15619--15629, 2023.

\bibitem[Baldassarre et~al.(2025)Baldassarre, Szafraniec, Terver, Khalidov, Massa, LeCun, Labatut, Seitzer, and Bojanowski]{baldassarre2025back}
Federico Baldassarre, Marc Szafraniec, Basile Terver, Vasil Khalidov, Francisco Massa, Yann LeCun, Patrick Labatut, Maximilian Seitzer, and Piotr Bojanowski.
\newblock Back to the features: {DINO} as a foundation for video world models.
\newblock \emph{arXiv preprint arXiv:2507.19468}, 2025.

\bibitem[Bar et~al.(2025)Bar, Zhou, Tran, Darrell, and LeCun]{bar2025navigation}
Amir Bar, Gaoyue Zhou, Danny Tran, Trevor Darrell, and Yann LeCun.
\newblock Navigation world models.
\newblock In \emph{Proceedings of the IEEE/CVF Conference on Computer Vision and Pattern Recognition}, pages 15791--15801, 2025.

\bibitem[Brohan et~al.(2022)Brohan, Brown, Carbajal, Chebotar, Dabis, Finn, Gopalakrishnan, Hausman, Herzog, Hsu, et~al.]{brohan2022rt}
Anthony Brohan, Noah Brown, Justice Carbajal, Yevgen Chebotar, Joseph Dabis, Chelsea Finn, Keerthana Gopalakrishnan, Karol Hausman, Alex Herzog, Jasmine Hsu, et~al.
\newblock Rt-1: Robotics transformer for real-world control at scale.
\newblock \emph{arXiv preprint arXiv:2212.06817}, 2022.

\bibitem[Brooks et~al.(2024)Brooks, Peebles, Holmes, DePue, Guo, Jing, Schnurr, Taylor, Luhman, Luhman, et~al.]{brooks2024video}
Tim Brooks, Bill Peebles, Connor Holmes, Will DePue, Yufei Guo, Li~Jing, David Schnurr, Joe Taylor, Troy Luhman, Eric Luhman, et~al.
\newblock Video generation models as world simulators.
\newblock \emph{OpenAI Blog}, 2024.

\bibitem[Bruce et~al.(2024)Bruce, Dennis, Edwards, Parker-Holder, Shi, Hughes, Lai, Mavalankar, Steigerwald, Apps, et~al.]{bruce2024genie}
Jake Bruce, Michael~D Dennis, Ashley Edwards, Jack Parker-Holder, Yuge Shi, Edward Hughes, Matthew Lai, Aditi Mavalankar, Richie Steigerwald, Chris Apps, et~al.
\newblock Genie: Generative interactive environments.
\newblock In \emph{International Conference on Machine Learning}, 2024.

\bibitem[Caesar et~al.(2020)Caesar, Bankiti, Lang, Vora, Liong, Xu, Krishnan, Pan, Baldan, and Beijbom]{caesar2020nuscenes}
Holger Caesar, Varun Bankiti, Alex~H Lang, Sourabh Vora, Venice~Erin Liong, Qiang Xu, Anush Krishnan, Yu~Pan, Giancarlo Baldan, and Oscar Beijbom.
\newblock nuscenes: A multimodal dataset for autonomous driving.
\newblock In \emph{IEEE/CVF Conf. Comput. Vis. Pattern Recog.}, pages 11621--11631, 2020.

\bibitem[Cen et~al.(2025)Cen, Yu, Yuan, Jiang, Huang, Guo, Li, Song, Luo, Wang, et~al.]{cen2025worldvla}
Jun Cen, Chaohui Yu, Hangjie Yuan, Yuming Jiang, Siteng Huang, Jiayan Guo, Xin Li, Yibing Song, Hao Luo, Fan Wang, et~al.
\newblock Worldvla: Towards autoregressive action world model.
\newblock \emph{arXiv preprint arXiv:2506.21539}, 2025.

\bibitem[Chen et~al.(2024)Chen, Mart{\'\i}~Mons{\'o}, Du, Simchowitz, Tedrake, and Sitzmann]{chen2024diffusion}
Boyuan Chen, Diego Mart{\'\i}~Mons{\'o}, Yilun Du, Max Simchowitz, Russ Tedrake, and Vincent Sitzmann.
\newblock Diffusion forcing: Next-token prediction meets full-sequence diffusion.
\newblock \emph{Advances in Neural Information Processing Systems}, 37:\penalty0 24081--24125, 2024.

\bibitem[Chern et~al.(2024)Chern, Su, Ma, and Liu]{chern2024anole}
Ethan Chern, Jiadi Su, Yan Ma, and Pengfei Liu.
\newblock Anole: An open, autoregressive, native large multimodal models for interleaved image-text generation.
\newblock \emph{arXiv preprint arXiv:2407.06135}, 2024.

\bibitem[Chi et~al.(2025)Chi, Xu, Feng, Cousineau, Du, Burchfiel, Tedrake, and Song]{chi2025diffusion}
Cheng Chi, Zhenjia Xu, Siyuan Feng, Eric Cousineau, Yilun Du, Benjamin Burchfiel, Russ Tedrake, and Shuran Song.
\newblock Diffusion policy: Visuomotor policy learning via action diffusion.
\newblock \emph{The International Journal of Robotics Research}, 44\penalty0 (10-11):\penalty0 1684--1704, 2025.

\bibitem[Chi et~al.(2026{\natexlab{a}})Chi, Liang, Song, Yang, Li, Zhang, and Lv]{chi2026pv}
Haozhuang Chi, Jingsong Liang, Ziying Song, Lei Yang, Shihao Li, Haoruo Zhang, and Chen Lv.
\newblock Pv-wm: A heterogeneous micro-macro world model for articulated pedestrian-vehicle co-rollout.
\newblock \emph{arXiv preprint arXiv:2609.07328}, 2026{\natexlab{a}}.

\bibitem[Chi et~al.(2026{\natexlab{b}})Chi, Qiu, Su, Liu, Li, Zhang, and Lv]{chi2026driver}
Haozhuang Chi, Daosheng Qiu, Hao Su, Haochen Liu, Zirui Li, Haoruo Zhang, and Chen Lv.
\newblock Driver-wm: A driver-centric traffic-conditioned latent world model for in-cabin dynamics rollout.
\newblock \emph{arXiv preprint arXiv:2605.05092}, 2026{\natexlab{b}}.

\bibitem[Chu et~al.(2026)Chu, Zhang, Lin, Kong, Zhang, Tu, Ma, Huang, Yang, Huang, et~al.]{chu2026agentic}
Meng Chu, Xuan~Billy Zhang, Kevin~Qinghong Lin, Lingdong Kong, Jize Zhang, Teng Tu, Weijian Ma, Ziqi Huang, Senqiao Yang, Wei Huang, et~al.
\newblock Agentic world modeling: Foundations, capabilities, laws, and beyond.
\newblock \emph{arXiv preprint arXiv:2604.22748}, 2026.

\bibitem[Chung et~al.(2023)Chung, Garcia, Roberts, Tay, Firat, Narang, and Constant]{chung2023unimax}
Hyung~Won Chung, Xavier Garcia, Adam Roberts, Yi~Tay, Orhan Firat, Sharan Narang, and Noah Constant.
\newblock Unimax: Fairer and more effective language sampling for large-scale multilingual pretraining.
\newblock In \emph{The Eleventh International Conference on Learning Representations}, 2023.

\bibitem[Dong et~al.(2025)Dong, Wu, He, Cheng, Li, Li, Cheng, Zhou, Sun, Dai, et~al.]{dong2025ha}
Yifei Dong, Fengyi Wu, Qi~He, Zhi-Qi Cheng, Heng Li, Minghan Li, Zebang Cheng, Yuxuan Zhou, Jingdong Sun, Qi~Dai, et~al.
\newblock Ha-vln 2.0: An open benchmark and leaderboard for human-aware navigation in discrete and continuous environments with dynamic multi-human interactions.
\newblock \emph{arXiv preprint arXiv:2503.14229}, 2025.

\bibitem[Dong et~al.(2026{\natexlab{a}})Dong, Wu, Chen, Kong, Hu, Zhou, Zhu, Sun, He, Dai, Hauptmann, and Cheng]{dong2026towards}
Yifei Dong, Fengyi Wu, Guangyu Chen, Lingdong Kong, Qiyu Hu, Yuxuan Zhou, Xu~Zhu, Jingdong Sun, Jun-Yan He, Qi~Dai, Alexander~G. Hauptmann, and Zhi-Qi Cheng.
\newblock Towards unified world models for visual navigation via memory-augmented planning and foresight.
\newblock In \emph{European Conference on Computer Vision}, pages 276--295. Springer, 2026{\natexlab{a}}.

\bibitem[Dong et~al.(2026{\natexlab{b}})Dong, Wu, Dai, Kong, Chen, Sha, Hu, Liu, Huang, Dai, and Cheng]{dong2026language}
Yifei Dong, Fengyi Wu, Yilong Dai, Lingdong Kong, Guangyu Chen, Yetong Sha, Qiyu Hu, Feng Liu, Siyu Huang, Qi~Dai, and Zhi-Qi Cheng.
\newblock {Language-Conditioned World Modeling for Visual Navigation}.
\newblock In \emph{Advances in Neural Information Processing Systems}, 2026{\natexlab{b}}.
\newblock \url{https://arxiv.org/abs/2603.26741}.

\bibitem[Du et~al.(2023)Du, Yang, Dai, Dai, Nachum, Tenenbaum, Schuurmans, and Abbeel]{du2023learning}
Yilun Du, Sherry Yang, Bo~Dai, Hanjun Dai, Ofir Nachum, Josh Tenenbaum, Dale Schuurmans, and Pieter Abbeel.
\newblock Learning universal policies via text-guided video generation.
\newblock \emph{Adv. Neural Inf. Process. Syst.}, 36:\penalty0 9156--9172, 2023.

\bibitem[Duan et~al.(2025)Duan, Yu, Chen, Fei-Fei, and Wu]{duan2025worldscore}
Haoyi Duan, Hong-Xing Yu, Sirui Chen, Li~Fei-Fei, and Jiajun Wu.
\newblock Worldscore: A unified evaluation benchmark for world generation.
\newblock \emph{arXiv preprint arXiv:2504.00983}, 2025.

\bibitem[Fan et~al.(2026)Fan, Chi, Ju, Li, Bao, Wang, Chen, Jiang, Ge, Li, et~al.]{fan2026wow}
Chun-Kai Fan, Xiaowei Chi, Xiaozhu Ju, Hao Li, Yong Bao, Yu-Kai Wang, Lizhang Chen, Zhiyuan Jiang, Kuangzhi Ge, Ying Li, et~al.
\newblock Wow, wo, val! a comprehensive embodied world model evaluation turing test.
\newblock \emph{arXiv preprint arXiv:2601.04137}, 2026.

\bibitem[Fu et~al.(2023)Fu, Tamir, Sundaram, Chai, Zhang, Dekel, and Isola]{fu2023dreamsim}
Stephanie Fu, Netanel Tamir, Shobhita Sundaram, Lucy Chai, Richard Zhang, Tali Dekel, and Phillip Isola.
\newblock {DreamSim}: Learning new dimensions of human visual similarity using synthetic data.
\newblock \emph{arXiv preprint arXiv:2306.09344}, 2023.

\bibitem[Ha and Schmidhuber(2018)]{ha2018world}
David Ha and J{\"u}rgen Schmidhuber.
\newblock World models.
\newblock \emph{arXiv preprint arXiv:1803.10122}, 2018.

\bibitem[Hafner et~al.(2023)Hafner, Pasukonis, Ba, and Lillicrap]{hafner2024masteringdiversedomainsworld}
Danijar Hafner, Jurgis Pasukonis, Jimmy Ba, and Timothy Lillicrap.
\newblock Mastering diverse domains through world models.
\newblock \emph{arXiv preprint arXiv:2301.04104}, 2023.

\bibitem[Hirose et~al.(2018)Hirose, Sadeghian, V{\'a}zquez, Goebel, and Savarese]{hirose2018gonet}
Noriaki Hirose, Amir Sadeghian, Marynel V{\'a}zquez, Patrick Goebel, and Silvio Savarese.
\newblock {GONet}: A semi-supervised deep learning approach for traversability estimation.
\newblock In \emph{IEEE/RSJ International Conference on Intelligent Robots and Systems}, pages 3044--3051, 2018.

\bibitem[Hirose et~al.(2023)Hirose, Shah, Sridhar, and Levine]{hirose2023sacson}
Noriaki Hirose, Dhruv Shah, Ajay Sridhar, and Sergey Levine.
\newblock {SACSoN}: Scalable autonomous control for social navigation.
\newblock \emph{IEEE Robotics and Automation Letters}, 9\penalty0 (1):\penalty0 49--56, 2023.

\bibitem[Hu et~al.(2022)Hu, Shen, Wallis, Allen-Zhu, Li, Wang, Wang, Chen, et~al.]{hu2022lora}
Edward~J Hu, Yelong Shen, Phillip Wallis, Zeyuan Allen-Zhu, Yuanzhi Li, Shean Wang, Lu~Wang, Weizhu Chen, et~al.
\newblock {LoRA}: Low-rank adaptation of large language models.
\newblock \emph{International Conference on Learning Representations}, 2022.

\bibitem[Huang et~al.(2023)Huang, He, Yu, Zhang, Si, Jiang, Zhang, Wu, Jin, Chanpaisit, Wang, Chen, Wang, Lin, Qiao, and Liu]{huang2023vbenchcomprehensivebenchmarksuite}
Ziqi Huang, Yinan He, Jiashuo Yu, Fan Zhang, Chenyang Si, Yuming Jiang, Yuanhan Zhang, Tianxing Wu, Qingyang Jin, Nattapol Chanpaisit, Yaohui Wang, Xinyuan Chen, Limin Wang, Dahua Lin, Yu~Qiao, and Ziwei Liu.
\newblock Vbench: Comprehensive benchmark suite for video generative models, 2023.
\newblock \url{https://arxiv.org/abs/2311.17982}.

\bibitem[Huber(1992)]{huber1992robust}
Peter~J Huber.
\newblock Robust estimation of a location parameter.
\newblock In \emph{Breakthroughs in statistics: Methodology and distribution}, pages 492--518. Springer, 1992.

\bibitem[Ilharco et~al.(2019)Ilharco, Jain, Ku, Ie, and Baldridge]{ilharco2019general}
Gabriel Ilharco, Vihan Jain, Alexander Ku, Eugene Ie, and Jason Baldridge.
\newblock General evaluation for instruction conditioned navigation using dynamic time warping.
\newblock \emph{arXiv preprint arXiv:1907.05446}, 2019.

\bibitem[Joachims(2002)]{joachims2002optimizing}
Thorsten Joachims.
\newblock Optimizing search engines using clickthrough data.
\newblock In \emph{Proceedings of the eighth ACM SIGKDD international conference on Knowledge discovery and data mining}, pages 133--142, 2002.

\bibitem[Karnan et~al.(2022)Karnan, Nair, Xiao, Warnell, Pirk, Toshev, Hart, Biswas, and Stone]{karnan2022socially}
Haresh Karnan, Anirudh Nair, Xuesu Xiao, Garrett Warnell, S{\"o}ren Pirk, Alexander Toshev, Justin Hart, Joydeep Biswas, and Peter Stone.
\newblock Socially compliant navigation dataset ({SCAND}): A large-scale dataset of demonstrations for social navigation.
\newblock \emph{IEEE Robotics and Automation Letters}, 7\penalty0 (4):\penalty0 11807--11814, 2022.

\bibitem[Khazatsky et~al.(2024)Khazatsky, Pertsch, Nair, Balakrishna, Dasari, Karamcheti, Nasiriany, Srirama, Chen, Ellis, et~al.]{khazatsky2024droid}
Alexander Khazatsky, Karl Pertsch, Suraj Nair, Ashwin Balakrishna, Sudeep Dasari, Siddharth Karamcheti, Soroush Nasiriany, Mohan~Kumar Srirama, Lawrence~Yunliang Chen, Kirsty Ellis, et~al.
\newblock Droid: A large-scale in-the-wild robot manipulation dataset.
\newblock \emph{arXiv preprint arXiv:2403.12945}, 2024.

\bibitem[Kingma and Welling(2013)]{kingma2013auto}
Diederik~P Kingma and Max Welling.
\newblock Auto-encoding variational {Bayes}.
\newblock \emph{arXiv preprint arXiv:1312.6114}, 2013.

\bibitem[Kolagar and Zarcone(2024)]{kolagar2024aligning}
Zahra Kolagar and Alessandra Zarcone.
\newblock Aligning uncertainty: Leveraging llms to analyze uncertainty transfer in text summarization.
\newblock In \emph{Workshop on Uncertainty-Aware NLP}, pages 41--61, 2024.

\bibitem[Kong et~al.(2024)Kong, Chen, Wang, Su, Hu, Yang, and Liu]{kong2024controllable}
Xianghao Kong, Jinyu Chen, Wenguan Wang, Hang Su, Xiaolin Hu, Yi~Yang, and Si~Liu.
\newblock Controllable navigation instruction generation with chain of thought prompting.
\newblock In \emph{European Conference on Computer Vision}, pages 37--54. Springer, 2024.

\bibitem[Kwon et~al.(2026)Kwon, Kim, Go, and Baek]{kwon2025stableworldmodelsmeasuring}
Soonwoo Kwon, Jin-Young Kim, Hyojun Go, and Kyungjune Baek.
\newblock Toward stable world models: Measuring and addressing world instability in generative environments.
\newblock \emph{Pattern Recognition}, page 113351, 2026.

\bibitem[Li et~al.(2025)Li, Wu, Zhang, Xia, Mao, Dong, Vuli{\'c}, and Wei]{li2025imagine}
Chengzu Li, Wenshan Wu, Huanyu Zhang, Yan Xia, Shaoguang Mao, Li~Dong, Ivan Vuli{\'c}, and Furu Wei.
\newblock Imagine while reasoning in space: Multimodal visualization-of-thought.
\newblock \emph{arXiv preprint arXiv:2501.07542}, 2025.

\bibitem[Li et~al.(2024)Li, Li, Wang, Lan, Yu, Ji, Li, Zhu, Kautz, Wu, et~al.]{li2024hydra}
Zhenxin Li, Kailin Li, Shihao Wang, Shiyi Lan, Zhiding Yu, Yishen Ji, Zhiqi Li, Ziyue Zhu, Jan Kautz, Zuxuan Wu, et~al.
\newblock Hydra-mdp: End-to-end multimodal planning with multi-target hydra-distillation.
\newblock \emph{arXiv preprint arXiv:2406.06978}, 2024.

\bibitem[Lian et~al.(2025)Lian, Cai, Du, and Liang]{lian2025toward}
Kewei Lian, Shaofei Cai, Yilun Du, and Yitao Liang.
\newblock Toward memory-aided world models: Benchmarking via spatial consistency.
\newblock \emph{arXiv preprint arXiv:2505.22976}, 2025.

\bibitem[Liang et~al.(2025)Liang, Kong, Yan, Liu, Yang, Huang, Yin, Zuo, Hu, Zhu, Lu, Liu, Jiang, Li, Li, Zhuo, Ng, Cottereau, Gao, Pan, Ooi, and Liu]{worldlens}
Ao~Liang, Lingdong Kong, Tianyi Yan, Hongsi Liu, Wesley Yang, Ziqi Huang, Wei Yin, Jialong Zuo, Yixuan Hu, Dekai Zhu, Dongyue Lu, Youquan Liu, Guangfeng Jiang, Linfeng Li, Xiangtai Li, Long Zhuo, Lai~Xing Ng, Benoit~R. Cottereau, Changxin Gao, Liang Pan, Wei~Tsang Ooi, and Ziwei Liu.
\newblock {WorldLens}: Full-spectrum evaluations of driving world models in real world.
\newblock \emph{arXiv preprint arXiv:2512.10958}, 2025.

\bibitem[Liao et~al.(2022)Liao, Xie, and Geiger]{liao2022kitti}
Yiyi Liao, Jun Xie, and Andreas Geiger.
\newblock Kitti-360: A novel dataset and benchmarks for urban scene understanding in 2d and 3d.
\newblock \emph{IEEE Transactions on Pattern Analysis and Machine Intelligence}, 45\penalty0 (3):\penalty0 3292--3310, 2022.

\bibitem[Liu et~al.(2023{\natexlab{a}})Liu, Zhu, Gao, Feng, Liu, Zhu, and Stone]{liu2023libero}
Bo~Liu, Yifeng Zhu, Chongkai Gao, Yihao Feng, Qiang Liu, Yuke Zhu, and Peter Stone.
\newblock Libero: Benchmarking knowledge transfer for lifelong robot learning.
\newblock \emph{arXiv preprint arXiv:2306.03310}, 2023{\natexlab{a}}.

\bibitem[Liu et~al.(2023{\natexlab{b}})Liu, Li, Wu, and Lee]{liu2023llava}
Haotian Liu, Chunyuan Li, Qingyang Wu, and Yong~Jae Lee.
\newblock Visual instruction tuning.
\newblock \emph{Advances in Neural Information Processing Systems}, 36:\penalty0 34892--34916, 2023{\natexlab{b}}.

\bibitem[Liu et~al.(2026{\natexlab{a}})Liu, Liu, Yang, Jiang, Li, Zhang, Wang, Nex, and Cheng]{liu2026streamvlo}
Mengmeng Liu, Jiuming Liu, Michael~Ying Yang, Chaokang Jiang, Jiangtao Li, Yunpeng Zhang, Hesheng Wang, Francesco Nex, and Hao Cheng.
\newblock Streamvlo: Streaming visual-lidar odometry with cumulative drift compensation.
\newblock In \emph{Proceedings of the IEEE/CVF Conference on Computer Vision and Pattern Recognition}, pages 39086--39097, 2026{\natexlab{a}}.

\bibitem[Liu et~al.(2026{\natexlab{b}})Liu, Zhang, Liu, Cui, Xie, Chen, Ye, Nex, Cheng, and Yang]{liu2026universe}
Mengmeng Liu, Diankun Zhang, Jiuming Liu, Jianfeng Cui, Hongwei Xie, Guang Chen, Hangjun Ye, Francesco Nex, Hao Cheng, and Michael~Ying Yang.
\newblock Universe: Unified video action models for autonomous driving with flexible mask-modulated modality generation.
\newblock \emph{arXiv preprint arXiv:2607.05133}, 2026{\natexlab{b}}.

\bibitem[Liu et~al.(2026{\natexlab{c}})Liu, Zhang, Liu, Cui, Xie, Chen, Ye, Yang, Nex, and Cheng]{liu2026driveva}
Mengmeng Liu, Diankun Zhang, Jiuming Liu, Jianfeng Cui, Hongwei Xie, Guang Chen, Hangjun Ye, Michael~Ying Yang, Francesco Nex, and Hao Cheng.
\newblock Driveva: Video action models are zero-shot drivers.
\newblock In \emph{European Conference on Computer Vision}, pages 315--335. Springer, 2026{\natexlab{c}}.

\bibitem[Liu et~al.(2025)Liu, Liu, Zhang, Xia, Wang, Shen, Jin, Ding, and Yu]{liu2025grl}
Yuze Liu, Tingjie Liu, Tiehua Zhang, Youhua Xia, Jinze Wang, Zhishu Shen, Jiong Jin, Zhijun Ding, and Fei~Richard Yu.
\newblock Grl-prompt: Towards prompts optimization via graph-empowered reinforcement learning using llms’ feedback.
\newblock In \emph{Pacific-Asia Conference on Knowledge Discovery and Data Mining}, pages 426--438. Springer, 2025.

\bibitem[Loshchilov and Hutter(2017)]{loshchilov2017decoupled}
Ilya Loshchilov and Frank Hutter.
\newblock Decoupled weight decay regularization.
\newblock \emph{arXiv preprint arXiv:1711.05101}, 2017.

\bibitem[Mees et~al.(2022)Mees, Hermann, Rosete-Beas, and Burgard]{mees2022calvinbenchmarklanguageconditionedpolicy}
Oier Mees, Lukas Hermann, Erick Rosete-Beas, and Wolfram Burgard.
\newblock Calvin: A benchmark for language-conditioned policy learning for long-horizon robot manipulation tasks, 2022.
\newblock \url{https://arxiv.org/abs/2112.03227}.

\bibitem[Pearce and Zhu(2022)]{pearce2021csgo}
Tim Pearce and Jun Zhu.
\newblock Counter-strike deathmatch with large-scale behavioural cloning.
\newblock In \emph{2022 IEEE Conference on Games (CoG)}, pages 104--111. IEEE, 2022.

\bibitem[Peebles and Xie(2023)]{peebles2023scalable}
William Peebles and Saining Xie.
\newblock Scalable diffusion models with transformers.
\newblock In \emph{Proceedings of the IEEE/CVF International Conference on Computer Vision}, pages 4195--4205, 2023.

\bibitem[{Qwen Team}(2026)]{qwen3.6-27b}
{Qwen Team}.
\newblock {Qwen3.6-27B}: Flagship-level coding in a {27B} dense model, April 2026.
\newblock \url{https://qwen.ai/blog?id=qwen3.6-27b}.

\bibitem[Shah et~al.(2021)Shah, Eysenbach, Kahn, Rhinehart, and Levine]{shah2021rapid}
Dhruv Shah, Benjamin Eysenbach, Gregory Kahn, Nicholas Rhinehart, and Sergey Levine.
\newblock Rapid exploration for open-world navigation with latent goal models.
\newblock \emph{arXiv preprint arXiv:2104.05859}, 2021.

\bibitem[Shah et~al.(2023)Shah, Sridhar, Bhorkar, Hirose, and Levine]{shah2022gnm}
Dhruv Shah, Ajay Sridhar, Arjun Bhorkar, Noriaki Hirose, and Sergey Levine.
\newblock Gnm: A general navigation model to drive any robot.
\newblock In \emph{2023 IEEE International Conference on Robotics and Automation (ICRA)}, pages 7226--7233. IEEE, 2023.

\bibitem[Shang et~al.(2026)Shang, Li, Ma, Su, Jin, Wang, Jin, Zhang, Tang, Su, et~al.]{shang2026worldarena}
Yu~Shang, Zhuohang Li, Yiding Ma, Weikang Su, Xin Jin, Ziyou Wang, Lei Jin, Xin Zhang, Yinzhou Tang, Haisheng Su, et~al.
\newblock Worldarena: A unified benchmark for evaluating perception and functional utility of embodied world models.
\newblock \emph{arXiv preprint arXiv:2602.08971}, 2026.

\bibitem[Sim{\'e}oni et~al.(2025)Sim{\'e}oni, Vo, Seitzer, Baldassarre, Oquab, Jose, Khalidov, Szafraniec, Yi, Ramamonjisoa, et~al.]{simeoni2025dinov3}
Oriane Sim{\'e}oni, Huy~V Vo, Maximilian Seitzer, Federico Baldassarre, Maxime Oquab, Cijo Jose, Vasil Khalidov, Marc Szafraniec, Seungeun Yi, Micha{\"e}l Ramamonjisoa, et~al.
\newblock Dinov3.
\newblock \emph{arXiv preprint arXiv:2508.10104}, 2025.

\bibitem[Song et~al.(2020)Song, Meng, and Ermon]{song2020denoising}
Jiaming Song, Chenlin Meng, and Stefano Ermon.
\newblock Denoising diffusion implicit models.
\newblock \emph{arXiv preprint arXiv:2010.02502}, 2020.

\bibitem[Sridhar et~al.(2024)Sridhar, Shah, Glossop, and Levine]{sridhar2024nomad}
Ajay Sridhar, Dhruv Shah, Catherine Glossop, and Sergey Levine.
\newblock {NoMaD}: Goal masked diffusion policies for navigation and exploration.
\newblock In \emph{IEEE International Conference on Robotics and Automation}, pages 63--70, 2024.

\bibitem[Sturm et~al.(2012)Sturm, Burgard, and Cremers]{sturm2012evaluating}
J{\"u}rgen Sturm, Wolfram Burgard, and Daniel Cremers.
\newblock Evaluating egomotion and structure-from-motion approaches using the {TUM} {RGB-D} benchmark.
\newblock In \emph{IEEE/RSJ International Conference on Intelligent Robot Systems Workshops}, volume~13, page~6, 2012.

\bibitem[Sun et~al.(2020)Sun, Kretzschmar, Dotiwalla, Chouard, Patnaik, Tsui, Guo, Zhou, Chai, Caine, et~al.]{sun2020scalability}
Pei Sun, Henrik Kretzschmar, Xerxes Dotiwalla, Aurelien Chouard, Vijaysai Patnaik, Paul Tsui, James Guo, Yin Zhou, Yuning Chai, Benjamin Caine, et~al.
\newblock Scalability in perception for autonomous driving: Waymo open dataset.
\newblock In \emph{IEEE/CVF Conf. Comput. Vis. Pattern Recog.}, pages 2446--2454, 2020.

\bibitem[Triest et~al.(2022)Triest, Sivaprakasam, Wang, Wang, Johnson, and Scherer]{triest2022tartandrive}
Samuel Triest, Matthew Sivaprakasam, Sean~J Wang, Wenshan Wang, Aaron~M Johnson, and Sebastian Scherer.
\newblock {TartanDrive}: A large-scale dataset for learning off-road dynamics models.
\newblock In \emph{IEEE International Conference on Robotics and Automation}, pages 2546--2552, 2022.

\bibitem[Unterthiner et~al.(2019)Unterthiner, Van~Steenkiste, Kurach, Marinier, Michalski, and Gelly]{unterthiner2019fvd}
Thomas Unterthiner, Sjoerd Van~Steenkiste, Karol Kurach, Rapha{\"e}l Marinier, Marcin Michalski, and Sylvain Gelly.
\newblock Fvd: A new metric for video generation.
\newblock 2019.

\bibitem[Walke et~al.(2023)Walke, Black, Zhao, Vuong, Zheng, Hansen-Estruch, He, Myers, Kim, Du, et~al.]{walke2023bridgedata}
Homer~Rich Walke, Kevin Black, Tony~Z Zhao, Quan Vuong, Chongyi Zheng, Philippe Hansen-Estruch, Andre~Wang He, Vivek Myers, Moo~Jin Kim, Max Du, et~al.
\newblock Bridgedata v2: A dataset for robot learning at scale.
\newblock In \emph{Conference on Robot Learning}, pages 1723--1736. PMLR, 2023.

\bibitem[Wan et~al.(2025)Wan, Wang, Ai, Wen, Mao, Xie, Chen, Yu, Zhao, Yang, et~al.]{wan2025wan}
Team Wan, Ang Wang, Baole Ai, Bin Wen, Chaojie Mao, Chen-Wei Xie, Di~Chen, Feiwu Yu, Haiming Zhao, Jianxiao Yang, et~al.
\newblock Wan: Open and advanced large-scale video generative models.
\newblock \emph{arXiv preprint arXiv:2503.20314}, 2025.

\bibitem[Wang et~al.(2025)Wang, Zhang, Cui, Zhang, Shen, Liu, Ma, and Jin]{wang2025we}
Jinze Wang, Lu~Zhang, Yiyang Cui, Tiehua Zhang, Zhishu Shen, Yuze Liu, Xingjun Ma, and Jiong Jin.
\newblock Do we really need sft? prompt-as-policy over knowledge graphs for cold-start next poi recommendation.
\newblock \emph{arXiv preprint arXiv:2510.08012}, 2025.

\bibitem[Wang et~al.(2024)Wang, Cheng, He, Wang, Dai, Chen, Xia, and Zhang]{wang2024drivingdojo}
Yuqi Wang, Ke~Cheng, Jiawei He, Qitai Wang, Hengchen Dai, Yuntao Chen, Fei Xia, and Zhaoxiang Zhang.
\newblock Drivingdojo dataset: Advancing interactive and knowledge-enriched driving world model.
\newblock \emph{Advances in Neural Information Processing Systems}, 37:\penalty0 13020--13034, 2024.

\bibitem[Wu et~al.(2026{\natexlab{a}})Wu, Dong, Dai, Chen, Wu, Huang, Wang, Dai, Hauptmann, and Cheng]{wu2025govig}
Fengyi Wu, Yifei Dong, Yilong Dai, Guangyu Chen, Qifeng Wu, Huiting Huang, Hang Wang, Qi~Dai, Alexander~G Hauptmann, and Zhi-Qi Cheng.
\newblock Govig: Goal-conditioned visual navigation instruction generation via multimodal reasoning.
\newblock In \emph{Findings of the Association for Computational Linguistics: ACL 2026}, pages 14899--14920, 2026{\natexlab{a}}.

\bibitem[Wu et~al.(2026{\natexlab{b}})Wu, Cai, Zhao, Feng, Dang, Song, Tian, Zhu, Lei, Dou, Tang, Sun, Wu, Chu, Liu, and Huang]{wu2026omniworldbenchcomprehensiveinteractioncentricevaluation}
Meiqi Wu, Zhixin Cai, Fufangchen Zhao, Xiaokun Feng, Rujing Dang, Bingze Song, Ruitian Tian, Jiashu Zhu, Jiachen Lei, Hao Dou, Jing Tang, Lei Sun, Jiahong Wu, Xiangxiang Chu, Zeming Liu, and Kaiqi Huang.
\newblock Omni-worldbench: Towards a comprehensive interaction-centric evaluation for world models, 2026{\natexlab{b}}.
\newblock \url{https://arxiv.org/abs/2603.22212}.

\bibitem[Xiang et~al.(2024)Xiang, Liu, Gu, Gao, Ning, Zha, Feng, Tao, Hao, Shi, et~al.]{xiang2024pandora}
Jiannan Xiang, Guangyi Liu, Yi~Gu, Qiyue Gao, Yuting Ning, Yuheng Zha, Zeyu Feng, Tianhua Tao, Shibo Hao, Yemin Shi, et~al.
\newblock Pandora: Towards general world model with natural language actions and video states.
\newblock \emph{arXiv preprint arXiv:2406.09455}, 2024.

\bibitem[Xiong et~al.(2026{\natexlab{a}})Xiong, Song, Kang, Yan, Jiang, Yang, Fu, Fotiadis, Wang, Liu, et~al.]{xiong2026actworld}
Zhexiao Xiong, Yizhi Song, Hao Kang, Qing Yan, Liming Jiang, Jenson Yang, Zhoujie Fu, Stathi Fotiadis, Angtian Wang, Zichuan Liu, et~al.
\newblock Actworld: From explorable to interactive world model via action-aware memory.
\newblock \emph{arXiv preprint arXiv:2606.17730}, 2026{\natexlab{a}}.

\bibitem[Xiong et~al.(2026{\natexlab{b}})Xiong, Ye, Yaman, Cheng, Lu, Luo, Jacobs, and Ren]{xiong2026unidrive}
Zhexiao Xiong, Xin Ye, Burhan Yaman, Sheng Cheng, Yiren Lu, Jingru Luo, Nathan Jacobs, and Liu Ren.
\newblock Unidrive-wm: Unified understanding, planning and generation world model for autonomous driving.
\newblock \emph{arXiv preprint arXiv:2601.04453}, 2026{\natexlab{b}}.

\bibitem[Yu et~al.(2025)Yu, Qin, Wang, Wan, Zhang, and Liu]{yu2025gamefactory}
Jiwen Yu, Yiran Qin, Xintao Wang, Pengfei Wan, Di~Zhang, and Xihui Liu.
\newblock {GameFactory}: Creating new games with generative interactive videos.
\newblock \emph{arXiv preprint arXiv:2501.08325}, 2025.

\bibitem[Yue et~al.(2025)Yue, Huang, Liao, Chen, Zhou, Chen, Yao, and Ren]{yue2025ewmbench}
Hu~Yue, Siyuan Huang, Yue Liao, Shengcong Chen, Pengfei Zhou, Liliang Chen, Maoqing Yao, and Guanghui Ren.
\newblock Ewmbench: Evaluating scene, motion, and semantic quality in embodied world models.
\newblock \emph{arXiv preprint arXiv:2505.09694}, 2025.

\bibitem[Zhang et~al.(2025)Zhang, Jiang, Dai, Lu, Uzunoglu, Zhang, Wei, Wang, Patel, Liang, et~al.]{zhang2025world}
Jiahan Zhang, Muqing Jiang, Nanru Dai, Taiming Lu, Arda Uzunoglu, Shunchi Zhang, Yana Wei, Jiahao Wang, Vishal~M Patel, Paul~Pu Liang, et~al.
\newblock World-in-world: World models in a closed-loop world.
\newblock \emph{arXiv preprint arXiv:2510.18135}, 2025.

\bibitem[Zhang et~al.(2026)Zhang, Shen, Zhang, Qin, Pei, and Meng]{zhang2026rae}
Mingkun Zhang, Wangtian Shen, Fan Zhang, Haijian Qin, Zihao Pei, and Ziyang Meng.
\newblock Rae-nwm: Navigation world model in dense visual representation space.
\newblock In \emph{European Conference on Computer Vision}, pages 455--473. Springer, 2026.

\bibitem[Zhang et~al.(2018)Zhang, Isola, Efros, Shechtman, and Wang]{zhang2018unreasonable}
Richard Zhang, Phillip Isola, Alexei~A Efros, Eli Shechtman, and Oliver Wang.
\newblock The unreasonable effectiveness of deep features as a perceptual metric.
\newblock In \emph{Proceedings of the IEEE/CVF Conference on Computer Vision and Pattern Recognition}, pages 586--595, 2018.

\bibitem[Zhao et~al.(2025)Zhao, Wang, Zhu, Chen, Huang, Bao, and Wang]{zhao2025drivedreamer}
Guosheng Zhao, Xiaofeng Wang, Zheng Zhu, Xinze Chen, Guan Huang, Xiaoyi Bao, and Xingang Wang.
\newblock {DriveDreamer-2}: {LLM}-enhanced world models for diverse driving video generation.
\newblock In \emph{AAAI Conference on Artificial Intelligence}, volume~39, pages 10412--10420, 2025.

\bibitem[Zhou et~al.(2024)Zhou, Du, Chen, Li, Yeung, and Gan]{zhou2024robodreamer}
Siyuan Zhou, Yilun Du, Jiaben Chen, Yandong Li, Dit-Yan Yeung, and Chuang Gan.
\newblock Robodreamer: Learning compositional world models for robot imagination.
\newblock In \emph{Int. Conf. Mach. Learn.}, pages 61885--61896. PMLR, 2024.

\bibitem[Zhu et~al.(2024)Zhu, Wang, Zhao, Min, Deng, Dou, Wang, Shi, Wang, Zhang, et~al.]{zhu2024sora}
Zheng Zhu, Xiaofeng Wang, Wangbo Zhao, Chen Min, Nianchen Deng, Min Dou, Yuqi Wang, Botian Shi, Kai Wang, Chi Zhang, et~al.
\newblock Is {Sora} a world simulator? a comprehensive survey on general world models and beyond.
\newblock \emph{arXiv preprint arXiv:2405.03520}, 2024.

\end{thebibliography}

\end{document}